\documentclass[10pt,twocolumn,letterpaper]{article}

\usepackage{wacv}
\usepackage{times}
\usepackage{epsfig}
\usepackage{graphicx}
\usepackage{amsmath}
\usepackage{amssymb}
\usepackage{booktabs}
\usepackage{caption}
\usepackage{array}
\usepackage{tabularx}
\usepackage[dvipsnames]{xcolor}
\usepackage[skip=0cm,list=true]{subcaption}
\usepackage{url}
\usepackage{multicol}
\usepackage{makecell}
\usepackage{comment}
\usepackage[accsupp]{axessibility}

\makeatletter
\@namedef{ver@everyshi.sty}{}
\makeatother
\usepackage{tikz}

\usepackage{cite}

\usepackage{microtype}
\usepackage{colortbl}

\usepackage{soul}
\usepackage{todonotes}

\newenvironment{tightitemize} % Defines the tightitemize environment which modifies the itemize environment to be more compact
{\vspace{-\topsep}\begin{itemize}\itemsep1pt \parskip0pt \parsep0pt}
{\end{itemize}\vspace{-\topsep}}

\definecolor{gold(bg)}{rgb}{0.87, 0.79, 0.32}
\definecolor{gold(metallic)}{rgb}{0.83, 0.69, 0.22}
\definecolor{gold(web)(golden)}{rgb}{1.0, 0.84, 0.0}

\definecolor{palesilver}{rgb}{0.79, 0.75, 0.73}
\definecolor{silver}{rgb}{0.75, 0.75, 0.75}
\definecolor{lightslategray}{rgb}{0.47, 0.53, 0.6}

\definecolor{bronze(bg)}{rgb}{0.9, 0.6, 0.3}
\definecolor{bronze}{rgb}{0.8, 0.5, 0.2}

\newcolumntype{L}[1]{>{\raggedright\let\newline\\\arraybackslash\hspace{0pt}}b{#1}}
\newcolumntype{C}[1]{>{\centering\let\newline\\\arraybackslash\hspace{0pt}}b{#1}}

\newcommand{\rankn}[1]{({\small\##1})}

\def\wacvPaperID{****} % Enter the WACV Paper ID here

\wacvfinalcopy % *** Uncomment this line for the final submission

\ifwacvfinal
\usepackage[breaklinks=true,bookmarks=false,hidelinks=True]{hyperref}
\else
\usepackage[pagebackref=true,breaklinks=true,colorlinks,bookmarks=false]{hyperref}
\fi

\begin{document}

%%%%%%%%% TITLE
\title{1\textsuperscript{st} Workshop on Maritime Computer Vision (MaCVi) 2023: Challenge Results}

\author{Benjamin Kiefer$^1$, Matej Kristan$^2$, Janez Perš$^2$, Lojze Žust$^2$, Fabio Poiesi$^3$,\\ 
Fabio Augusto de Alcantara Andrade$^4$, Alexandre Bernardino$^5$, Matthew Dawkins$^6$, \\ Jenni Raitoharju$^7$, Yitong Quan$^1$, Adem Atmaca$^1$, Timon Höfer$^1$, \\
Qiming Zhang$^8$, Yufei Xu$^8$, Jing Zhang$^8$, Dacheng Tao$^{9,8}$, Lars Sommer$^{10}$, Raphael Spraul$^{10}$,\\ 
Hangyue Zhao$^{11}$, Hongpu Zhang$^{11}$, Yanyun Zhao$^{11}$, Jan Lukas Augustin$^{12}$,Eui-ik Jeon$^{13}$, \\ 
Impyeong Lee$^{13}$, Luca Zedda$^{14}$, Andrea Loddo$^{14}$, Cecilia Di Ruberto$^{14}$, Sagar Verma$^{15}$, \\ 
Siddharth Gupta$^{15}$, Shishir Muralidhara$^{16}$, Niharika Hegde$^{16}$, Daitao Xing$^{17}$, \\ 
Nikolaos Evangeliou$^{18}$, Anthony Tzes$^{18}$, Vojtěch Bartl$^{19}$, Jakub Špaňhel$^{19}$, Adam Herout$^{19}$, \\
Neelanjan Bhowmik$^{20}$, Toby P. Breckon$^{20}$, Shivanand Kundargi*$^{21}$,Tejas Anvekar*$^{21}$,\\
 Chaitra Desai$^{21}$, Ramesh Ashok Tabib$^{21}$, Uma Mudengudi$^{21}$, Arpita Vats$^{22}$, Yang Song$^{11}$,\\
Delong Liu$^{11}$, Yonglin Li$^{23}$, Shuman Li$^{23}$, Chenhao Tan$^{23}$, Long Lan$^{23}$,
Vladimir Somers$^{24,25,26}$,\\ Christophe De Vleeschouwer$^{25}$, Alexandre Alahi$^{24}$,
Hsiang-Wei Huang$^{27}$, Cheng-Yen Yang$^{27}$,\\ Jenq-Neng Hwang$^{27}$,
Pyong-Kun Kim$^{28}$, Kwangju Kim$^{28}$, Kyoungoh Lee$^{28}$,
Shuai Jiang$^{11}$,\\ Haiwen Li$^{11}$, Zheng Ziqiang$^{29}$, Tuan-Anh Vu$^{29}$, Hai Nguyen-Truong$^{29}$,\\ Sai-Kit Yeung$^{29}$, Zhuang Jia$^{30}$,
Sophia Yang$^{31}$, Chih-Chung Hsu$^{32}$, \\ Xiu-Yu Hou$^{32}$, Yu-An Jhang$^{32}$, Simon Yang$^{33}$, Mau-Tsuen Yang$^{34}$
{\tt\small  }
%add technical reports authors
\and
$^1$University of Tuebingen, 
$^2$University of Ljubljana, 
$^3$Fondazione Bruno Kessler, 
\\$^4$University of South-Eastern Norway, 
$^5$University of Lisbon, 
$^6$Kitware, {\tt\small }
$^7$University of Jyväskylä,\\
$^8$The University of Sydney, 
$^9$JD Explore Academy, 
$^{10}$Fraunhofer IOSB, \\
$^{11}$Beijing University of Posts and Telecommunications, 
$^{12}$Helmut Schmidt University, \\
$^{13}$University of Seoul, 
$^{14}$University of Cagliari, 
$^{15}$Granular AI, 
$^{16}$TU Kaiserslautern,\\
$^{17}$New York University, 
$^{18}$New York University Abu Dhabi,\\
$^{19}$Brno University of Technology, 
$^{20}$Durham University,
$^{21}$KLE Technological University,\\
$^{22}$Santa Clara University,
$^{23}$National University of Defense Technology,
$^{24}$EPFL,
$^{25}$UCLouvain,\\
$^{26}$Sportradar,
$^{27}$University of Washington,
$^{28}$Electronics and Telecommunications Research Institute,\\
$^{29}$Hong Kong University of Science and Technology,
$^{30}$Xiaomi Inc,
$^{31}$National Tsing Hua University,\\
$^{32}$National Cheng Kung University,
$^{33}$National Taiwan Ocean University,\\
$^{34}$National Dong-Hwa University}
\maketitle
\thispagestyle{empty}

\begin{abstract}
\vspace{0mm}
The 1$^{\text{st}}$ Workshop on Maritime Computer Vision (MaCVi) 2023 focused on maritime computer vision for Unmanned Aerial Vehicles (UAV) and Unmanned Surface Vehicle (USV), and organized several subchallenges in this domain: (i) UAV-based Maritime Object Detection, (ii) UAV-based Maritime Object Tracking, (iii) USV-based Maritime Obstacle Segmentation and (iv) USV-based Maritime Obstacle Detection. The subchallenges were based on the SeaDronesSee and MODS benchmarks. 
%This report summarizes the main findings of the individual subchallenges, which are (1) UAV-based Maritime Object Detection, (2) UAV-based Maritime Object Tracking, (3) USV-based Maritime Obstacle Segmentation and (4) USV-based Maritime Obstacle Detection. 
This report summarizes the main findings of the individual subchallenges and introduces 
%Furthermore, we introduce 
a new benchmark, called SeaDronesSee Object Detection v2, which extends the previous benchmark by including more classes and footage. We provide statistical and qualitative analyses, and assess trends in the best-performing methodologies of over 130 submissions. The methods are summarized in the appendix.
%The tech report for most of the top performing methods is attached. 
The datasets, evaluation code and the %competition's final standing 
leaderboard are publicly available (\url{https://seadronessee.cs.uni-tuebingen.de/macvi}).
\end{abstract}

%%%%%%%%% BODY TEXT
\section{Introduction}

The open water covers over 70\% of our planet and accounts for 80\% of international trade \cite{United_Nations_Conference_on_Trade_and_Development2021-dz}. 
The use of cameras and robotic platforms in this domain is growing fast, and ocean scientists are gathering large amounts of visual data using various sensors with a clear need for robust and reliable methods for quantification, detection, classification and understanding \cite{varga2022seadronessee,marques2015unmanned,zhai2016novel,prasad2019object,kanjir2018vessel,gallego2019detection,miri2022guide,kaur2022sea}. 
Evidence of this is also reported in recent literature surveys \cite{zhang2021survey,miri2022guide,bo2021ship}.

\begin{figure}[t]
\centering
\begin{subfigure}{.35\textwidth}
   \includegraphics[width=\textwidth]{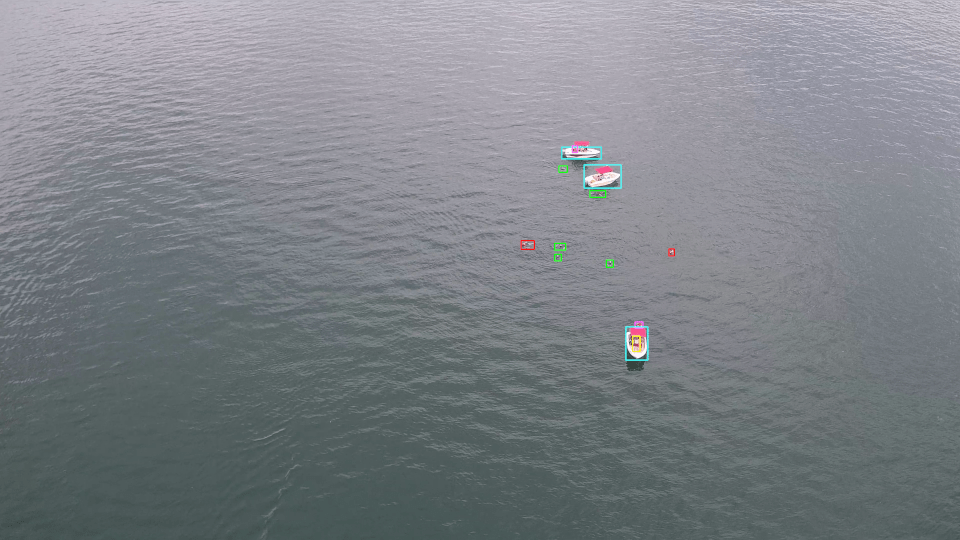}
   \caption{UAV-based Maritime Object Detection}
\label{fig:g1}
\end{subfigure}%
\hfill
\begin{subfigure}{.35\textwidth}
   \includegraphics[width=\textwidth]{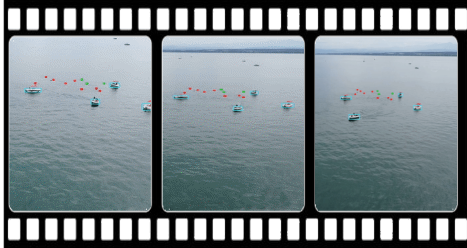}
   \caption{UAV-based Maritime Object Tracking}
\label{fig:g2}
\end{subfigure}
\begin{subfigure}{.35\textwidth}
   \includegraphics[width=\textwidth]{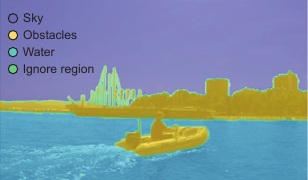}
   \caption{USV-based Maritime Obstacle Segmentation}
\label{fig:g3}
\end{subfigure}
\begin{subfigure}{.35\textwidth}
   \includegraphics[width=\textwidth]{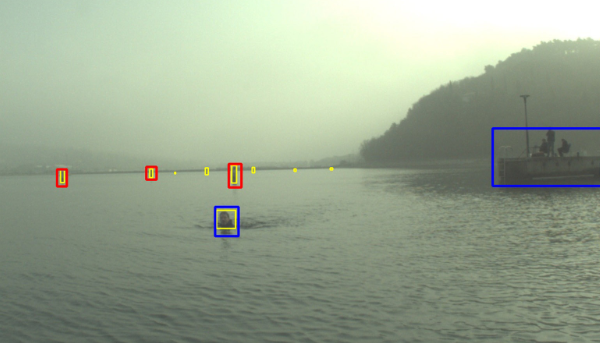}
   \caption{USV-based Maritime Obstacle Detection}
\label{fig:g4}
\end{subfigure}
\caption{Overview of MaCVi challenges.}
\label{fig:challenges_overview}
\end{figure}

In particular, significant efforts have been invested in recent decades into development of autonomous robots that operate on and above the water surface. Unmanned surface vehicles (USVs) are emerging from this research in form of autonomous boats and ships. Their autonomy profoundly depends on perception capability, particularly in busy maritime traffic, near the coast or in inland waters. Despite important advances made in maritime computer vision for USVs~\cite{Bovcon2021,Nigrudkar2021,Muhovic2020,nunes2022real}, this remains an unsolved problem.
Another class of autonomous robots that is emerging are Unmanned Aerial Vehicles (UAVs), or drones, which provide an aerial view over a scenery, allowing to oversee large areas quickly and in a relatively inexpensive manner. While computer vision methods are not as crucial for navigation in UAVs, perception capabilities are required for automated perimeter inspection \cite{zhang2017automatic}. 

%On the one hand, Unmanned Surface Vehicles (USVs) are used for environmental monitoring tasks,
%~\cite{marinelife}. 
%a crucial element for autonomous operation of autonomous boats and ships is environmental perception. 
%On the other hand, Unmanned Aerial Vehicles (UAVs) provide an aerial view over a scenery, allowing to oversee large areas quickly and in a relatively inexpensive manner. 
%USVs provide a different view of the scenery, they are slower, but are capable of significantly longer mission duration.
%Differently from UAVs, USVs require computer vision methods for their own navigation as well, especially in busy maritime traffic, near the coast or in inland waters. This is still an unsolved problem, despite being an active research area~\cite{Bovcon2021,Nigrudkar2021,Muhovic2020,nunes2022real}.

\begin{figure*}[tb]
\centering
\includegraphics[width=\textwidth]{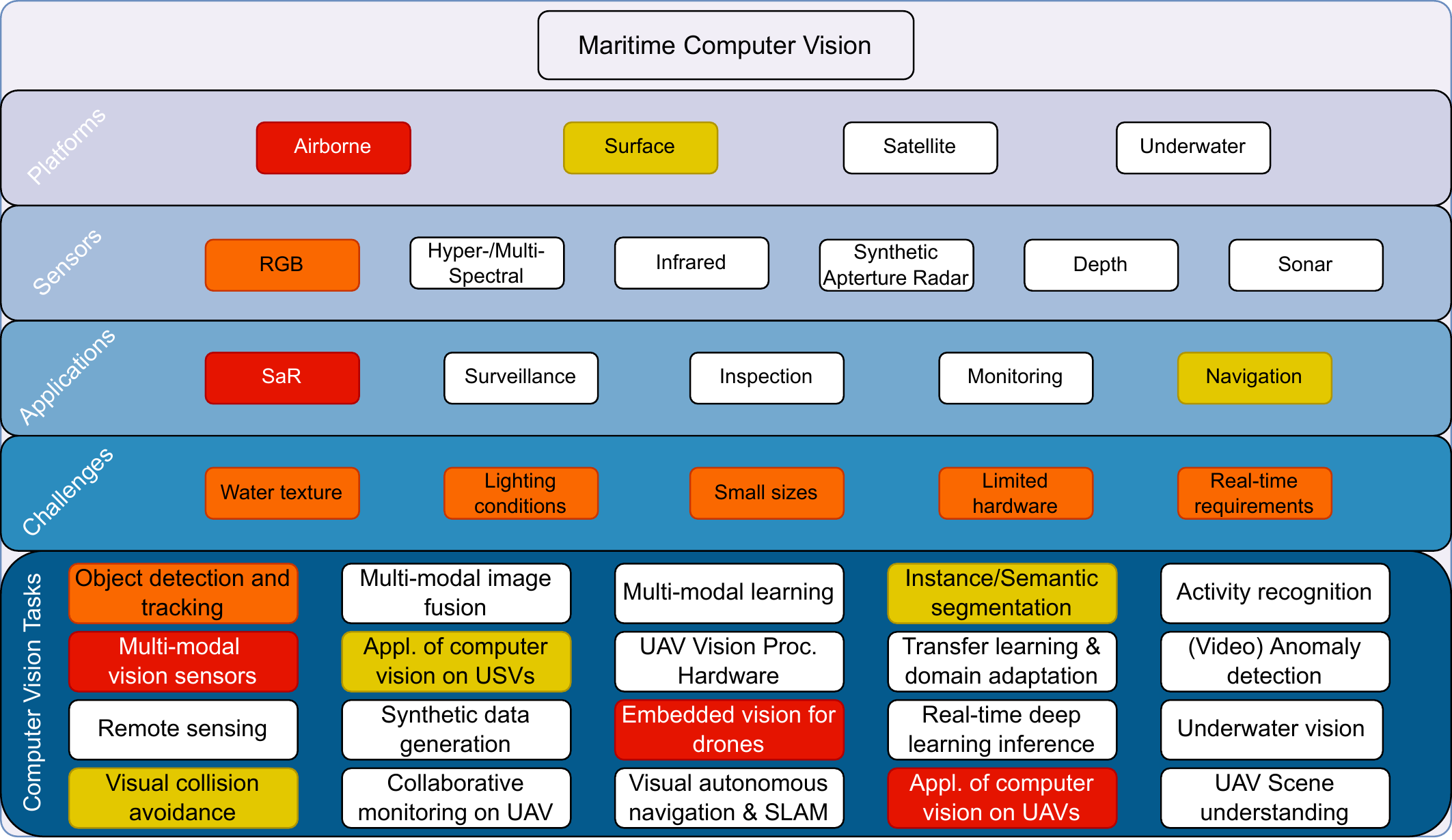}
\caption{Overview diagram of what the maritime computer vision is comprised of and what applications the four challenges, UAV-based Object Detection \& Tracking (red) and USV-based Obstacle Detection \& Segmentation (yellow), are aimed at tackling (orange applies to both).
\label{fig:applicationsoverview}
}
\end{figure*}

Indeed, UAVs and USVs cater a wide range of maritime applications, such as maritime Search and Rescue (SaR) \cite{varga2022seadronessee,lygouras2019unsupervised}, maritime patrol \cite{xu2014bgcut,panico2020adaptive}, monitoring of oil and sewage spills from ships \cite{guangbo2019oil,han2021automatic}, trash detection \cite{cheng2021flow,syed2021computationally}, illegal fishing prevention \cite{prayudi2020surveillance,bloom2019drones}, animal population surveying \cite{hyun2020remotely,mustafa2019detecting}, wind farm and oil rig inspection \cite{xu2019wind,car2020autonomous}, and coral reef monitoring \cite{bennett2020automating,david2021structure} to name a few.
All of these applications require robust vision systems for UAVs and USVs for practical use. 
Therefore, the maritime domain poses several unique challenges:
\begin{itemize}
    \item {\bf Water texture:} Naturally, the most distinguishing property comes from the water surface itself. Sea foam or waves are unpredictable and inhibit reliable detection. Furthermore, sun reflections result in random artifacts in the case of standard, thermal and multi-spectral cameras. While the water surface seems to be homogeneous, it differs drastically between different bodies of water. Furthermore, plants, animals, trash and other confounders make the detection even harder.
    
    \item {\bf Lighting conditions:} Relative camera orientation with respect to the sun position affects the apparent scene lighting. Acute angles with horizon visible may render certain image areas underexposed while other overexposed and saturated.
    
%    Partly related to the water texture is the difficulty in changes of lighting conditions, which results in sun reflections as a consequence from certain viewing angles of the camera. Furthermore, for acute angles that make the horizon visible, the exposure time varies between different areas in the image. 
    %On the other hand, the homogeneous water surface 
    
    \item {\bf Size of objects and obstacles:} UAVs fly at high altitudes to increase the field of view, which makes objects appear very small.
%    n many use cases, objects appear small when observed by a UAv from a  high altitude. 
    %of interest are typically very small as UAVs fly a high altitude to cover large areas. 
    This requires models to operate with large resolutions and large foreground-background imbalances. For USVs, a comparable situation is detection of small crafts from a vantage point of a large ship.% that cannot change its course quickly.
    
    \item {\bf Limited Hardware:} Typical relevant UAVs have a small payload and a limited power supply, only allowing embedded hardware to be deployed. In long-distance missions, where it is not possible to transmit a video stream, this restricts the use of computer vision algorithms to small models. Meanwhile, USVs \emph{must} process most of the sensory data on board in real-time for sailing control and timely obstacle avoidance. %reaction and to get real-time information, which is then used to control the vehicle.
    
    \item {\bf Real-time requirements:} Maritime SaR missions and other applications require models running in real-time, such that there are no false negatives and a quick response is possible. Furthermore, the high speed of UAVs demand fast synchronization between navigation sensors and potentially multiple cameras to allow for georeferencing or tracking applications, and in both, UAVs and USVs, real-time requirements are crucial for navigation itself.

\end{itemize}

To address these challenges in a way that unites many maritime applications and to spark interest in the maritime domain, the 1\textsuperscript{st} Workshop on Maritime Computer Vision (MaCVi) 2023 was organized in conjunction with the IEEE/CVF Winter Conference on Applications of Computer Vision (WACV) 2023.
An integral part of the workshop were the challenges listed in Figure \ref{fig:challenges_overview}, i.e.~UAV-based Object Detection \& Tracking, and USV-based Obstacle Detection \& Segmentation. 

The first group of challenge tracks are geared towards Maritime Search and Rescue (SaR) applications, where the footage aims to simulate such scenarios (see Figure \ref{fig:applicationsoverview} for a challenge categorization). These two tracks are mainly based on the SeaDronesSee benchmark \cite{varga2022seadronessee}, although we extended it considerably in the case of the object detection part. 
Sections \ref{sec:uavobjectdetectionchallenge} and \ref{sec:uavobjecttrackingchallenge} describe in detail the dataset and the challenge of these tracks. 
The second group of challenge tracks are aimed at autonomous boats applications. They resemble real-world application challenges in the context of unmanned water vehicles. The challenges are based on the MODS benchmark \cite{MODSBenchmark2022} and the challenge tracks and dataset will be described in Section \ref{sec:usvchallenges}.

%See Figure \ref{fig:applicationsoverview} for an overview of the maritime computer vision domain with its applications and where the challenges can be categorized into.

The rest of the paper is organized as follows. 
First, we provide an overview of the challenge protocol before we review the outcomes of the individual challenge tracks with their underlying benchmarks and datasets.

\section{Challenge Participation Protocol}

The challenge tracks were announced on the 20th of August 2022 and ran until the 25th of October 2022. At the announcement date, participants could download the datasets and evaluation and visualization toolkits from the workshop homepage\footnote{\url{https://seadronessee.cs.uni-tuebingen.de/wacv23}}. 
Participants could experiment with their methods on this data before they could upload their predictions on the individual tracks' test sets on the webserver from the 14th of September onwards. The predictions were compared with the corresponding ground truth annotations on the server-side. Lastly, participants could choose to show their result on the leaderboard or to delete the submission.

At the start of the uploading phase, participants were allowed to upload predictions three times per day independent of the challenge track. The submitted predictions were said to be subject to further inspection on our side regarding the exact performances and participants were required to provide information on their used methods, and in the USV-based tracks, participants were required to submit their code as well. The respective metrics for the individual challenge tracks decided whether the submission reached a top-3 position. Furthermore, we required every participant to submit information on the speed of their method measured in frames per second wall clock time and their hardware. Lastly, participants needed to indicate which data sets (also for pretraining) they used during training.

Additionally, the teams that reached a performance above our least performing baseline were asked to submit a short technical report, describing their methods and training configurations. 
These reports are attached to this paper.

\subsection{Evaluation Server}

The evaluation server is an extended version of original webserver for the SeaDronesSee benchmark.  
The updated version of the evaluation server has been available online for several months before the start of the challenges.
In addition to the challenge tracks, it also supports the following tracks: 
Boat-MNIST (toy dataset for image classification), 
UAV-based Object Detection v1, 
Single-Object Tracking and 
a variant of the Multi-Object Tracking task focusing on swimmers only.

\section{UAV-based Object Detection Challenge}
\label{sec:uavobjectdetectionchallenge}

The goal of this challenge was to 
% advance computer vision algorithms in Maritime Search and Rescue missions. 
% It is aimed at detecting 
detect humans, boats and other objects in open water. 
The task of object detection in maritime SaR is far from solved. 
For example, the best performing model of the SeaDronesSee object detection track currently achieves 36\% mAP, as opposed to the COCO benchmark with the best performer achieving over 60\% mAP. 
SeaDronesSee is more challenging due to lighting conditions and sun reflections, different appearances coming from various altitudes and viewing angles. 
While the sparsity of object locations often results in false positives, the small sizes of objects along partial occlusion due to water lead to false negatives.

\begin{figure}[t]
\centering
   \includegraphics[width=0.46\textwidth]{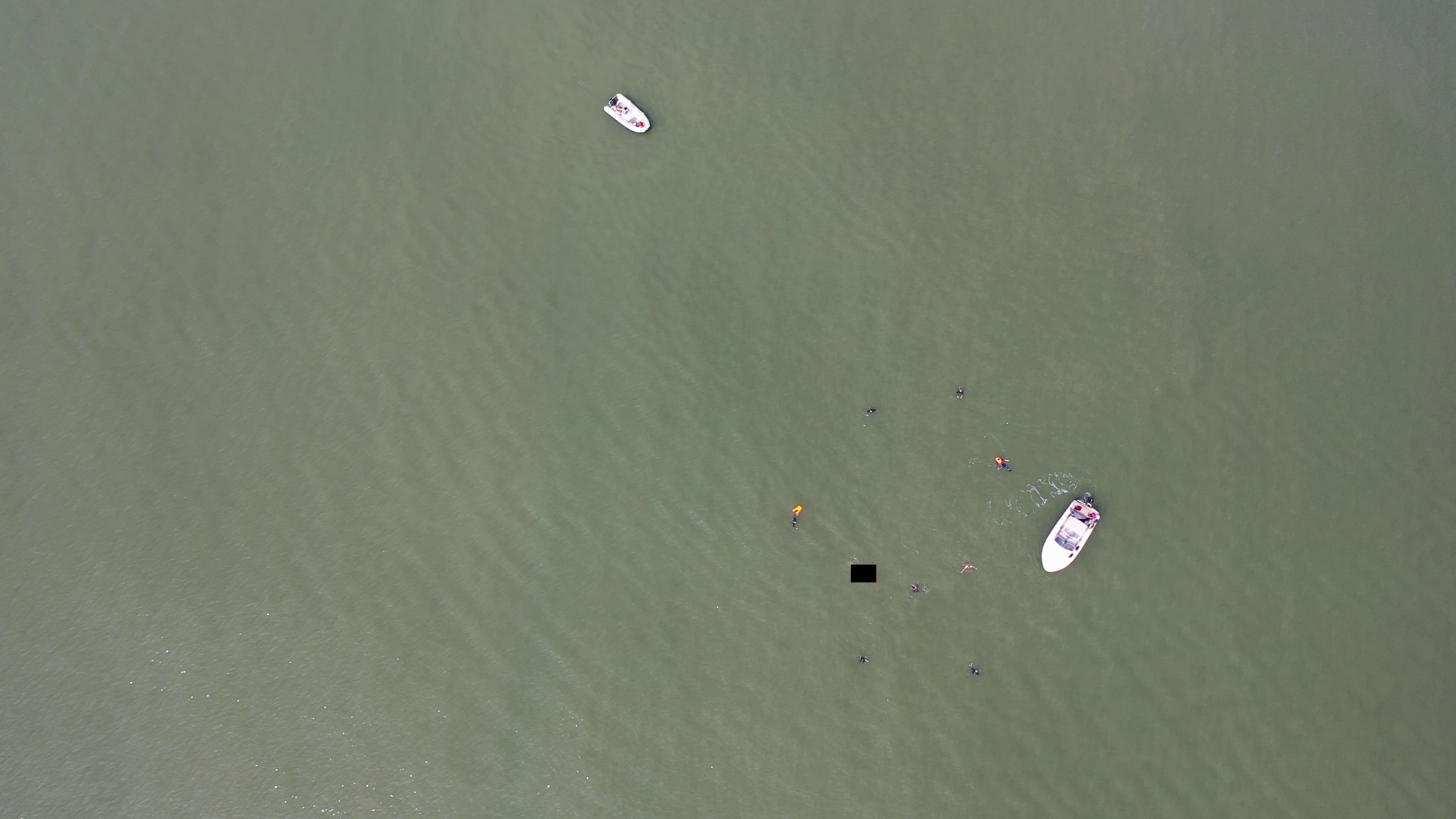}
    \includegraphics[width=0.46\textwidth]{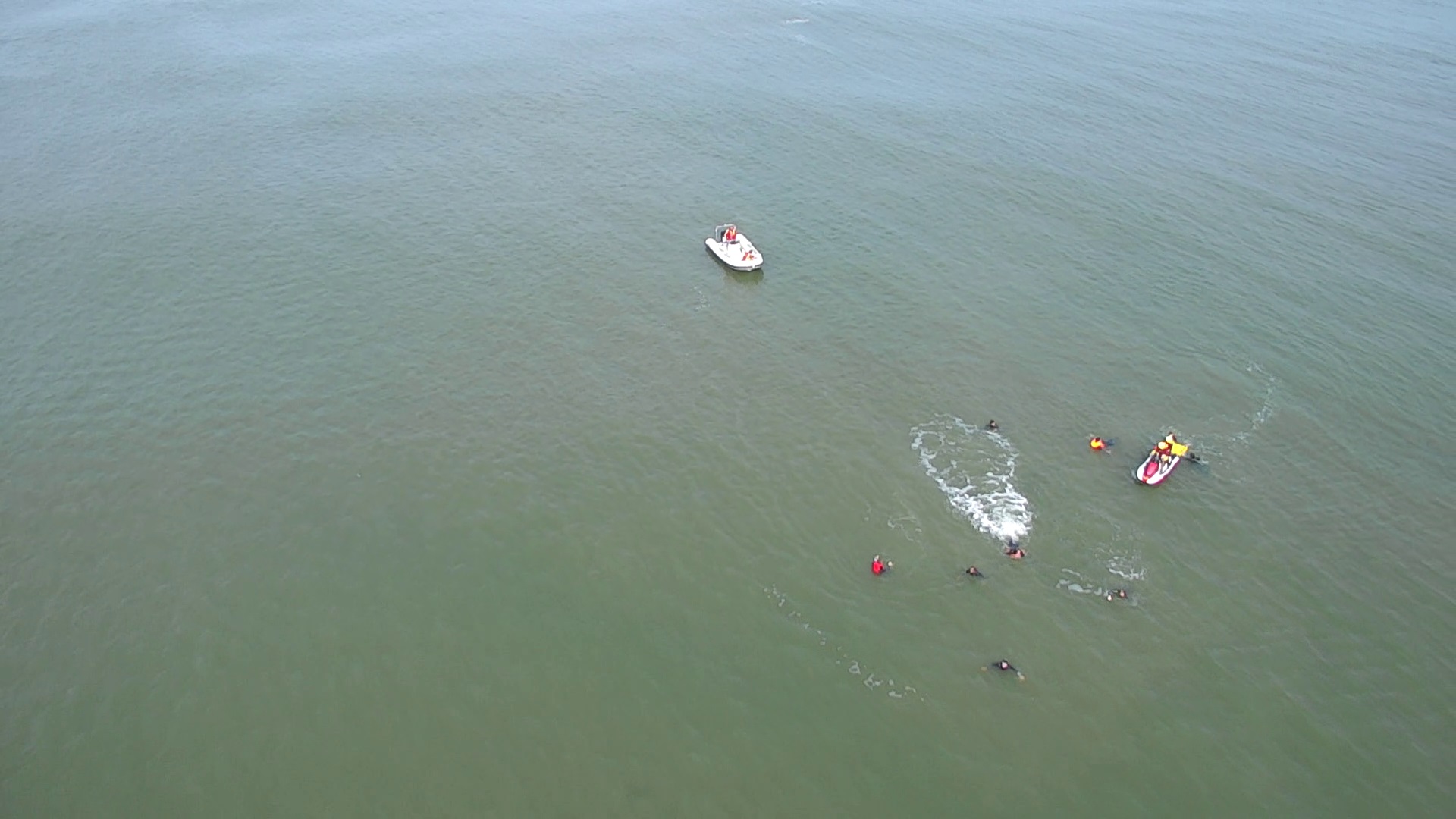}
\caption{Example images of newly generated images in the SDS ODv2 dataset. The black rectangle denotes an ignored region.}
\label{fig:OD_new_images}
\end{figure}

For the challenge of the workshop, we made a few changes from the original SeaDronesSee object detection benchmark. 
In addition to the already publicly available data, we collected further training data, which is included at the start of the challenge. 
In particular, we extend the object detection track of SeaDronesSee by roughly 9k newly captured images depicting the sea surface from the viewpoint of a UAV. See Figure \ref{fig:OD_new_images} for examples images.
The ground-truth bounding boxes are available and the evaluation protocol is based on the standard mean average precision. Owing to the application scenario, we also evaluate the class-agnostic performances, which resembles the use-case of detecting anything that is not water.

\subsection{Dataset}

%TODO maybe image with meta data visualization left and top?

\begin{figure}[t]
\centering
   \includegraphics[width=0.46\textwidth]{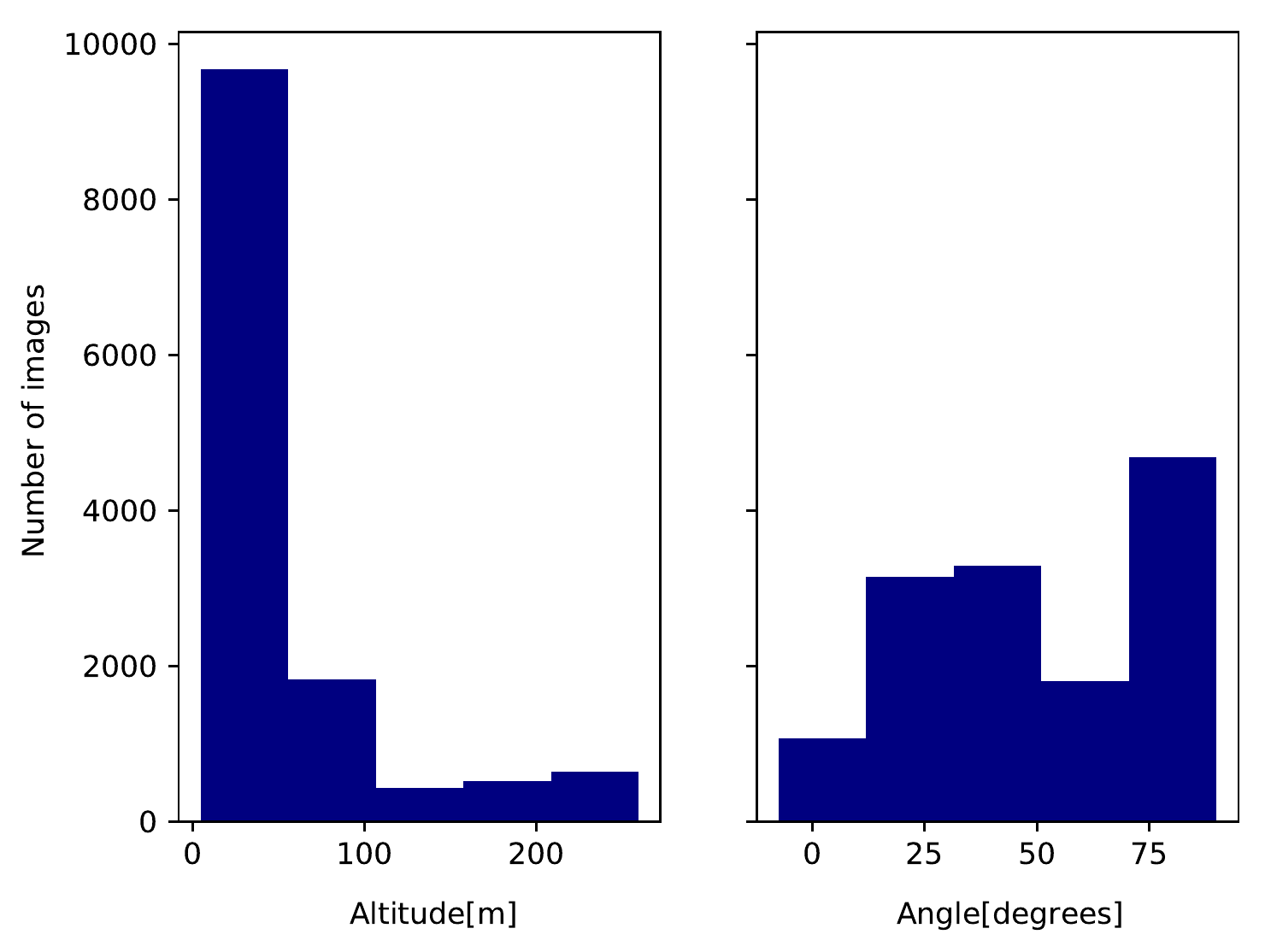}
\caption{Altitude and gimbal pitch angle distribution of images in SDS OD v2.}
\label{fig:altangleOD}
\end{figure}

\begin{figure}[t]
\centering
   \includegraphics[width=0.46\textwidth]{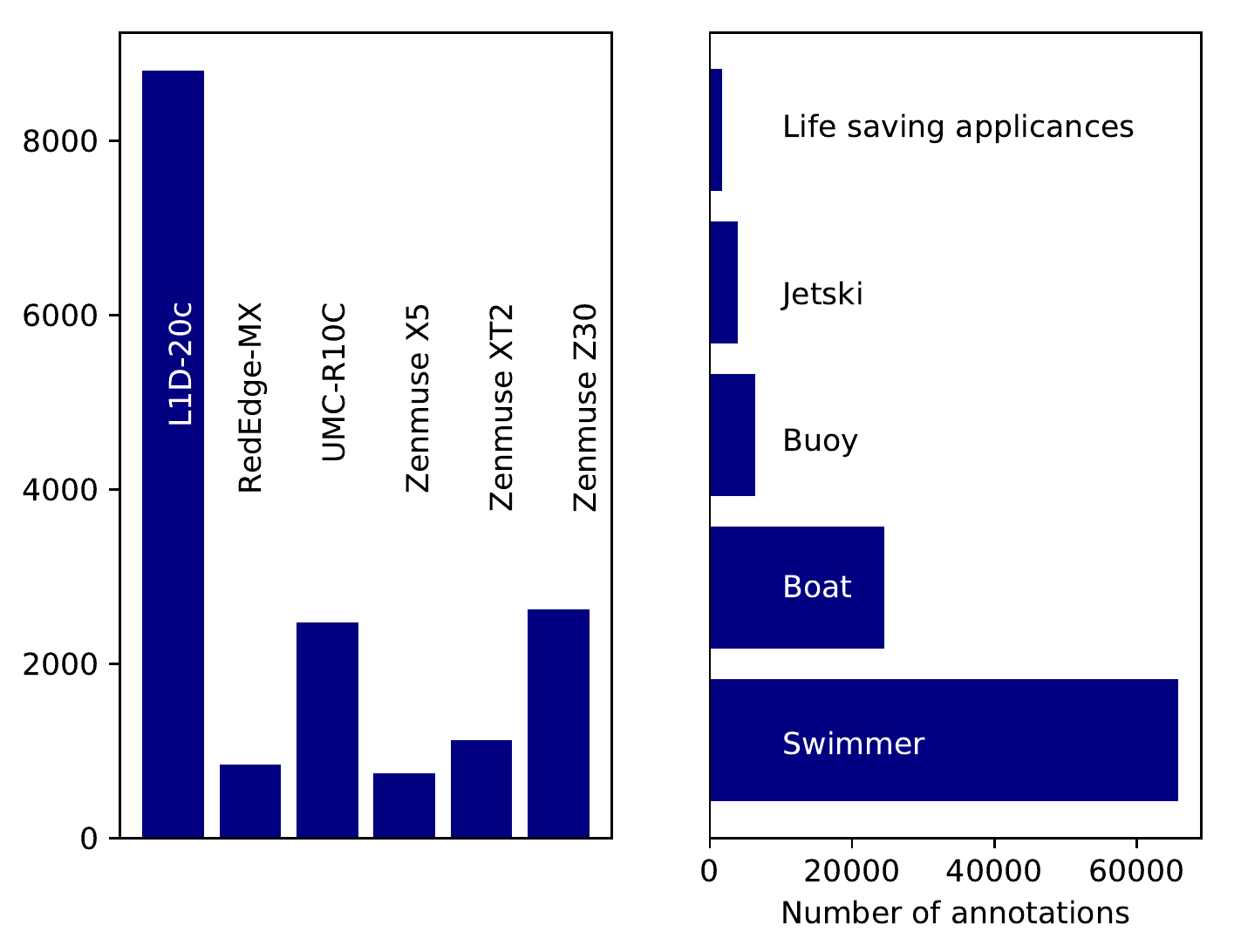}
\caption{Camera and class distribution in SDS ODv2.}
\label{fig:camdistOD}
\end{figure}

\begin{figure}[t]
\centering
   \includegraphics[width=0.46\textwidth]{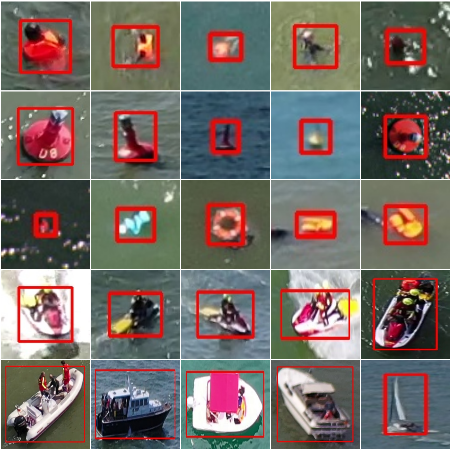}
\caption{Class instances in SDS ODv2. From top to bottom: swimmer, buoy, life vest/life belt, jetski, boat.}
\label{fig:instancesOD}
\end{figure}

\begin{table}[t]
\centering
\caption{Overview of used cameras, its properties  (all had RGB functionality) and the corresponding UAV to transport it for the SeaDronesSee Object Detection v2 dataset. Note that the zoom function was only used in a few images ($<$100), in the case of the multispectral and dual-thermal camera, only RGB channels were taken.}
\label{tab:camerasOD}
\vspace{-.2cm}
    \begin{tabular}{lrrr}
        \toprule
        Camera & Resolution & Type & UAV    \\
        \midrule
         L1D-20C &  3840x2160 & Vid & Mavic \\
         RedEdge-MX & 1280x960 & Multispectral & Trinity  \\
         UMC-R10C & 5456x3632 &  & Trinity\\
         Zenmuse X5 & 3840x2160 & Vid & M100 \\
         Zenmuse XT2 & 3840x2160 & Vid+Thermal & M210 \\
         Zenmuse Z30 & 1920x1080 & Vid+Zoom & M210 \\
        \bottomrule
    \end{tabular}
\end{table}

\begin{table}[t]
\centering
\caption{Meta data that comes with most images.}
\vspace{-.2cm}
    \label{table:meta_data}
    \begin{tabular}{lrrr}
        \toprule
        Data  & Unit & Min. value & Max.value  \\
        \midrule
        %$300,000$
        Time since start & ms & 0 & $\infty$ \\
        Date and Time & ISO 8601 & -- & -- \\
        Latitude & degrees & $-90$ & $+90$ \\
        Longitude & degrees & $-90$ & $+90$ \\
        Altitude & meters & $0$ & $\infty$ \\
        Gimbal pitch & degrees & $0$ & 90 \\
        UAV roll & degrees & $-90$ & $+90$ \\
        UAV pitch & degrees & $-90$ & $+90$ \\
        UAV yaw & degrees & $-180$ & $+180$ \\
        $x$-axis speed & m/s & $0$ & $\infty$ \\
        $y$-axis speed & m/s & $0$ & $\infty$ \\
        $z$-axis speed& m/s & $0$ & $\infty$\\
        \bottomrule
    \end{tabular}
\end{table}

The SeaDronesSee-Object Detection v2 (S-ODv2) dataset contains 14,227 RGB images (training: 8,930; validation: 1,547; testing: 3,750). The images are captured from various altitudes and viewing angles ranging from 5 to 260 meters and 0 to 90° degrees (gimbal pitch angle) while providing the respective meta information for altitude, viewing angle and other meta data for almost all frames. See Figure \ref{fig:altangleOD} for the altitude and viewing angle distribution. Most images come with additional meta data as depicted in Table \ref{table:meta_data}. 
Note that there are 2,830 images without any meta data labels and 686 images with only gimbal pitch angle labels. 
The images were captured with six different cameras as depicted in Table \ref{tab:camerasOD}. Note that we only used the RGB channels if more channels were available. Figure \ref{fig:camdistOD} shows the unbalanced camera distribution in the dataset. Each image is annotated with labels for the classes
\begin{tabularx}{\linewidth}{ @{} p{2.1cm} p{1.5cm} p{3.5cm} @{} }
  \begin{tightitemize}
    \item swimmer
    \item boat
  \end{tightitemize} &

  \begin{tightitemize}
    \item jetski
    \item buoy
  \end{tightitemize} &

  \begin{tightitemize}
    \item life saving appliance (life vest/belt).
  \end{tightitemize}
\end{tabularx}
See sample instances of these classes in Figure \ref{fig:instancesOD}.
Additionally, there is an ignore class. This region contains difficult to label or ambiguous objects. We blackened out these regions in the images already. Figure \ref{fig:camdistOD} shows the class distribution and the heavy class imbalance in the dataset.
Although the bounding box annotations for the test set are withheld, the meta data labels for the test set were provided.

\subsection{Evaluation Protocol}

We evaluate the predictions on the commonly used AP, AP50, AP75, AR1 and AR10 from the COCO evaluation protocol \cite{lin2014microsoft}. We provided the full evaluation protocol as part of our evaluation kits available on Github \cite{seadronessee}. For the first subtrack, we average the AP results over all classes. For the second subtrack, denoted binary object detection, we only have a single class called non-water. We further analyze the models using other metrics, such as TIDE \cite{bolya2020tide} and by leveraging the available meta data. The determining metric for winning will be AP. In case of a draw, AP50 counts.

\subsection{Submissions, Analysis and Trends}

\begin{table*}[tb]
\centering
\caption{Object Detection v2  submissions overview. For brevity, we denoted S=SeaDronesSee, O=Object Detection v2, all=(t)rain and (v)al set, IN=ImageNet, C=COCO, TTA=test-time augmentations. Augmentations only lists non-common augmentations and do not include techniques, such as resizing, color changes, cropping and more.}
\label{tab:od_submissions_overview}

\begin{tabular}{lrrrrrr}
\toprule
                Model name & Data & Type & Backbone & Module & Augmentations & Ref. \\
                \midrule
                Maritime-VSA (\ref{trod:Maritime-VSA}) & IN-22k, C, S-O$_t$ & Transf. & DB-Swin-S & Casc. R-CNN & VSA, TTA & \cite{zhang2022vsa}\\
                DetectoRS (\ref{trod:DetectoRS}) & C, S-O$_{all}$ & 2-stg.-CNN & ResNet-50 & Casc. R-CNN & TTA & \cite{qiao2021detectors} \\
                YOLOv7-Sea (\ref{trod:YOLOv7Sea}) & C, S-O$_t$ & 1-stg.-CNN & E-ELAN & SimAM & TTA, WBF & \cite{wang2022yolov7} \\
                DyHead (\ref{trod:DyHead}) & IN22k,C,S-O$_t$ & Transf. & Swin-L & Dynamic Head & TTA  & \cite{dai2021dynamic} \\
                YOLOv7-X (\ref{trod:YOLOv7X}) & C, S-O$_{t}$ & 1-stg.-CNN & YOLOv7-X & &  & \cite{wang2022yolov7} \\
                YOLO-CNS (\ref{trod:YOLOCNS}) & C, S-O$_t$ & Transf./CNN & Swin Transf. & CBAM, NAM & & \cite{yolov5github} \\
                YOLOv7-W6 (\ref{trod:YOLOv7W6}) & C, S-O$_t$ & 1-stg.CNN & YOLOv7-W6 & &  & \cite{wang2022yolov7} \\
                M10 (\ref{trod:M10}) & IN, S-O$_t$ & 1-stg.CNN & ResNeXt-101 & VarifocalNet & TTA  & \cite{zhang2021varifocalnet} \\
                YOLOv7-NYU (\ref{trod:YOLOv7NYU}) & C, S-O$_t$ & 1-stg.CNN & E-ELAN & Super-Res. & TTA, SAHI & \cite{wang2022yolov7} \\
                YOLOv7-FIT (\ref{trod:YOLOv7FIT}) & C, S-O$_t$ & 1-stg.CNN &  YOLOv7-E6 & &  & \cite{wang2022yolov7} \\
                DurObj (\ref{trod:DurObj}) & VisDrone, S-O$_t$ & 1-stg.CNN & ResNet-101 & TOOD &  & \cite{feng2021tood} \\
                APX (\ref{trod:APX}) & C, S-O$_t$ & 1-stg.CNN & Yolov7 & APX &  & \cite{wang2022yolov7} \\
                YOLOv7-TILE (\ref{trod:YOLOv7TILE}) & C, S-O$_t$ & 1-stg.CNN & YOLOv7 & & SAHI, TTA & \cite{wang2022yolov7} \\
                \bottomrule
\end{tabular}
\end{table*}

%TODO: mention yolov7-w6 for real-time use
%TODO fill missing fps values in object detection table

\begin{table*}[tb]
\centering
\caption{Final leaderboard for SeaDronesSee Object Detection v2.}
\label{tab:OD_finalstanding}
\vspace{-.2cm}
\begin{tabular}{lrlllllll}
\toprule
   Model name &  FPS &      Hardware &                          AP &                   AP$_{50}$ &                   AP$_{75}$ &                    AR$_{1}$ &                   AR$_{10}$ &               Binary$_{AP}$ \\
\midrule
 Maritime-VSA &    1 &          A100 &  \color{gold(metallic)}0.62 &  \color{gold(metallic)}0.91 &  \color{gold(metallic)}0.68 &  \color{gold(metallic)}0.48 &  \color{gold(metallic)}0.70 &  \color{gold(metallic)}0.56 \\
    DetectoRS &    1 &    Tesla V100 &  \color{lightslategray}0.60 &  \color{lightslategray}0.90 &  \color{lightslategray}0.66 &  \color{lightslategray}0.47 &          \color{bronze}0.67 &  \color{lightslategray}0.54 \\
   YOLOv7-Sea &    1 &    Tesla V100 &          \color{bronze}0.59 &  \color{gold(metallic)}0.91 &          \color{bronze}0.64 &          \color{bronze}0.46 &  \color{lightslategray}0.68 &  \color{lightslategray}0.54 \\
       DyHead &    1 &          A100 &                        0.57 &          \color{bronze}0.89 &                        0.62 &                        0.45 &  \color{lightslategray}0.68 &          \color{bronze}0.52 \\
     YOLOv7-X &   15 &      RTX 3090 &                        0.54 &                        0.85 &                        0.57 &                        0.44 &                        0.61 &                        0.50 \\
     Yolo-CNS &   60 &       TeslaP6 &                        0.53 &                        0.83 &                        0.56 &                        0.44 &                        0.62 &                        0.49 \\
    YOLOv7-W6 &   10 &      RTX 3090 &                        0.53 &                        0.84 &                        0.56 &                        0.44 &                        0.62 &                        0.49 \\
          M10 &    1 &       RTX3090 &                        0.53 &                        0.84 &                        0.55 &                        0.43 &                        0.60 &                        0.47 \\
   YOLOv7-NYU &   -1 &          2080 &                        0.52 &                        0.86 &                        0.54 &                        0.43 &                        0.60 &                        0.46 \\
   YOLOv7-FIT &    6 &       RTX3090 &                        0.52 &                        0.80 &                        0.55 &                        0.42 &                        0.58 &                        0.49 \\
       DurObj &    4 &      TITAN XP &                        0.50 &                        0.79 &                        0.51 &                        0.42 &                        0.58 &                        0.47 \\
          APX &   60 &      RTX 3050 &                        0.50 &                        0.83 &                        0.50 &                        0.41 &                        0.58 &                        0.45 \\
  YOLOv7-TILE &    3 &  Nvidia Titan &                        0.42 &                        0.71 &                        0.44 &                        0.36 &                        0.50 &                        0.44 \\
    YOLOv7-BL &   66 &      RTX 3080 &                        0.42 &                        0.72 &                        0.42 &                        0.36 &                        0.49 &                        0.41 \\
  FRCNN-RN-BL &   29 &   GTX 1080 Ti &                        0.24 &                        0.52 &                        0.20 &                        0.24 &                        0.32 &                        0.21 \\
\bottomrule
\end{tabular}
\end{table*}

We received 77 submissions from 18 different teams. 
We also provided two additional baselines, a YOLOv7 (\ref{trod:yolov7baseline}) and a Faster R-CNN with ResNet-18 backbone (\ref{trod:rcnnbaseline}). 
In our analysis, we will focus on the top 13 models that outperformed both these baselines.
None of the methods employed ensembles or were trained on any uncommon dataset.
Only some submissions used the SDS ODv2 validation set for training.
%TODO put refs to tech reports in parantheses
Three of the submitted models were transformer-based, which originally were especially hard to tune for small object detection, but was recently found popular also in the aerial object detection domain \cite{cao2021visdrone}. 
More precisely, the winner of this challenge, Maritime-VSA (\ref{trod:Maritime-VSA}), the 4$^{\text{th}}$ place, DyHead (\ref{trod:DyHead}), and the 6$^{\text{th}}$ place (\ref{trod:YOLOCNS}) rely either entirely or partly on transformer-based blocks. Maritime-VSA showcase their recently published varied-size window attention, which is suitable for processing large image resolutions compared to more traditional transformer architectures. 
In conjunction with the popular Cascade R-CNN as a detection head and test-time augmentations, they obtained a significant lead. DyHead leverage the recent so-called dynamic heads to unify the object detection heads for localization and classification via attention mechanisms \cite{dai2021dynamic}. 
Test-time augmentations and large image resolutions were employed. 
The method rightfully mentions the problems with annotation errors, which will be analyzed below. 

\begin{figure}[t]
\centering
   \includegraphics[width=0.46\textwidth]{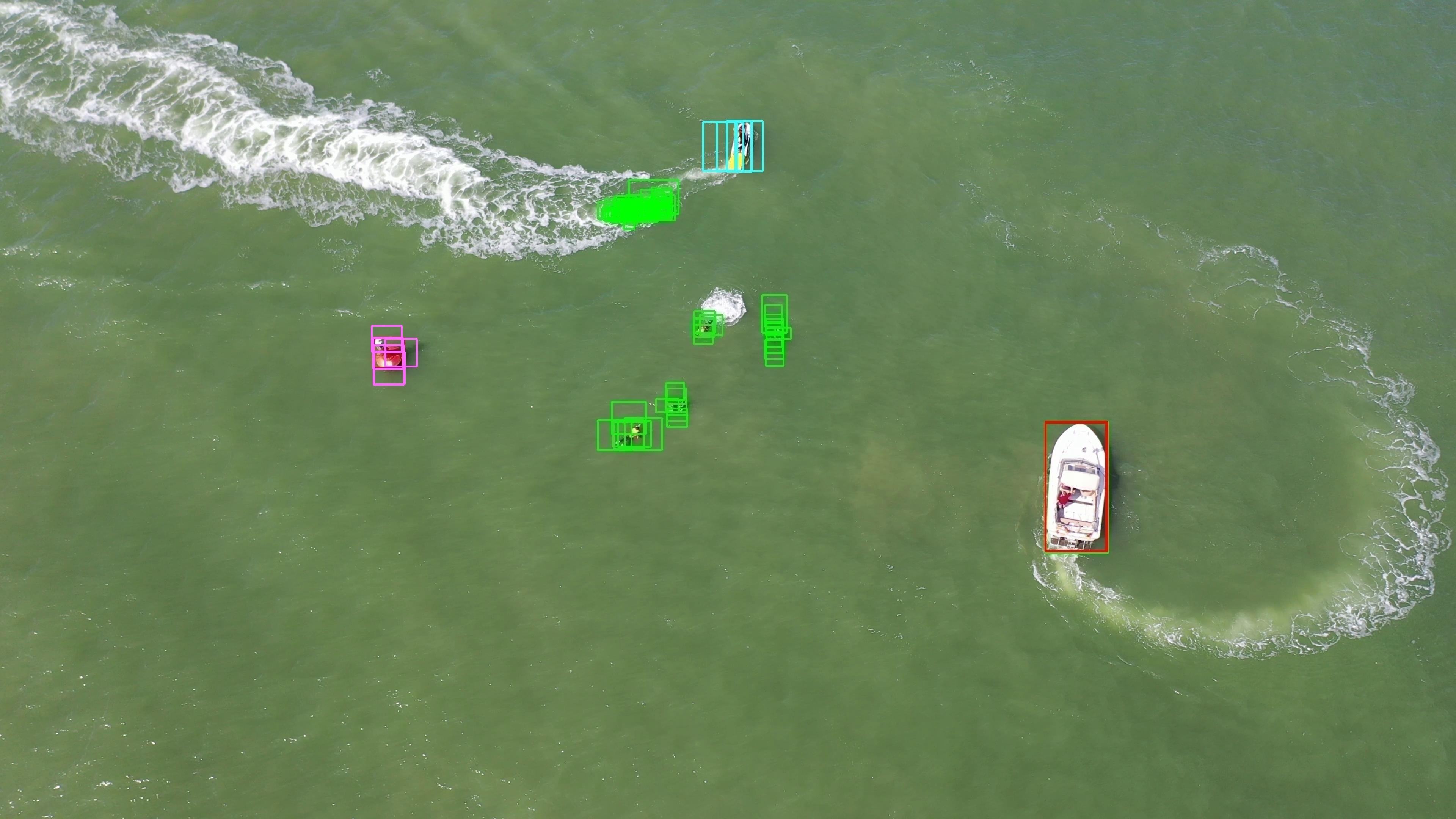}
   \includegraphics[width=0.46\textwidth]{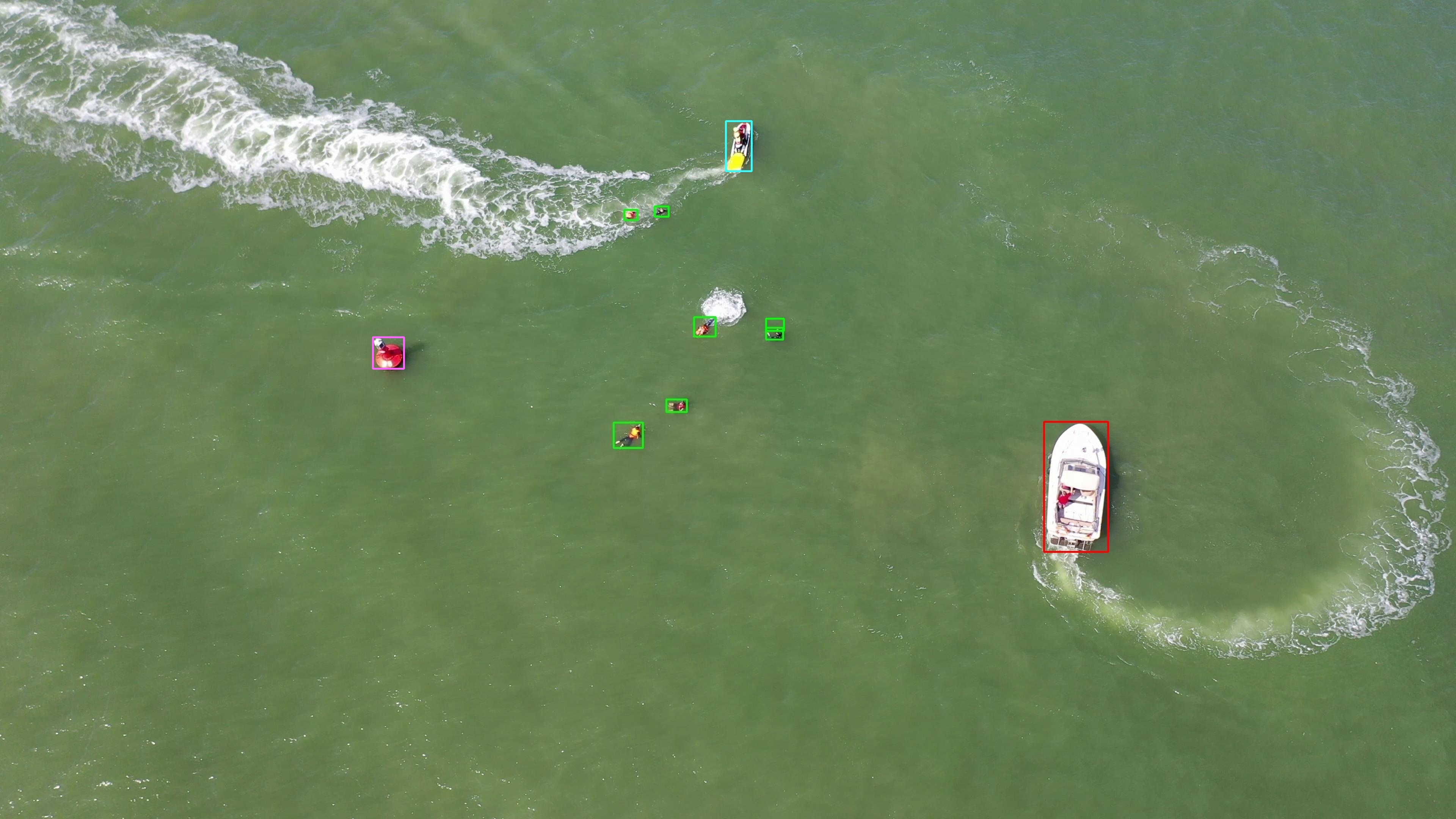}
\caption{Example predictions from Maritime-VSA (top) and DetectoRS (bottom). Note that we did not filter based on the confidence score, which is why the first method has many predictions for each object. The confidence score for most of them is very low, which is why the AP won't suffer from these.
}
\label{fig:OD_sample_imgs}
\end{figure}

The remaining models are different types of CNNs. 
The 2$^{\text{nd}}$ place, DetectoRS (\ref{trod:DetectoRS}), base their submission on Cascade R-CNN (\cite{cai2018cascade}), which is well known for its performance in small object detection (see e.g.~performance on VisDrone workshop \cite{cao2021visdrone}). 
A likely significant addition is that they employed large resolutions and multi-scale testing. 
Several other methods are based on a YOLO-variant, most prominently the current YOLOv7 \cite{wang2022yolov7} architecture. 
In fact, the 3$^{\text{rd}}$ (\ref{trod:YOLOv7Sea},\cite{Zhao_2023_WACV}), 5$^{\text{th}}$ (\ref{trod:YOLOv7X}), 7$^{\text{th}}$ (\ref{trod:YOLOv7W6}), 9$^{\text{th}}$ (\ref{trod:YOLOv7NYU}), 10$^{\text{th}}$ (\ref{trod:YOLOv7FIT}), 12$^{\text{th}}$ (\ref{trod:APX}) and 13$^{\text{th}}$ (\ref{trod:YOLOv7TILE}) places all base their submissions on YOLOv7. 
Many YOLOv7 submissions either adapted the architecture to include an attention module (\ref{trod:YOLOv7Sea}) or tuned hyperparameters, such as considerably increasing the image size (\ref{trod:YOLOv7X}, \ref{trod:YOLOv7W6}, \ref{trod:YOLOv7FIT}),  or included augmentations, such as random cropping (\ref{trod:YOLOv7NYU}), mosaicing (\ref{trod:YOLOv7NYU}) or color changes (\ref{trod:YOLOv7W6}) just to name a few. \ref{trod:YOLOv7NYU} has an interesting take by applying a super-resolution network before applying the object detector.
Authors in \ref{trod:APX} take a more targeted approach to the maritime domain by clustering the pixel colors via Kmeans, such that mostly blue-green appearing water pixels can better be distinguished by the downstream YOLOv7 detector.

The remaining methods use more specific architectures, such as VarifocalNet \cite{zhang2021varifocalnet} (\ref{trod:M10}), a single-stage object detector, which itself is based on FCOS \cite{tian2019fcos}. 
Further augmentations, such as tiling (also multi-scale) improved the performance significantly.  
Authors in \ref{trod:DurObj} base their submission on a one-stage detector proposing to better align the outputs from the two subbranches, classification and localization \cite{feng2021tood}. See Table \ref{tab:od_submissions_overview} for an overview of the submitted methods.

Table \ref{tab:OD_finalstanding} shows the final standing of this challenge track. 
Notably, the performance of the top models is above 90 AP$_{50}$. 
Owing to the aerial nature and potentially sub-optimal label accuracy (e.g. shifted labels), the averaged AP is far lower, which is also reflected in the lower $AR_1$ and $AR_{10}$ scores. The binary AP, which measures the foreground vs. background performance, is slightly worse for almost all models which is likely caused by the class imbalance with the majority of the instances being swimmers, which is generally a hard class to predict (see Table \ref{tab:OD_final_standing_classes}). 

\begin{figure}[t]
\centering
   \includegraphics[width=0.45\textwidth]{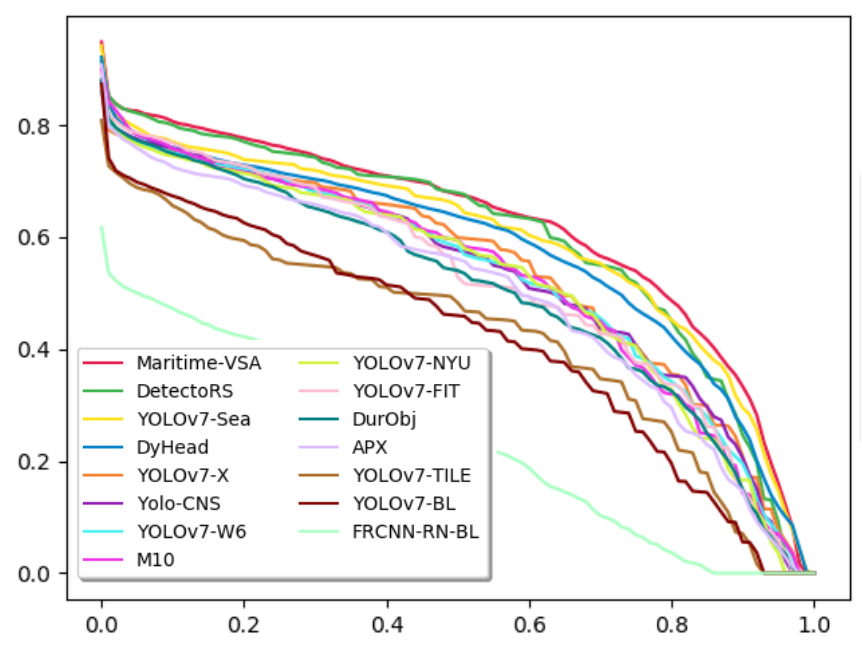}
\caption{Precision (x-axis)-recall (y-axis)-curve, for submitted methods.
}
\label{fig:prec_recall_odv2}
\end{figure}

Generally, the classes swimmers and life saving appliances are believed to be the hardest classes as their appearance vary the most and they are the smallest (and thus hardest to predict) objects (see also Figure \ref{fig:instancesOD}). Furthermore, these two classes are harder to distinguish and there are only few instances of life saving appliances. Furthermore, the methods' ranks in performance across different precision levels are consistent as can be seen from Figure \ref{fig:prec_recall_odv2}, i.e. every model is more or less better or worse than any other model for all precision scores consistently. 

\begin{figure}[t]
\centering
   \includegraphics[width=0.2\textwidth]{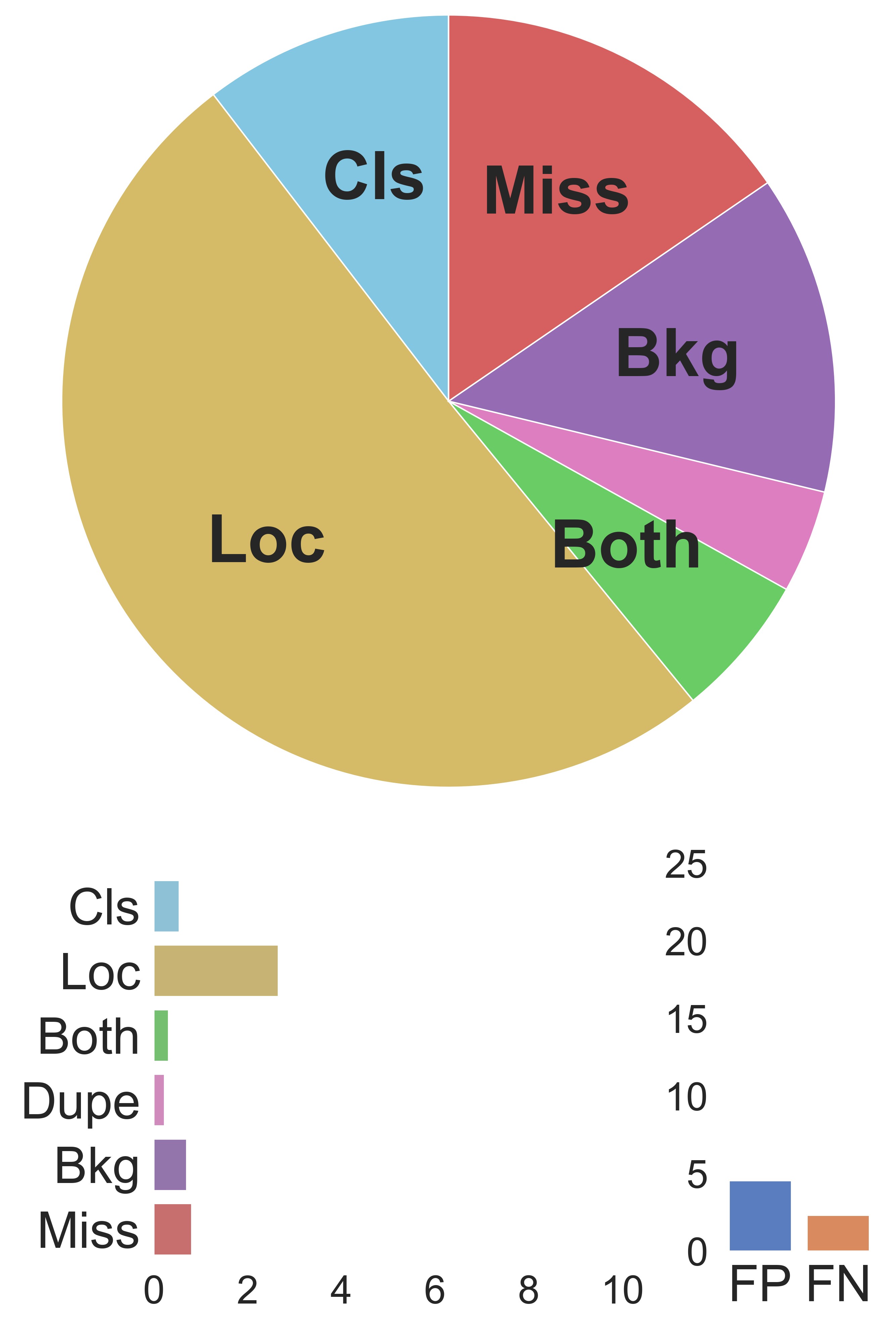}
   \includegraphics[width=0.2\textwidth]{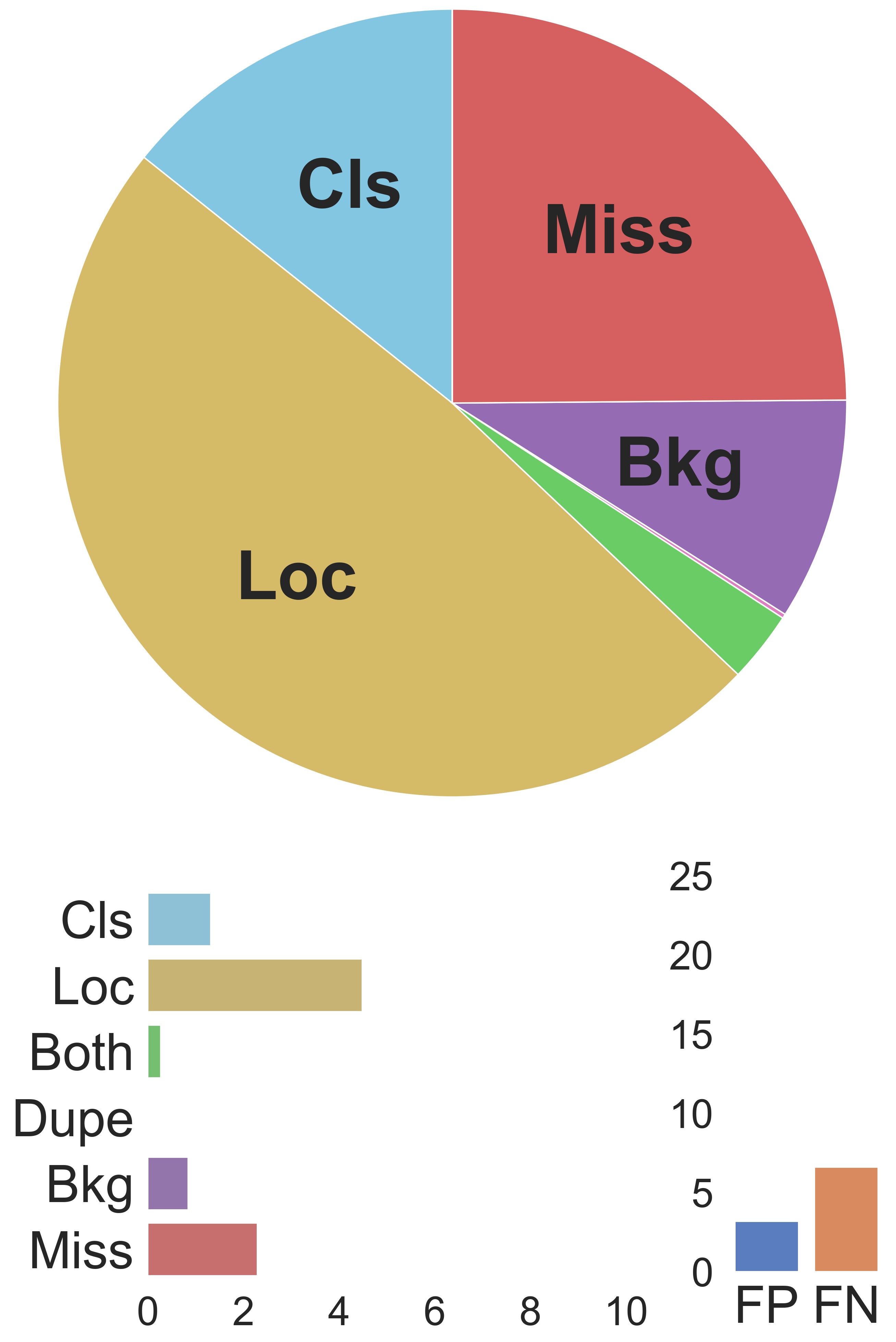}
\caption{TIDE evaluations for Maritime-VSA (left) and DetectoRS (right).}
\label{fig:tide_best_two_odv2}
\end{figure}

A closer analysis on the type of error can be seen from the TIDE plots in Figure \ref{fig:tide_best_two_odv2}. There, we plot the different error types of the two best performing submissions, Maritme-VSA and DetectoRS. Both models behave similarly in their error type influence distribution, e.g. most of the errors come from localization errors (roughly 50\%). Background errors (falsely predicting background to be any class instance) are a similarly often cause of errors as missing to detect objects in the case of Maritime-VSA. However, DetectoRS takes a different trade-off and mostly only misses objects as opposed to detecting background as foreground objects. Note, however, that the overall magnitude of errors is lower for Maritime-VSA for both types of errors (bottom bar charts). The less common duplicate detections errors only play a role in Maritime-VSA, which aligns with the qualitative prediction example in Figure \ref{fig:OD_sample_imgs}. Note, however, that these duplicate predictions have low confidence and hence do not matter too much in the overall AP calculation.

\begin{figure}[tb]
\centering
   \includegraphics[width=0.46\textwidth]{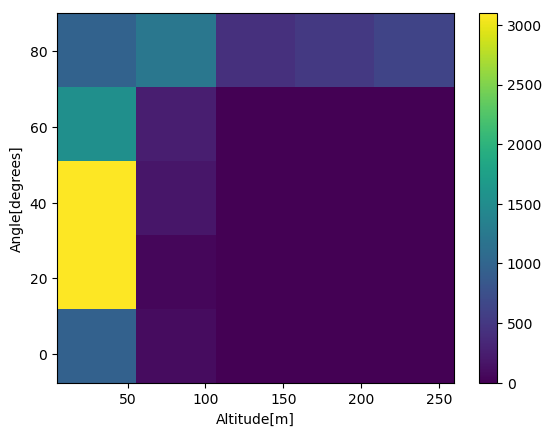}
\caption{Altitude angle distribution of images in SDS ODv2.
}
\label{fig:2dhisto}
\end{figure}

Table \ref{tab:OD_final_standing_meta} shows the AP values broken down by different meta data configuration intervals. Again, the models perform mostly consistently across different domains. Generally significantly visible, the performance for acute angles is low across all models. While this may simply be the cause of having fewer images in that domain (compare to Fig. \ref{fig:altangleOD}), these images often contain very small objects in the distant horizon. Furthermore, in the case of swimmers, these are hardly visible as only their body parts above the water are visible (see e.g. the first swimmer of Figure \ref{fig:instancesOD} compared to the third one). Surprisingly, the performance on high altitudes is the highest. This could be the cause of consistent viewpoints, as images from high altitude exhibit viewpoints almost always of close to 90$^\circ$ (looking downwards; see Figure \ref{fig:2dhisto}. The performances broken down by different cameras is not conclusive. The performance for the M210 UAV is very low, which can only be hypothesized to be partly attributed to the M210 UAV carrying the lower resolution Zenmuse Z30 (see Tab. \ref{tab:camerasOD}), although there exist many images (compare to Fig. \ref{fig:camdistOD}). The high performance for the trinity drone is again believed to be caused by the consistent 90$^\circ$ facing downwards viewpoint as this UAV only has facing downwards cameras.

\begin{figure}[t]
\centering
   \includegraphics[width=0.46\textwidth]{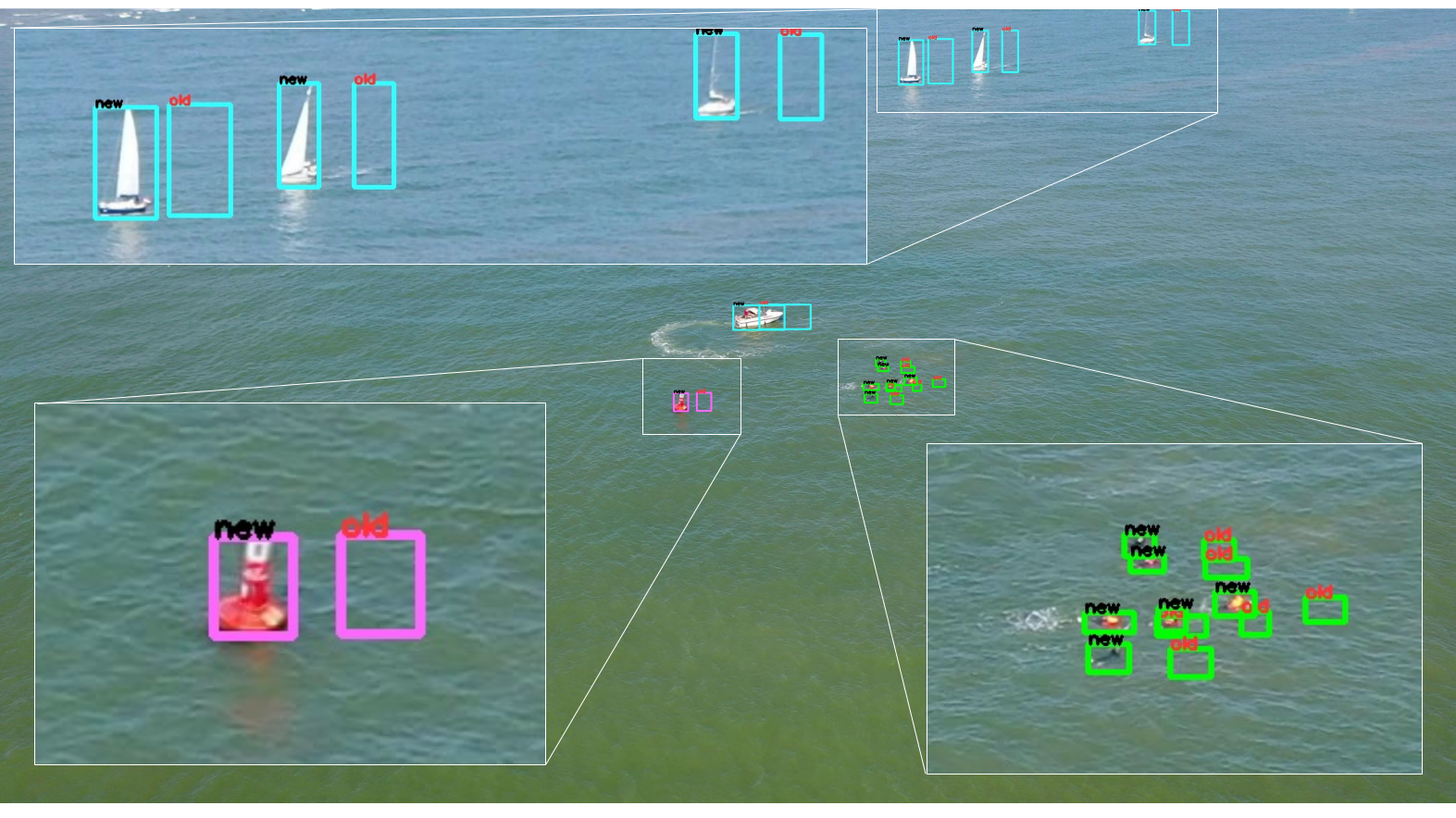}
   \includegraphics[width=0.46\textwidth]{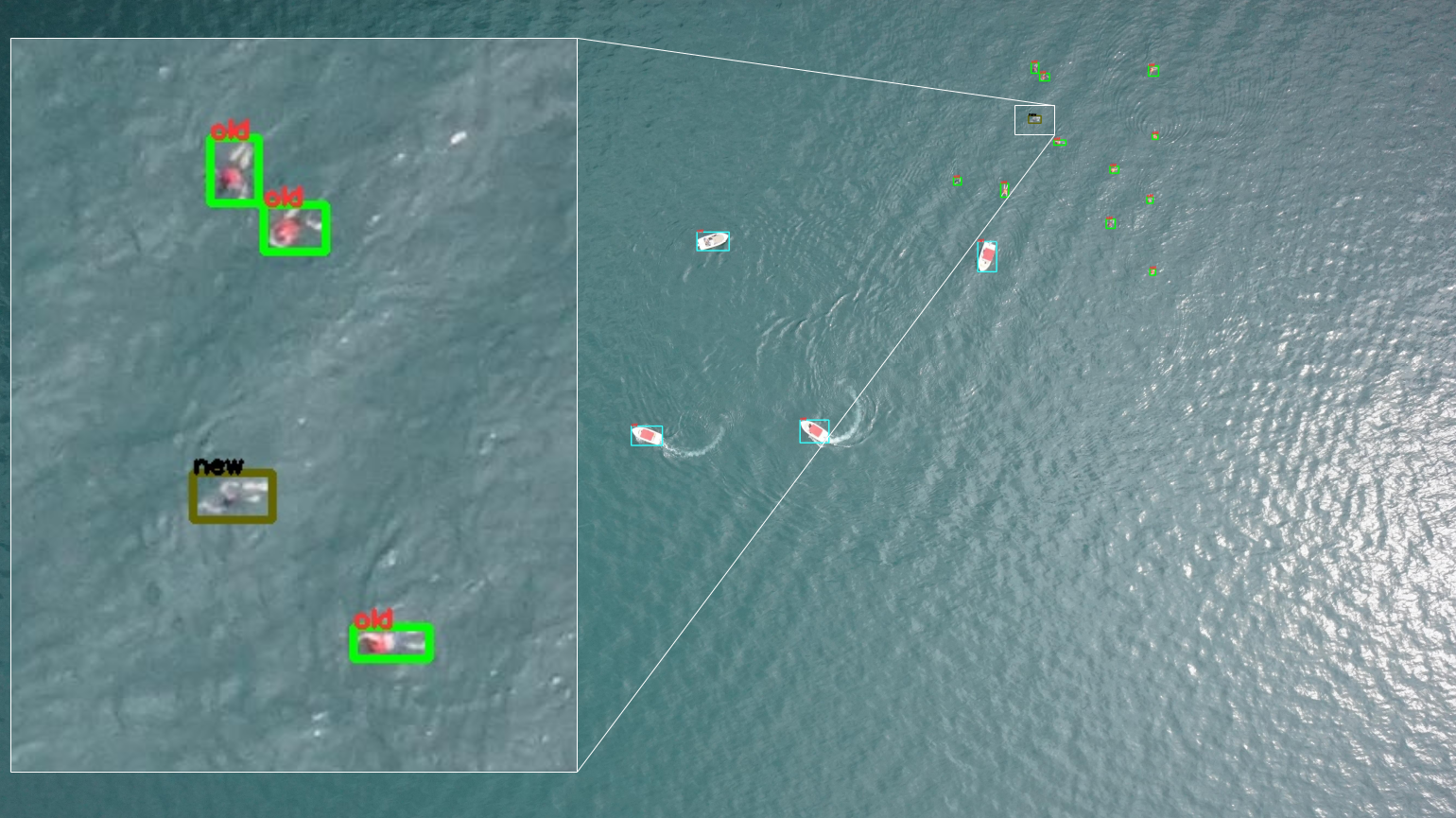}
\caption{Label errors revealed after another iteration of manual annotation. Top: Displaced labels, Bottom: missing labels (red font: old, black font: new).}
\label{fig:OD_label_errors}
\end{figure}

The dataset contains a fair amount of label errors, which we found upon reiterating a whole manual annotation pass over the dataset. Table \ref{table:OD_label_error_statistic} shows the number of found label errors. 
See examples of label errors in Figure \ref{fig:OD_label_errors}. 
Displaced label errors come from the used annotation tool Darklabel's tracking functionality\footnote{\url{github.com/darkpgmr/DarkLabel}, accessed: Nov 2022.}, which causes a drag in the bounding box labels in scenes where there is a lot of camera or UAV movement. 
Missing labels mostly occur in static images where there was no underlying video that aided the human annotators in finding objects to label. Table \ref{tab:OD_final_standing_corrected_dataset} shows the performances of the individual submissions on the cleaned/corrected dataset. It shows that the performances indeed improve across all models but the overall order stays almost the same.

\subsection{Discussion and Challenge Winners}

The challenge results have shown that transformer architectures start to gain traction in the aerial domain as well, while CNN architectures are still the standard choice for such tasks. The easy-to-use and yet strong one-stage detector YOLOv7 is a very popular choice. As is common for these kind of challenges (compare to VisDrone \cite{visdroneworkshop}), test-time augmentations are applied and significantly boost the performance at the cost of slower run times. Furthermore, using large resolutions is one of the keys to obtaining high accuracies, be it by means of architecturally supporting large resolutions or by targeted augmentations, such as cropping. 

The observation above is exemplified by the winner trio: The first place from The University of Sydney, Maritime-VSA (\ref{trod:Maritime-VSA}), employed transformers, the second place from Fraunhofer IOSB, DetectoRS (\ref{trod:DetectoRS}), leveraged the popular two-stage detector Cascade R-CNN, and the third place from Beijing University of Posts and Telecommunications, YOLOv7-Sea (\ref{trod:YOLOv7Sea}), built upon the current YOLOv7 detector.
%TODO: check at other uav/sv sections whether they included a winners discussion

Furthermore, most submitted object detectors run far from real-time. While \ref{trod:YOLOv7W6} made experiments with a real-time capable YOLOv7-tiny, they obtained detrimental accuracies. Furthermore, special consideration should be given to the used hardware in that case since in this challenge, participants mostly relied on high-end GPUs, such as V100s.

Therefore, research in this domain needs to consider runtime constraints imposed in real applications of these detectors. In future iterations of MaCVi, this would need to be a focus.

\begin{table}[t]
\centering
\caption{Class APs for (Sw)immer, (Bo)at, (Je)tski, (L)ife (s)aving appliances and (Bu)oy.
}
\label{tab:OD_final_standing_classes}
\vspace{-.2cm}
\begin{tabular}{llllll}
\toprule
   Model name &                          Sw &                          Bo &                          Je &                          Ls &                          Bu \\
\midrule
 Maritime-VSA &  \color{gold(metallic)}0.44 &  \color{gold(metallic)}0.80 &  \color{gold(metallic)}0.64 &  \color{gold(metallic)}0.50 &  \color{gold(metallic)}0.69 \\
    DetectoRS &  \color{lightslategray}0.43 &  \color{lightslategray}0.78 &          \color{bronze}0.62 &  \color{lightslategray}0.49 &          \color{bronze}0.66 \\
   YOLOv7-Sea &  \color{lightslategray}0.43 &          \color{bronze}0.77 &                        0.61 &          \color{bronze}0.47 &  \color{lightslategray}0.67 \\
       DyHead &          \color{bronze}0.41 &  \color{lightslategray}0.78 &  \color{lightslategray}0.63 &                        0.39 &                        0.64 \\
     YOLOv7-X &                        0.38 &                        0.74 &                        0.59 &                        0.34 &                        0.64 \\
     Yolo-CNS &                        0.37 &                        0.73 &                        0.58 &                        0.32 &                        0.64 \\
    YOLOv7-W6 &                        0.36 &                        0.74 &                        0.58 &                        0.33 &                        0.61 \\
          M10 &                        0.34 &                        0.75 &                        0.58 &                        0.34 &                        0.62 \\
   YOLOv7-NYU &                        0.35 &                        0.70 &                        0.56 &                        0.39 &                        0.59 \\
   YOLOv7-FIT &                        0.37 &                        0.74 &                        0.59 &                        0.25 &                        0.63 \\
       DurObj &                        0.36 &                        0.74 &                        0.58 &                        0.21 &                        0.62 \\
          APX &                        0.33 &                        0.70 &                        0.55 &                        0.30 &                        0.61 \\
  YOLOv7-TILE &                        0.33 &                        0.66 &                        0.50 &                        0.08 &                        0.55 \\
    YOLOv7-BL &                        0.30 &                        0.64 &                        0.50 &                        0.15 &                        0.50 \\
  FRCNN-RN-BL &                        0.13 &                        0.42 &                        0.35 &                        0.00 &                        0.32 \\
\bottomrule
\end{tabular}
\end{table}

\begin{table*}[t]
\centering
\caption{AP results for subsets, divided by altitude ((L)ow, (M)edium, H(igh)), gimbal pitch ((A)cute, (A)cute to (R)ight, (R)ight), and camera ((Mav)ic, M210 and (Tri)nity. We divide the 'Altitudes' and 'Angles' into three equidistant intervals.}
\label{tab:OD_final_standing_meta}
\vspace{-.2cm}
\begin{tabular}{llllllllll}
\toprule
   Model name &                    AP$_{L}$ &                    AP$_{M}$ &                    AP$_{H}$ &                    AP$_{A}$ &                   AP$_{AR}$ &                    AP$_{R}$ &                  AP$_{Mav}$ &                 AP$_{M210}$ &                  AP$_{Tri}$ \\
\midrule
 Maritime-VSA &  \color{gold(metallic)}0.62 &  \color{gold(metallic)}0.57 &  \color{lightslategray}0.68 &  \color{gold(metallic)}0.23 &  \color{gold(metallic)}0.65 &  \color{gold(metallic)}0.64 &  \color{gold(metallic)}0.61 &  \color{gold(metallic)}0.18 &  \color{gold(metallic)}0.71 \\
    DetectoRS &  \color{lightslategray}0.61 &  \color{lightslategray}0.55 &  \color{gold(metallic)}0.70 &  \color{lightslategray}0.22 &  \color{lightslategray}0.63 &  \color{lightslategray}0.63 &  \color{lightslategray}0.59 &  \color{lightslategray}0.17 &  \color{lightslategray}0.69 \\
   YOLOv7-Sea &  \color{lightslategray}0.61 &          \color{bronze}0.53 &          \color{bronze}0.67 &          \color{bronze}0.21 &  \color{lightslategray}0.63 &          \color{bronze}0.62 &  \color{lightslategray}0.59 &          \color{bronze}0.16 &          \color{bronze}0.66 \\
       DyHead &          \color{bronze}0.59 &                        0.49 &                        0.63 &                        0.18 &          \color{bronze}0.62 &                        0.60 &          \color{bronze}0.57 &  \color{lightslategray}0.17 &                        0.64 \\
     YOLOv7-X &                        0.56 &                        0.48 &                        0.56 &                        0.18 &                        0.59 &                        0.55 &                        0.54 &          \color{bronze}0.16 &                        0.59 \\
     Yolo-CNS &                        0.56 &                        0.42 &                        0.64 &                        0.18 &                        0.58 &                        0.55 &                        0.53 &                        0.13 &                        0.61 \\
    YOLOv7-W6 &                        0.55 &                        0.47 &                        0.62 &                        0.17 &                        0.58 &                        0.58 &                        0.53 &                        0.12 &                        0.62 \\
          M10 &                        0.55 &                        0.41 &          \color{bronze}0.67 &                        0.14 &                        0.57 &                        0.60 &                        0.52 &                        0.13 &                        0.61 \\
   YOLOv7-NYU &                        0.53 &                        0.49 &                        0.63 &                        0.13 &                        0.57 &                        0.61 &                        0.51 &                        0.14 &                        0.55 \\
   YOLOv7-FIT &                        0.53 &                        0.42 &          \color{bronze}0.67 &                        0.17 &                        0.56 &                        0.58 &                        0.51 &                        0.13 &                        0.63 \\
       DurObj &                        0.52 &                        0.40 &                        0.64 &                        0.13 &                        0.56 &                        0.56 &                        0.49 &                        0.14 &                        0.58 \\
          APX &                        0.54 &                        0.39 &                        0.54 &                        0.13 &                        0.56 &                        0.53 &                        0.50 &                        0.11 &                        0.54 \\
  YOLOv7-TILE &                        0.45 &                        0.36 &                        0.30 &                        0.13 &                        0.49 &                        0.35 &                        0.43 &                        0.10 &                        0.47 \\
    YOLOv7-BL &                        0.44 &                        0.34 &                        0.42 &                        0.06 &                        0.50 &                        0.45 &                        0.42 &                        0.12 &                        0.43 \\
  FRCNN-RN-BL &                        0.26 &                        0.23 &                        0.26 &                       0.00 &                        0.33 &                        0.27 &                        0.25 &                        0.03 &                        0.21 \\
\bottomrule
\end{tabular}
\end{table*}

\begin{table}[tb]
\centering
\caption{Annotation error statistics.}
\label{table:OD_label_error_statistic}
\vspace{-.2cm}
    \begin{tabular}{lrrrr}
        \toprule
          & Train & Val & Test   \\
        \midrule
        \# missed boxes & 404 & 81 & 193 \\
        \# displaced boxes & 257 & 118 & 240 \\
        \bottomrule
    \end{tabular}
\end{table}

\begin{table}[tb]
\centering
\caption{Prediction results on datasets with corrected annotations.
}
\label{tab:OD_final_standing_corrected_dataset}
\begin{tabular}{lrr}
\toprule
   Model name &                          AP &                   AP$_{50}$ \\
\midrule
 Maritime-VSA &  \color{gold(metallic)}0.64 &  \color{gold(metallic)}0.95 \\
    DetectoRS &  \color{lightslategray}0.63 &  \color{lightslategray}0.93 \\
   YOLOv7-Sea &          \color{bronze}0.62 &  \color{gold(metallic)}0.95 \\
       DyHead &                        0.60 &          \color{bronze}0.92 \\
     YOLOv7-X &                        0.56 &                        0.89 \\
     Yolo-CNS &                        0.56 &                        0.87 \\
    YOLOv7-W6 &                        0.55 &                        0.87 \\
          M10 &                        0.56 &                        0.88 \\
   YOLOv7-NYU &                        0.54 &                        0.89 \\
   YOLOv7-FIT &                        0.54 &                        0.83 \\
       DurObj &                        0.52 &                        0.82 \\
          APX &                        0.52 &                        0.87 \\
  YOLOv7-TILE &                        0.45 &                        0.74 \\
    YOLOv7-BL &                        0.44 &                        0.76 \\
  FRCNN-RN-BL &                        0.26 &                        0.55 \\
\bottomrule
\end{tabular}
\end{table}

\section{UAV-based Object Tracking Challenge}
\label{sec:uavobjecttrackingchallenge}

Part of the SeaDronesSee benchmark was the Multi-Object Tracking track. This track focuses on tracking objects in water which are of interest in SaR scenarios, while it could also be leveraged for surveillance.
In SaR scenarios, it might be of interest to track the detection and position of people or boats over time, so that the found subjects are easily distinguishable. 
However, tracking small, partly occluded subjects, which change their appearance based on their movement and occlusion level due to water, is non-trivial. Gimbal movement and altitude change cause objects to move quickly within the video frames. 
For these reasons, we hosted the first SeaDronesSee-MOT challenge track, which will be discussed in the following.

%TODO not anymore: insert clip and 2d-3d visualization

\subsection{Dataset}

The SeaDronesSee-MOT dataset consists of 21 clips in the train set, 17 clips in the validation set and 19 clips in the test set with a total of 54,105 frames and 403,192 annotated instances. Every frame is annotated with the ground-truth bounding boxes along unique ids for the following classes:
\begin{multicols}{2}
\begin{tightitemize}
\item swimmer
\item floater
\item life jacket
\end{tightitemize}
\columnbreak
\begin{tightitemize}
\item swimmer on boat
\item floater on boat
\item boat
\end{tightitemize}
\end{multicols}
Floater denotes a swimmer wearing a life jacket. Following \cite{varga2022seadronessee}, for the SeaDronesSee-MOT challenge track, we restrict the task as follows. We only require the objects boats, swimmer and floater to be tracked in a one-class setting, where we do not distinguish between different classes. We note that this is a \emph{short-term} tracking task~\cite{KristanPAMI2016}, i.e. objects that disappear from the scene need not be tracked anymore. Each frame comes with precise meta data labels regarding altitude, angles of the UAV and the gimbal, GPS, and more.

\subsection{Evaluation Protocol}

We evaluate the submissions by using the following metrics: HOTA, MOTA, IDF1, MOTP, MT, ML, FP, FN, Recall, Precision, ID Switches, Frag \cite{luiten2021hota,leal2015motchallenge}. The determining metric for winning is HOTA. In case of a tie, MOTA is the tiebreaker.

Furthermore, we require every participant to submit information on the computational runtime of their method measured in frames per second wall-clock time along their used hardware.

\subsection{Submissions, Analysis and Trends}

\begin{table*}[t]
\centering
\caption{Multi-Object Tracking  submissions overview. For brevity, we denoted d=detector, t=tracker, S=SeaDronesSee, O=Object Detection v2, all=train and val set. Furthermore, "+"=adding and "-"=removing" the respective module.}
\label{tab:mot_submissions_overview}
\vspace{-.2cm}
\begin{tabular}{lrrrrrr}
\toprule
                Model name & Data & Detector & Modules & FPS & C/GPU & Reference \\
                \midrule
                 \begin{tabular}{l}
                      MoveSORT \\
                       (\ref{tr:MoveSORT})
                 \end{tabular}           & 
                 \begin{tabular}{r}
                      COCO, \\ 
                      S-MOT  
                 \end{tabular} & 
                 \begin{tabular}{r}
                      YOLOv7  \\
                      (\cite{wang2022yolov7})
                 \end{tabular}   &
                  \begin{tabular}{r}
                      +ECC \cite{evangelidis2008parametric}, \\
                      +NSA K. \cite{du2021giaotracker}
                 \end{tabular}                 &
                 \begin{tabular}{r}
                      10  \\
                      (d+t)
                 \end{tabular}                  & 
                 \begin{tabular}{r}
                  T4  \\
                 \end{tabular}                   & 
                 \begin{tabular}{r}
                  DeepSORT  \\
                 \cite{wojke2017simple}
                 \end{tabular}                   
                    \\
                    \midrule
                 \begin{tabular}{l}
                      byteTracker  \\
                    (\ref{tr:byteTracker})
                 \end{tabular}                   &
                  \begin{tabular}{r}
                      COCO,    \\
                      S-MOT
                 \end{tabular}                 &
                \begin{tabular}{r}
                      YOLOX    \\
                      \cite{ge2021yolox,yoloxrepo}
                \end{tabular}                   & &
                  \begin{tabular}{r}
                      6  \\
                      (d+t)
                 \end{tabular}                  &                 
                  \begin{tabular}{r}
                  A100  \\
                 \end{tabular} & 
                   \begin{tabular}{r}
                      ByteTrack  \\
                     \cite{zhang2022bytetrack,bytetrackrepo}
                 \end{tabular}                
                    \\
                \midrule
                \begin{tabular}{l}
                      StrongerSORT  \\
                    (\ref{tr:StrongerSORT})
                 \end{tabular}                    & 
                 \begin{tabular}{r}
                      S-MOT    \\ Market1501
                 \end{tabular}                  &  
                 \begin{tabular}{r}
                     pub. det.   \\
                      (YOLOv7)
                 \end{tabular}                   & 
                  \begin{tabular}{r}
                      -AFLink, -GSI    \\
                        \cite{Du2022}, +PCB \cite{PCB}
                 \end{tabular}
                 & 
                \begin{tabular}{r}
                     10   \\
                     (t) 
                 \end{tabular}                    & 
                    \begin{tabular}{r}
                  M1  \\
                 \end{tabular} & 
                \begin{tabular}{r}
                      StrongSORT  \\
                    \cite{Du2022,strongsortgithub}
                 \end{tabular}                
                       \\ 
                \midrule
                \begin{tabular}{l}
                      MOT  \\
                    (\ref{tr:MOT})
                 \end{tabular}                  &  
                \begin{tabular}{r}
                      COCO,    \\
                      S-O$^\text{all}$
                 \end{tabular} &
                \begin{tabular}{r}
                      Casc. R-CNN      \\
                      ResNet-50,
                 \end{tabular}                     & & 
                 \begin{tabular}{r}
                      1  \\
                    (d+t) 
                 \end{tabular}                      & 
                     \begin{tabular}{r}
                  V100  \\
                 \end{tabular}                      &
                 \begin{tabular}{r}
                      DeepSORT  \\
                    \cite{wojke2017simple}
                 \end{tabular}                       \\
                \midrule
                \begin{tabular}{l}
                      OCSORT  \\
                    (\ref{tr:OCSORT})
                 \end{tabular}                 &  
                \begin{tabular}{r}
                      S-O$^\text{all}$,    \\
                       S-MOT$^\text{all}$
                 \end{tabular}                 & 
                \begin{tabular}{r}
                      YOLOX-XL    \\
                        \cite{ge2021yolox,yoloxrepo}
                 \end{tabular}                & & 
                \begin{tabular}{r}
                      20  \\
                    (d+t) 
                 \end{tabular}                  & 
                 \begin{tabular}{r}
                  V100  \\
                 \end{tabular}                  & 
                \begin{tabular}{r}
                      OCSORT  \\
                    \cite{cao2022observation,ocsortgithub}
                 \end{tabular}                    \\
                  \midrule
                \begin{tabular}{l}
                  Tracktor   \\
                Baseline (\ref{tr:Tracktor Baseline})
             \end{tabular}                      & 
                \begin{tabular}{r}
                       COCO, \\
                       S-MOT
                 \end{tabular}                & 
                \begin{tabular}{r}
                       F. R-CNN,    \\
                        ResNet-50
                 \end{tabular}                & 
                \begin{tabular}{r}
                       ECC \cite{evangelidis2008parametric}   \\
                        
                 \end{tabular}                 & 
                \begin{tabular}{r}
                      10  \\
                    (d+t) 
                 \end{tabular}            & 
                 \begin{tabular}{r}
                  RTX  \\ 3080
                 \end{tabular}                  & 
                 \begin{tabular}{r}
                      Tracktor  \\
                    \cite{chen2019mmdetection}
                 \end{tabular} 
                  \\

\bottomrule
\end{tabular}
\end{table*}

We received 18 submissions from 7 different institutions. Additionally, we provided a baseline, i.e.~a Tracktor-based tracker using ECC with a Faster R-CNN ResNet-50 detector (\ref{tr:Tracktor Baseline}). We used the mmtracking implementation \cite{chen2019mmdetection} with default hyperparameters. We also provided public detections so that participants do not need to train their own detectors. These are from a YOLOv7 model pretrained on COCO and trained on SeaDronesSee-MOT train set for 8 epochs yielding an AP of roughly $0.5$. For reference, the same model (except for the number of class outputs) has an AP of $0.4181$ on Object Detection v2, which is not optimal (compare to best models).

All of the 18 submitted trackers outperformed the baseline. See an overview of the submitted methods in Table \ref{tab:mot_submissions_overview}. Table \ref{tab:mot_results} shows the results of the best submissions of the best five teams. All submissions followed the tracking-by-detection paradigm. Since it was allowed to train on any data, most submissions did so and incorporated stronger detectors as the provided public detection baseline.

\begin{table*}[tb]
\centering
\caption{Multi-Object Tracking  results on the SeaDronesSee-MOT test set. The submissions are ranked based on HOTA. The last row indicates the baseline. Gold, silver and bronze denote the first, second and third place, respectively.}
\label{tab:mot_results}
\vspace{-.2cm}
\begin{tabular}{lrrrrrrrrrrrrr}
\toprule
                Model name &     HOTA & 	MOTA & 	IDF1 &	MOTP & 	MT & 	ML & 	FP & 	FN & 	Re & 	Pr & 	IDs & 	Frag \\
\midrule
                \color{gold(metallic)} MoveSORT &  \color{gold(metallic)}	0.67 & \color{gold(metallic)}	0.80 & \color{gold(metallic)}	0.77 & \color{bronze}	0.19 & \color{gold(metallic)}	311 & \color{gold(metallic)}	71 & \color{lightslategray} 8761 & \color{gold(metallic)}	10009 & \color{gold(metallic)}	0.89 & \color{gold(metallic)}	0.91 & \color{gold(metallic)}	44 & \color{bronze} 805 \\
                
                \color{lightslategray} byteTracker &  \color{lightslategray}	0.65 & \color{lightslategray}	0.77 &	\color{gold(metallic)} 0.77 & \color{gold(metallic)}	0.21 &	260 &	113 & \color{bronze}	10569 & \color{bronze}	11123 & \color{lightslategray}	0.88 & \color{lightslategray} 	0.89 & \color{lightslategray}	68 & 	841 \\
                
                \color{bronze} StrongerSORT & \color{bronze} 0.63 &	0.74 & \color{lightslategray}	0.75 & \color{lightslategray}	0.20 & \color{bronze}	303 & \color{lightslategray} 73 	& 10779 &	13308 &	\color{bronze} 0.86 & \color{bronze}	0.88 &	243 &	1396 \\
                
                MOT &  	0.62 & \color{bronze} 0.76 & \color{bronze}	0.71 &	\color{bronze} 0.19 & \color{lightslategray}	305 & \color{bronze}	79 &	11534 & \color{lightslategray}	10657 & \color{gold(metallic)} 0.89 & \color{bronze}	0.88 &	445 & \color{lightslategray}	672 \\
                
                OCSORT & 0.61 &	0.72 &	0.69 & \color{bronze}	0.19 &	291 &	97 & \color{gold(metallic)} 7836 &	18018 &	0.81 & \color{gold(metallic)}	0.91 & \color{bronze}	106 &	\color{gold(metallic)} 671 \\
                
                Tracktor Baseline & 0.46 &	0.48 &	0.50 & \color{gold(metallic)}	0.21 &	175 &	157 &	11960 &	35765 &	0.62 &	0.83 &	1435 &	2522\\

\bottomrule
\end{tabular}
\end{table*}

\begin{figure*}[tb]
\centering
   \includegraphics[width=\textwidth]{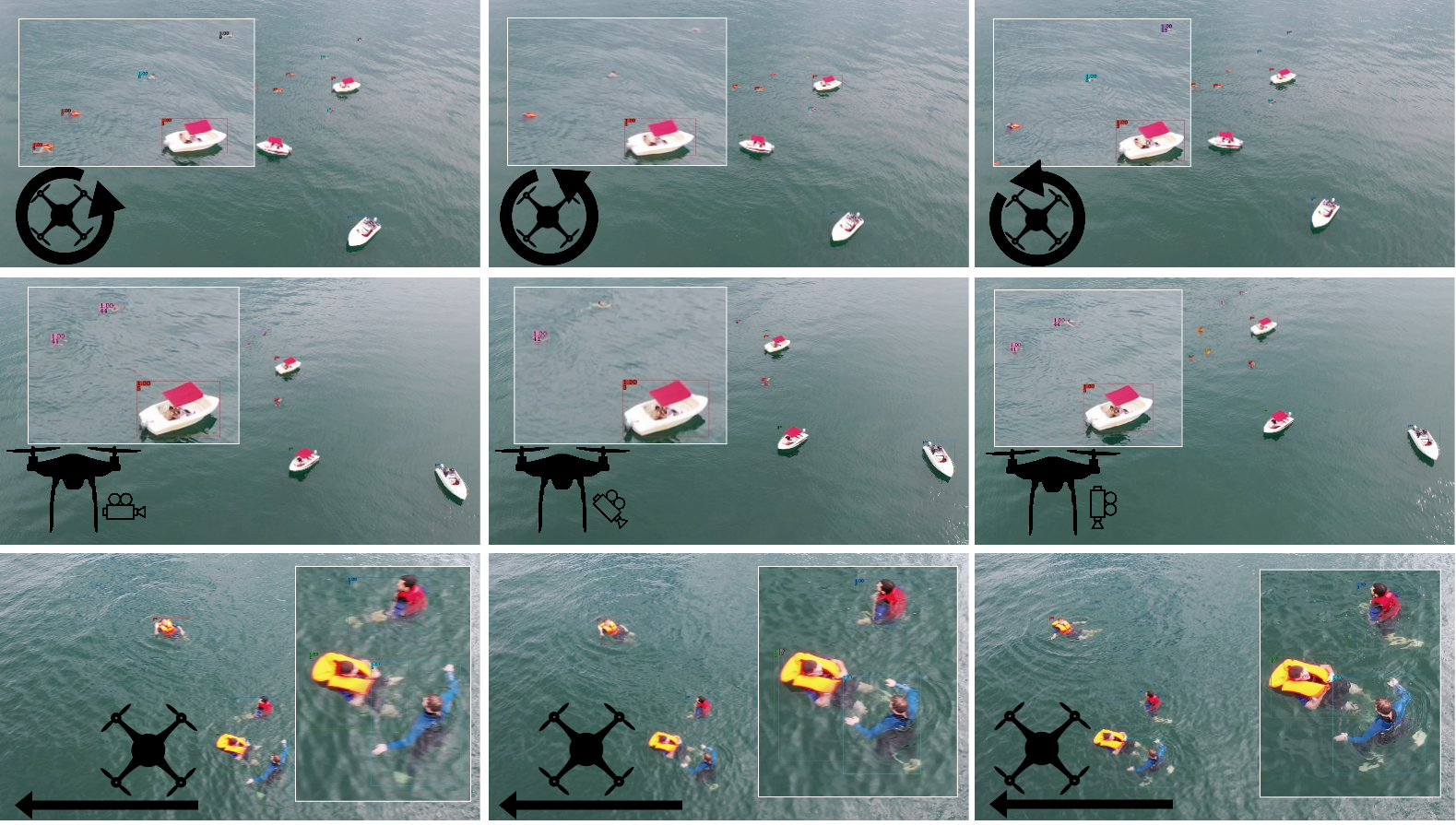}
\caption{Three common types of error causes. Predicted tracks from MoveSORT. Top: Panning by changing the heading angle cause the tracker to lose the three swimmer in the second frame. Middle: Tilting the camera has the same effect. Bottom: Fast movements of the UAV cause a duplicate detection.}
\label{figMOT:overviewtrackingerrors}
\end{figure*}

MoveSORT (\ref{tr:MoveSORT}) performed best in terms of HOTA, MOTA and IDF1 metrics although they only trained on SeaDronesSee-MOT. Being the best model in these metrics suggests that it is a very robust model w.r.t. detection and association accuracy. 
However, they relied on the recent YOLOv7 \cite{wang2022yolov7} detector, which may yield good detection results to work with. Notably, it only made 44 ID switches, which may the cause of the underlying DeepSORT implementation, which focuses specifically on decreasing the number of ID switches. However, also note all models have rather low id switch numbers, which is due to the sparse nature of the dataset where objects are not too cluttered (see e.g. Fig. \ref{figMOT:overviewtrackingerrors}).
MoveSORT further claims to improve on DeepSORT by using the enhanced correlation coefficient maximization module (ECC) to estimate the global rotation and translation between adjacent frames (\ref{tr:MoveSORT}). Indeed, association between frames for fast moving camera movements is a problem in certain video clips of SeaDronesSee-MOT as exemplified qualitatively in Figure \ref{figMOT:overviewtrackingerrors}.
Furthermore, they added the NSA Kalman filter module \cite{du2021giaotracker} from the second place tracker of the VisDrone 2021 MOT challenge \cite{chen2021visdrone}.

%TODO not anymore
%\textbf{In Figure TODO}, we visualize ECC and a linear motion module from the implementation of the Tracktor baseline \cite{chen2019mmdetection}, where the ground truth bounding box of an object from the previous frame is plotted on the next frames. \textbf{TODO ECC} where a fast pan of the UAV camera causes the qualitative and IoU plot -surprising considering that not many \textbf{features TODO}.

\begin{table*}[tbh]
\centering
\caption{Multi-Object Tracking rounded HOTA results in \% on video clips $0-21$ (7, 8, 20 do not exist) on the SeaDronesSee-MOT test set. The last row indicates the baseline. Gold, silver and bronze denote the first, second and third place, respectively.}
\label{tab:mot_clip_results}
\vspace{-.2cm}
\begin{tabular}{lrrrrrrrrrrrrrrrrrrr}
\toprule
                        Model name &                         0 &                         1 &                         2 &                         3 &                         4 &                         5 &                         6 &                         9 &                        10 &                        11 &                        12 &                        13 &                        14 &                        15 &                        16 &                        17 &                        18 &                        19 &                        21 \\
\midrule
                 MoveSORT &  \color{lightslategray}63 &                 \color{bronze}       46 &                        80 &  \color{lightslategray}67 &  \color{gold(metallic)}70 &          \color{bronze}79 &  \color{gold(metallic)}42 &  \color{lightslategray}59 &                        51 &  \color{lightslategray}65 &  \color{lightslategray}86 &  \color{lightslategray}57 &  \color{gold(metallic)}66 &  \color{lightslategray}75 &  \color{gold(metallic)}67 &  \color{gold(metallic)}91 &  \color{lightslategray}52 &                        92 &  \color{lightslategray}66 \\
 \color{lightslategray}byteTracker &          \color{bronze}60 &  \color{lightslategray}61 &          \color{bronze}82 &          \color{bronze}64 &                        65 &  \color{lightslategray}82 &  \color{lightslategray}38 &          \color{bronze}53 &  \color{lightslategray}55 &                        57 &                        83 &  \color{gold(metallic)}62 &          \color{bronze}61 &          \color{bronze}72 &          \color{bronze}60 &  \color{lightslategray}88 &  \color{gold(metallic)}56 &  \color{lightslategray}96 &  \color{gold(metallic)}67 \\
        \color{bronze}StrongerSORT &  \color{gold(metallic)}64 &                        37 &                        77 &                        47 &  \color{lightslategray}69 &                        72 &          \color{bronze}35 &  \color{gold(metallic)}60 &          \color{bronze}53 &                        61 &                        83 &                        52 &                        56 &  \color{lightslategray}75 &                        56 &                  \color{bronze}      85 &                        47 &                        87 &  \color{lightslategray}66 \\
                               MOT &                        57 &  \color{gold(metallic)}65 &  \color{lightslategray}84 &                        52 &          \color{bronze}68 &  \color{gold(metallic)}85 &                        31 &                        47 &  \color{lightslategray}55 &          \color{bronze}64 &  \color{gold(metallic)}87 &                        52 &  \color{lightslategray}65 &  \color{gold(metallic)}78 &  \color{lightslategray}63 &  \color{gold(metallic)}91 &                        43 &  \color{gold(metallic)}97 &                        60 \\
                            OCSORT &                        57 &                        43 &  \color{gold(metallic)}85 &                        22 &                        65 &  \color{lightslategray}82 &  \color{lightslategray}38 &                        52 &  \color{gold(metallic)}61 &  \color{gold(metallic)}68 &          \color{bronze}85 &          \color{bronze}56 &                        15 &                        66 &                        57 &  \color{gold(metallic)}91 &          \color{bronze}50 &          \color{bronze}94 &          \color{bronze}63 \\
                          Baseline &                        47 &                        31 &                        56 &   \color{gold(metallic)}8 &                        49 &                        31 &                        22 &                        35 &                        18 &                        47 &                        61 &                        37 &                        20 &                        50 &                        37 &                        80 &                        40 &  \color{gold(metallic)}97 &                        56 \\
                          \midrule
          Average &                        58 &          47 &                        77 &                        43 &                        64 &                        71 &                        34 &                        51 &                        48 &                        60 &                        80 &                        52 &                        47 &                        69 &                        57 &          87 &                        48 &                        93 &          63 \\
\bottomrule
\end{tabular}
\end{table*}

\begin{figure*}[tbh]
\centering
   \includegraphics[width=\textwidth]{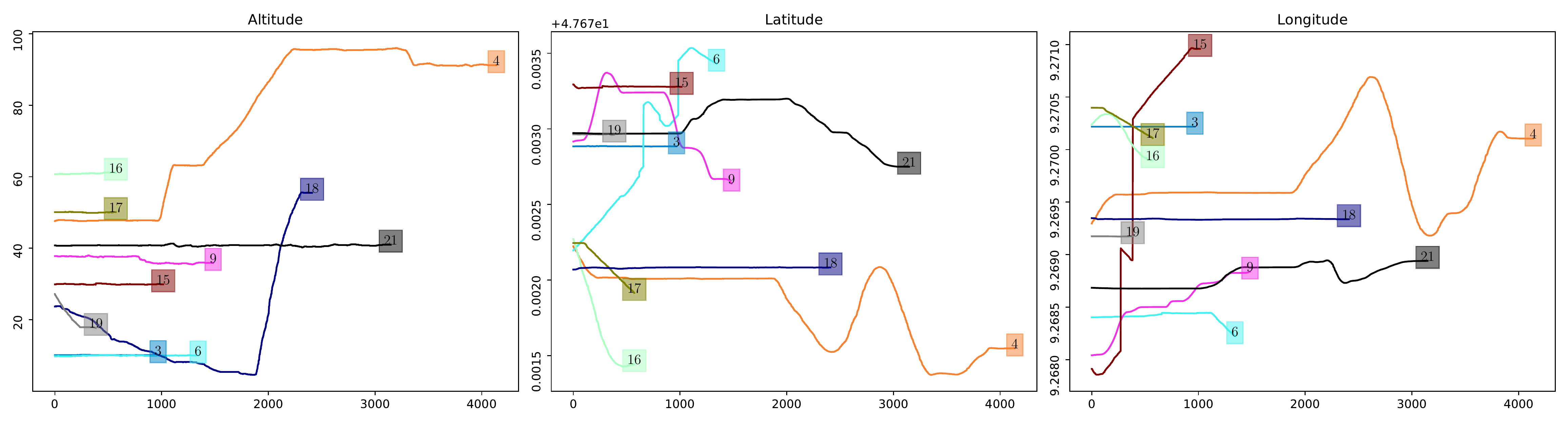}\\
   \includegraphics[width=\textwidth]{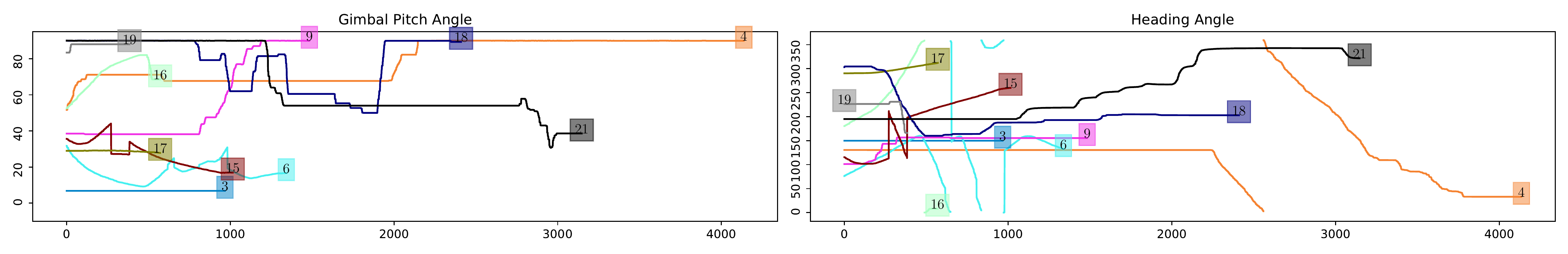}
\caption{Meta data visualization of video clips with lengths longer than 300 frames.}
\label{figMOT:longer300}
\end{figure*}

\begin{figure}[tbh]
\centering
   \includegraphics[width=0.46\textwidth]{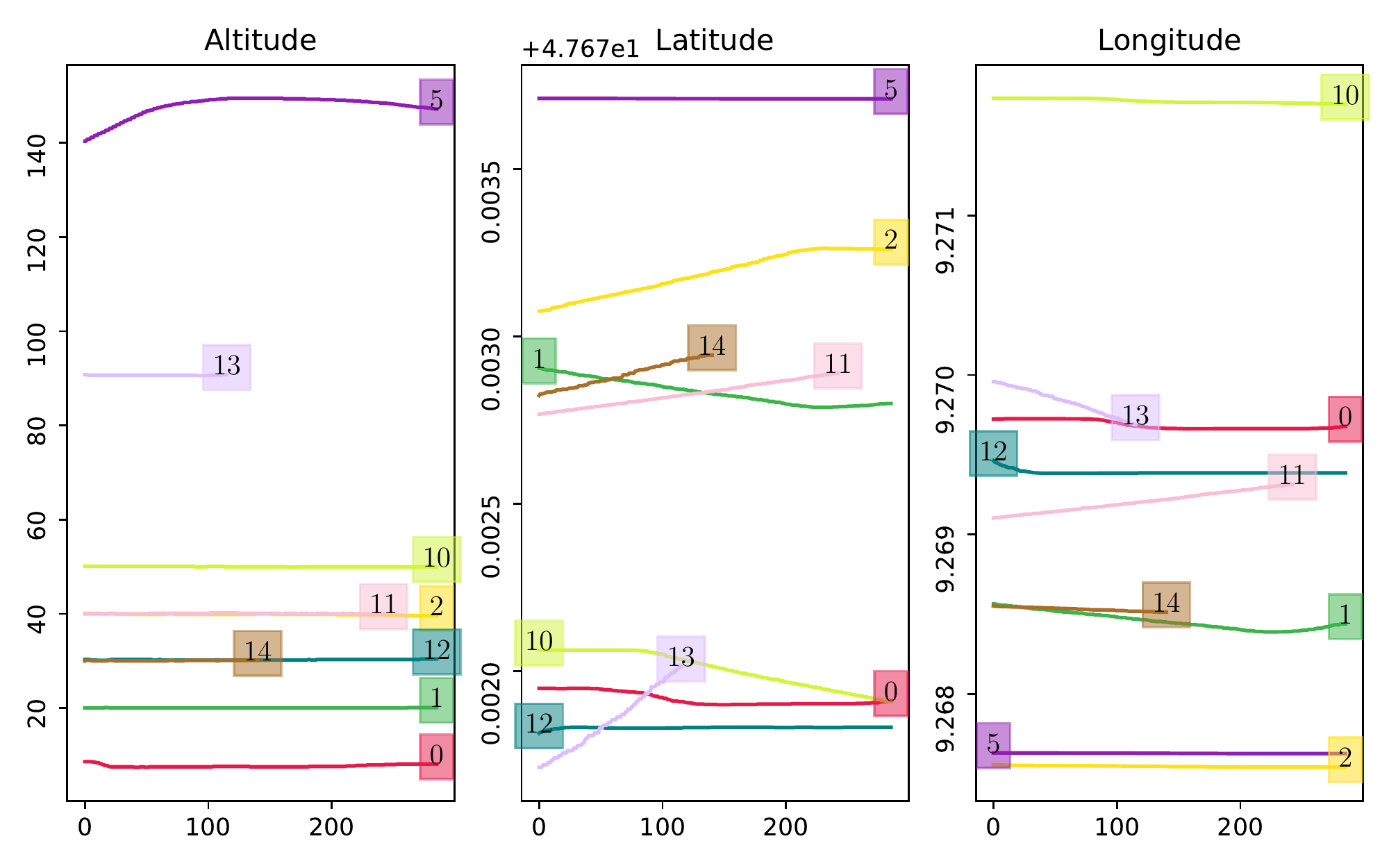}\\
   \includegraphics[width=0.46\textwidth]{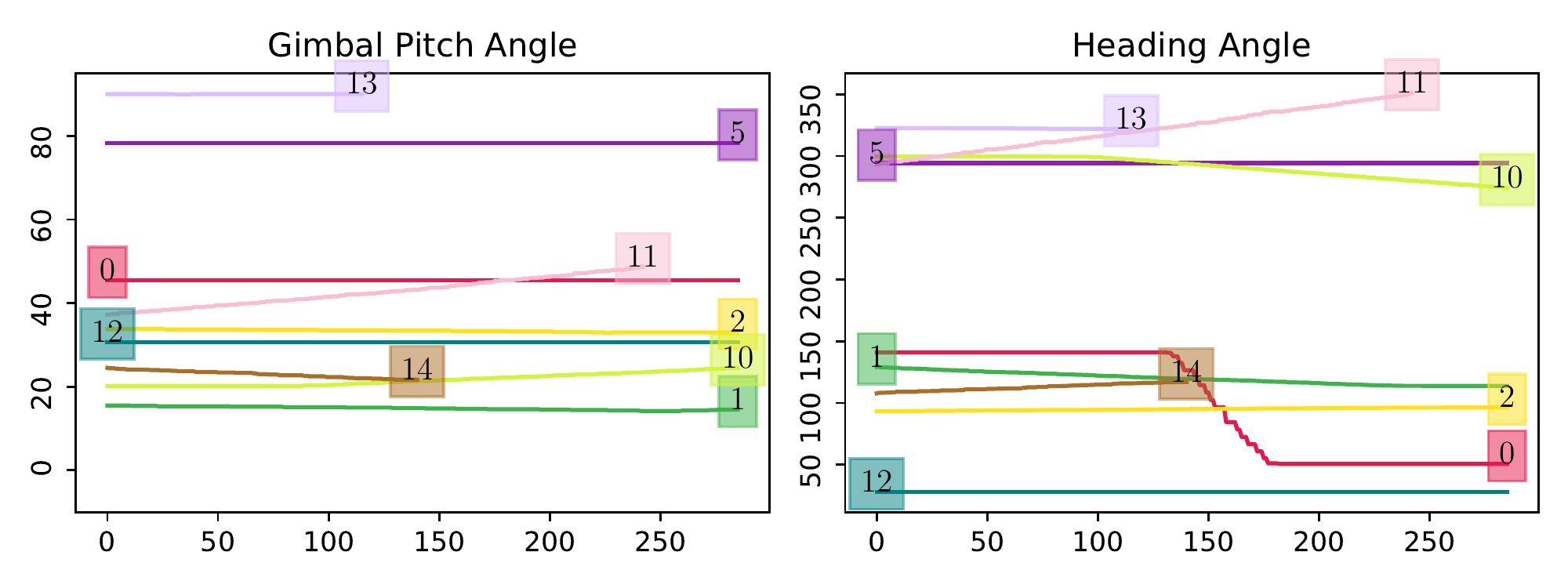}
\caption{Meta data visualization of video clips with lengths shorter than 300 frames.}
\label{figMOT:shorter300}
\end{figure}

The method byteTracker (\ref{tr:byteTracker}) placed second basing their submission on the recent ByteTrack implementation\cite{bytetrackrepo}. They adapted the tracker's focus on the MOT17 challenge \cite{MOT16} to the maritime setting by removing the vertical bounding box restriction and by changing hyperparameters, such as increasing the non-max-suppression threshold to remove potential false associations and to decrease the number of ID switches (\ref{tr:byteTracker}). The underlying detector was a large YOLOX-x model, which might explain some of the good performance.

StrongerSORT (\ref{tr:StrongerSORT}) placed third, mostly owing to its ability to reliably associate tracklets as indicated by its high IDF1 score and few lost tracklets (ML). They removed the newly introduced GSI and AFLink and added the part-based re-identification model PCB, which is pretrained on Market1501. 
This submission relied on the sub-optimal provided public detections. 
Moreover, in a second submission, STI-StrongSORT, they hypothesized that spatio-temporal information is more important than appearance-based information from a re-identification model. 
They based this hypothesis on the observation that objects have a very similar appearance and because occlusions are rare. 
With their proposed changes they manage to increase the speed from 10fps to 30fps. 
With a competitive HOTA score, they managed to decrease the number of ID switches fivefold.

MOT (\ref{tr:MOT}) placed fourth as measured in HOTA, but placed third as measured in MOTA and IDF1. They based their submission also on DeepSORT.  However, they trained their detector on the whole SeaDronesSee ODv2 dataset (train+val), which has a larger domain/appearance variance, but fewer (yet less correlated) images. Furthermore, their backbone is a ResNet-50 ($\sim$23M parameters), which is small compared to, e.g., a YOLOX-x with almost 100M parameters. They adapted to the aerial domain by setting appropriate scale parameters for the anchors and employed several train and test augmentation strategies while tuning respective hyperparameters. Similar to others, they set hyperparameters so as to ignore occlusion cases %\textbf{[TODO: not anymore maybe show video with occlusion from very low altitude?]}. 
They set the detection score for updating tracks to $0$, which may come from a similar motivation to that of ByteTrack \cite{zhang2022bytetrack}.

OCSORT (\ref{tr:OCSORT}) placed fifth as measured in HOTA, while having the smallest amount of fragmentations and the highest precision. They also employ the large YOLOX-x detector and train on all of SeaDronesSee ODv2 and MOT.

Table \ref{tab:mot_clip_results} shows the HOTA results of the models on all the test video clips. Interestingly, there is no clear best model on the majority of the clips. In the following, we try to explain some of the results on the clips, ordered from easiest to hardest clip.

In clip 19, only a single boat needs to be tracked which explains the high performances of all trackers. Similarly, clip 17 also shows only three boats which have to be tracked in a near static scene (compare to Figure \ref{figMOT:longer300}). Clips 2 and 12 are also static scenes (compare to Figure \ref{figMOT:shorter300}). While clip 5 is also static (small altitude increase), the high altitude causes many trackers to not detect and track the small swimmers. Clip 15 also only features boats although some of them are further away in the horizon and the movement and heading rotation of the drone in addition to the camera pitch angle change cause some trackers to fail to reliably track. See also Figure \ref{figMOT:overviewtrackingerrors} for examples of these errors. Clip 4 is the most dynamic one with camera and UAV panning and tilting and movement of the UAV in x,y and z directions. However, these movements are rather gently such that successful tracking can still be done by most of the trackers. Clip 21 features many swimmers and three boats. Furthermore, there are quick pitch angle changes along with a UAV movement and rotation. Clip 11 only shows a boat and a swimmer while the UAV is rotating around itself, although the swimmer is quite far away and hardly visible. Missing detections are punished relatively hard since the clip is short with few objects. While clip 0 is at very low altitude, there are many swimmer with a fast moving and rotating drone. Clip 16 shows many swimmers and boats and inherits a high dynamic range w.r.t. camera panning and tilting and movement of the UAV. Clip 13 shows a 90$^\circ$ scene where the UAV is moving at quite high altitude at constant speed. The swimmers are close to boats which is why it is hard to detect and track them. Clip 9 shows many swimmers with sudden changes in camera pitch and heading angle, resulting in many fragments and id switches. Clip 10 shows a few swimmers and several boats with a slow minimal camera pan. However, the acute angle lets swimmers appear very small and hard to detect. Similarly, clip 1 shows a scene with slowly rotating UAV and acute pitch angle, which results in many very far away swimmers that are quite small and are failed to detect robustly by most trackers. The hardest clip, 6, shows several boats and swimmers with a great amount of movement and dynamic camera panning and tilting. Furthermore, objects are hardly visible due to sun reflections.

\subsection{Discussion}
%TODO: check at other uav/sv sections whether they included a winners discussion

The submitted methods are already very strong. Many of the errors are still caused by very hard detections. However, the nature of UAV camera movements also cause several errors. Both, the detection and tracking errors could potentially be mitigated by using the available meta data. 

The winner method from Beijing University of Posts and Telecommunications, MoveSORT (\ref{tr:MoveSORT}), leveraged a recent YOLOv7 detectors but included several modules to enhance the performance. The second place from National University of Defense Technology, byteTracker (\ref{tr:byteTracker}) also employed the recent ByteTrack framework. The third place from EPFL, StrongerSORT (\ref{tr:StrongerSORT}), use the sub-optimal provided public detections to achieve the third place.

Further analysis would be necessary to discriminate based on classes and having real-time capable trackers with potentially worse but faster detection backbones. The necessity of certain tracking modules is also questionable in this setting, such as the reidentification module. Also, it is not clear how good the ECC module can really perform in the case of feature-poor maritime sceneries. 

%-discriminate based on classes
%-correct label errors
%-future research:
%    -fast camera movements -> ecc and linear might both fail 
%    -not real-time -> next time more focus on that (it was not the goal of this chal)
%    -reidentification really not helpful? -> analyze more
%    -reidentification when gone from scene (incoporate in dataset gts)

\section{USV-based Perception Challenges}
\label{sec:usvchallenges}

%\janezcomment{Whose text is below? Lojze?}
%\st{Unmanned Surface Vehicles (USVs) are robotic boats, that are able to operate without a crew. Their use cases range from small-scale such as automated inspection of marinas and trash collection to large-scale such as long-distance cargo shipping and ferry transport. The autonomy of USVs crucially depends on their perception capabilities.}

\begin{figure*}[!htb]
    \centering
        \includegraphics[width=\textwidth]{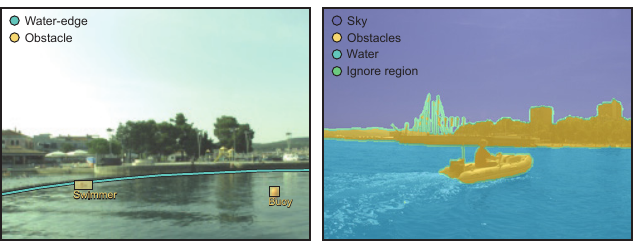}
    \caption{Two major perception tasks crucial for USV collision avoidance as defined by both MODS benchmark~\cite{MODSBenchmark2022} and the two MaCVi USV challenges: maritime object detection (left) and obstacle segmentation (right). The first problem assumes that static obstacles such as shoreline are known in advance by maps, while the second problem addresses prediction of the entire navigable area in which obstacle localization is implicit.}
\label{fig:MODS}
\end{figure*}

\begin{figure*}
    \centering
        \includegraphics[width=0.95\textwidth]{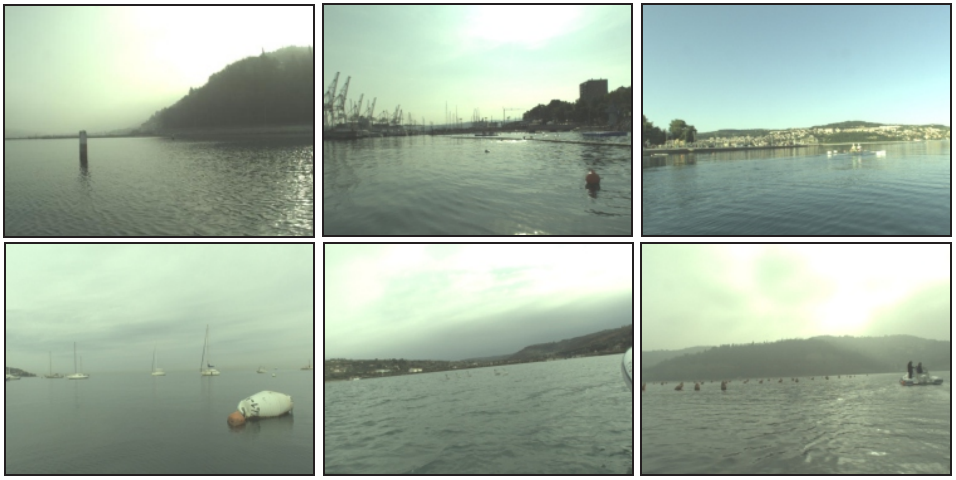}
    \caption{
    MaCVi USV challenges cover a broad range of obstacle appearances and types. In the above examples various boats, buoys, swimmers and even swans are visible.
    }
\label{fig:MODSsample}
\end{figure*}

Two USV-oriented challenges focusing on perception for maritime navigation were considered -- the obstacle segmentation and obstacle detection challenges (Fig.~\ref{fig:MODS}). Both challenges were based on the recent MODS Benchmark~\cite{MODSBenchmark2022}.
The challenges provided images from the viewpoint of a small USV, with the overall goal to detect obstacles and the boundaries of the visible water surface and thus prevent any kind of collisions that would endanger either the USV or its environment.
The challenges include a wide variety of obstacles, as it can be seen in Fig.~\ref{fig:MODSsample}.

%\janez{Two USV-centered challenges were considered. The foundation for both challenges was provided by the recently published MODS Benchmark~\cite{MODSBenchmark2022}
%. The benchmark and the challenge itself are 
%geared towards computer vision for autonomous navigation by USVs in the maritime environment.}

%\janez{The challenges provided images from the viewpoint of a small USV, with the overall goal to detect obstacles and the boundaries of the visible water surface and thus prevent any kind of collisions that would endanger either the USV or its environment. The MODS benchmark contains two tasks, which are geared towards different kind of computer vision algorithms and are shown in Fig.~\ref{fig:MODS}. MODS Benchmark and both challenges include wide variety of obstacles, as it can be seen in Fig.}~\ref{fig:MODSsample}.

%\janez{This challenge covers one of the important aspects of USV perception -- robust and efficient obstacle detection, which is an essential component and impacts downstream tasks like navigation and path planning.}

% With the release of several segmentation-based datasets (\textbf{TODO}) in recent years, several works tried to address maritime obstacle detection using semantic segmentation methods as they are able to address different types of obstacles (dynamic obstacles and static background) is a unified way. Due to the specifics of the marine domain, popular methods from other domains do not generalize well to the marine domain. For this reason several methods have been developed specifically for the maritime domain. To evaluate these segmentation methods from the perspective of USV safety and navigation, the MODS benchmark~\cite{MODSBenchmark2022} has been proposed.

\subsection{Dataset}

For both USV challenges, the MODS dataset~\cite{MODSBenchmark2022} was provided by the challenge organizers. A large corpus of initial sequences was acquired during eight voyages with our over a span of seven months in the years 2018-2019. The sequences were captured in two geographically disjoint areas of Slovenian coastal waters (port of Koper and close to resort village of Strunjan) to diversify the obstacles and environment appearance. To further diversify the dataset and capture the realism of USV missions, the voyages were planned at different times of the day and under different weather conditions. An expert manually piloted the USV and included realistic navigation scenarios with dangerous situations in which the boat is heading straight towards an obstacle or passing it by in a close range. Illustrative selection of images from the MODS dataset can be seen in the Fig.~\ref{fig:MODSsample}.

%\subsubsection{Data acquisition}

\textbf{Data acquisition.} Approximately forty-eight hours of footage with on-board synchronized sensors (in particular, stereo cameras, IMU, compass and GPS) was captured under the described protocol. The recordings were cut into sequences with interesting navigation scenarios and out of these, $94$ sequences, jointly containing $80,828$ images were selected. In the sequence selection, care was taken to include many diverse obstacle interactions as well as phenomenons challenging for visual recognition such as prominent sun-glitter, distinct sea-foam and driving through dense shellfish farms and floating debris.

%\subsubsection{Data annotation and refinement}

\textbf{Data annotation and refinement.} To reduce the annotation burden, while maintaining the dataset diversity, only every 10-th frame was annotated (i.e., once per second). The annotation task involved placing a tight bounding box over each dynamic obstacle and assign it a high-level label: \textit{vessel}, \textit{person} or \textit{other}. The MODD protocol from~\cite{KristanCYB2015,bovcon2018stereo} was followed for static obstacles annotation by drawing a polygon over their lower edge, where the obstacle touches the water (i.e., the water-obstacle edge). This type of annotation was chosen since the obstacle-water edge denotes the most informative part used for practical robotic navigation. For example, inaccurate segmentation of the upper part of a pier does not affect navigation, however incorrect segmentation of the part touching the water can lead to collision.

Finally, the data was refined and corrected by experienced researchers with background in maritime computer vision. A Matlab tool was designed for this stage to allow easy manipulation of the existing annotations, addition of categorical labels (vessel, person, other) to the dynamic obstacles and cross-frame label propagation. The final annotations were screened by another expert to ensure labeling consistency. This amounted to $63,579$ dynamic object annotations and $10,706$ obstacle-water edge annotations which appeared in $99.3\%$ of frames.

\subsubsection{The danger zone}\label{sec:danger-zone}

\begin{figure}
    \centering
    \includegraphics[width=0.95\columnwidth]{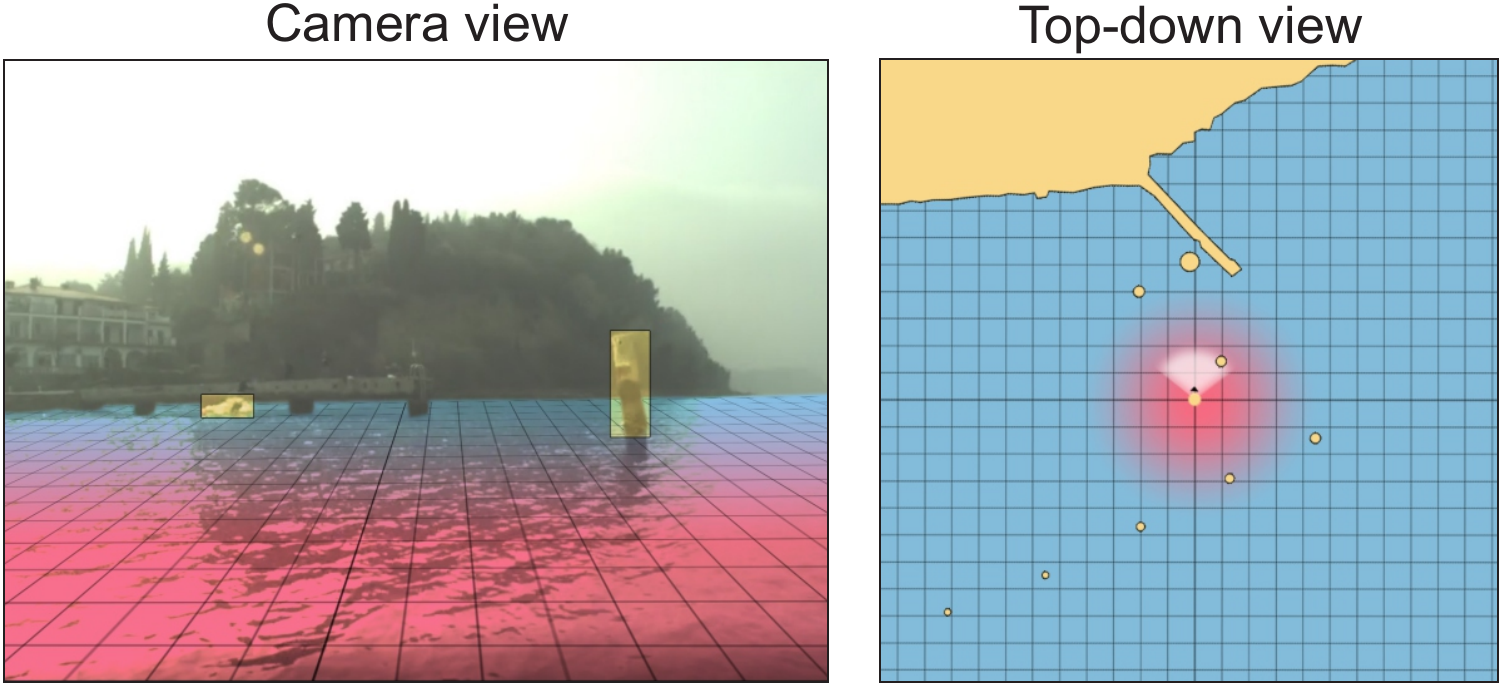}
    \caption{Nearby objects require immediate attention to avoid potential collision. A 15m hazardous area around USV (i.e., danger zone) is thus specified and visualized by a color gradient, ranging from red (dangerous) to blue (safe).}
    \label{fig:danger_zone}
\end{figure}

The danger that obstacles pose to the USV depends on their distance. Obstacles located in close proximity are more hazardous than distant ones. To address this, we defined a danger zone as a radial area, centered at the location of the USV. The radius is chosen in such a way, that the farthest point of the area is reachable within ten seconds when travelling continuously with an average speed of $1.5$m/s. Following~\cite{bovcon2019mastr} we thus estimate the danger zone in each image (see Figure~\ref{fig:danger_zone}) from the camera-IMU geometry. This opens the way for reporting method performance both on the whole image as well as constrained only to the danger zone.

\subsection{USV-based Obstacle Segmentation Challenge}

% TODO: Uvod -- [MK] OK, probably just state what is the task (predict three semantic ompoinents), what shoukd be the output of the method 

The goal of USV-based Obstacle Segmentation Challenge was to classify the pixels of an input image into three semantic categories -- obstacles, water or sky. To train semantic segmentation models for this purpose we suggested the use of the MaSTr1325 dataset~\cite{bovcon2019mastr}. Authors were allowed to use other datasets as well.

\subsubsection{Evaluation Protocol}

To evaluate segmentation predictions, we employ the MODS~\cite{MODSBenchmark2022} segmentation evaluation protocol. Segmentation methods provide per-pixel labels of semantic components (water, sky and obstacles). Traditional approaches for segmentation evaluation (\eg mIoU) do not reflect the aspects relevant for USV navigation. Instead the MODS protocol focuses on two important aspects of obstacle segmentation: water-edge estimation (static obstacles) and dynamic obstacle detection performance. The water-edge detection is analysed in terms of (1) localization accuracy ($\mu_A$), defined as the root mean square error (RMSE) computed from the distances between the ground truth water edge and the per-pixel vertical nearest water edge in the segmentation mask, and (2) detection robustness ($\mu_R$), defined as the percentage of correctly detected water edge pixels. A water-edge pixel is considered correctly detected when the vertical distance to the nearest water edge in the predicted segmentation is less than $\Theta_w = 20\text{px}$.

The dynamic obstacles detection accuracy is computed from the predicted obstacle segmentation mask as follows. First, true positives (TP) and false negatives (FN) are computed. A ground-truth dynamic obstacle counts as a TP if its bounding box region is covered sufficiently by the predicted segmentation, otherwise it counts as a false positives (FP). The coverage threshold is determined based on the automatically-estimated segmentation of the obstacle. Then, FP can be computed from predicted segmentations that fall outside GT obstacle bounding boxes. Regions that correspond to static obstacles (\ie above the water edge) are also removed from the segmentation mask. We determine the individual FP predictions using a connected-components-based approach on the remaining obstacle segmentations. For further details please see~\cite{MODSBenchmark2022}. Finally, the dynamic obstacle detection accuracy is summarized by the precision (Pr), recall (Re) and the F1-score metrics. We also report these metrics separately within the danger zone (Section~\ref{sec:danger-zone}).

\subsubsection{Submissions, Analysis and Trends}

\begin{table*}[t]
\centering
\caption{Overview of the submissions for the USV Obstacle Segmentation challenge. We outline the base model from which the methods were derived and whether the method used an ensemble approach. Ranking of the method on the leaderboard as well as the final placement of the teams are indicated. We also include the self-reported inference speeds.}
\label{tab:usvseg-overview}
% \vspace{-.2cm}
\begin{tabular}{lcrrcccccc}
\toprule
Place & Rank & Team & Model name & Section & Base & Ens & FPS & C/GPU & Avg. score \\
\midrule
\rowcolor{gold(bg)}
1st &    1  & BUPT      & Multi-WaSR   & \ref{usv-seg:multi-wasr}    &  WaSR       &  \checkmark & 12    & V100      &  93.5 \\
\rowcolor{silver}
2nd &    3  & HKUST     & MariFormer   & \ref{usv-seg:mari-former}   &  SegFormer  &             & 4     & RTX3090   &  93.2 \\
    % 6  & BUPT      & WaSR+        &  WaSR       &  \checkmark & 16    & RTX3090   &  92.5 \\
    &    13 & HKUST     & RevDeep      & \ref{usv-seg:hkust-revdeep} &  DeepLabv3  &             & 10    & RTX3090   &  91.6 \\
    &    15 & UL        & WaSR         & -                           &  WaSR       &             & 14    & RTX2080Ti &  91.3 \\
3rd &    16 & Xiaomi    & APTX003      & \ref{usv-seg:xi-aptx}       &  DeepLabv3+ &             & 3     & RTX3090   &  89.9 \\
4th &    17 & NCKU      & HRNet-OCR    & \ref{usv-seg:hrnet}         &  HRNet-OCR  &             & 4     & V100      &  89.6 \\
    &    18 & UL        & DeepLabv3    & -                           &  DeepLabv3  &             & 20    & RTX2080Ti &  89.5 \\
\midrule
5th &    24 & Couger AI & Lightnet     & -                           &  -          &             & -     & -         &  36.3 \\
\bottomrule
\end{tabular}
\end{table*}

\begin{figure}[t]
\centering
   \includegraphics[width=\linewidth]{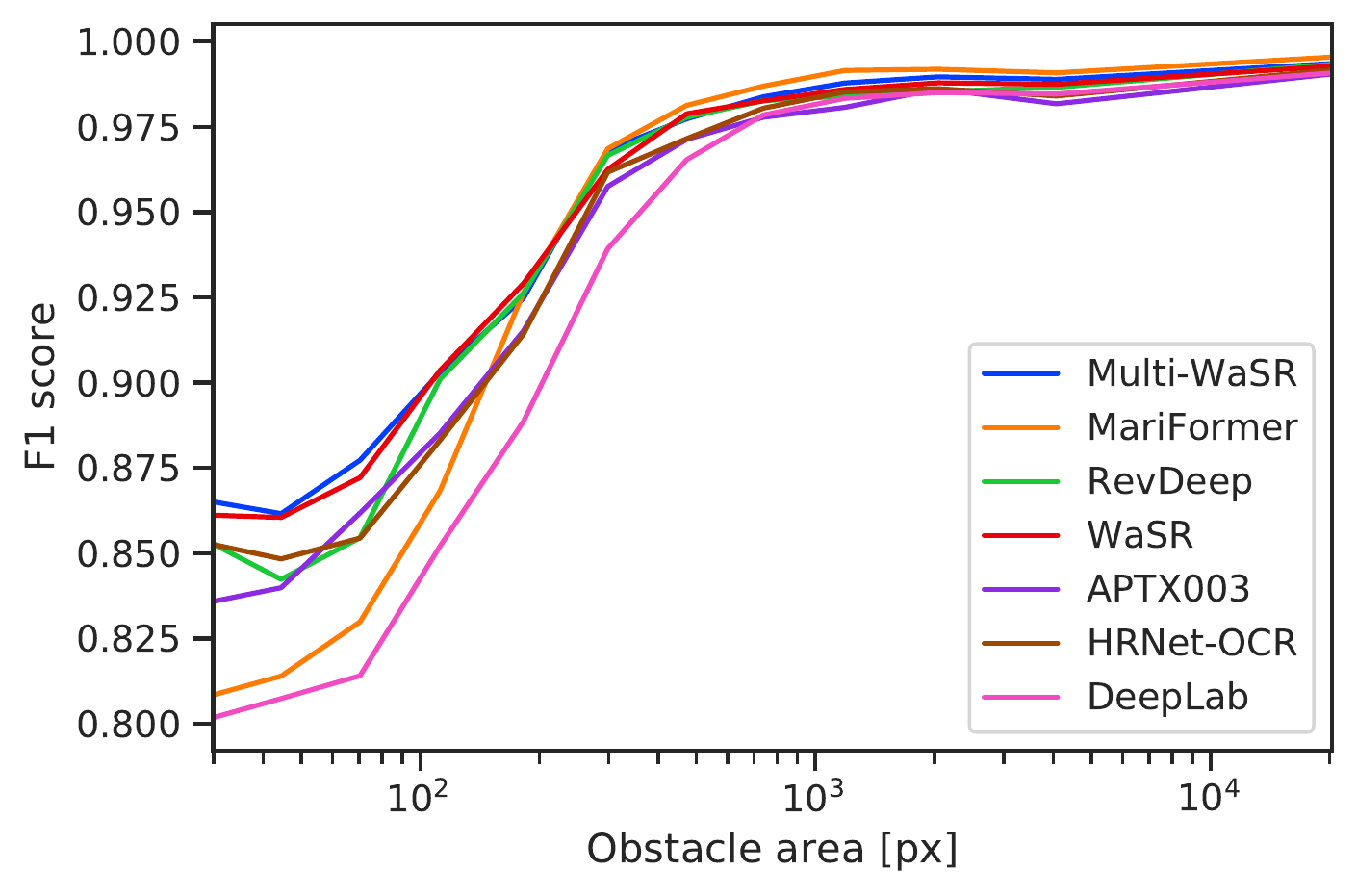}
\caption{Methods detection performance across different obstacle sizes.}
\label{fig:usvseg-sizes}
\end{figure}

% [Lojze] TODO: gold, silver, bronze metrics (like Benjamin)
\begin{table*}
\caption{Performance of submitted segmentation methods on MODS. Performance is reported in terms of F1 score, precision (Pr) and recall (Re) for dynamic obstacle detection, and water-edge detection accuracy ($\mu_A$) and robustness ($\mu_R$).}
\label{tab:usvseg-results}
\centering
\begin{tabular}{rlccccccccccccc}
\toprule
 & & & & & \multicolumn{3}{c}{Overall} & \phantom{a} & \multicolumn{3}{c}{Danger zone (\textless 15m)} & & \\
 \cmidrule{6-8} \cmidrule{10-12}
& method & $\mu_A$ & $\mu_R$ & & Pr & Re & F1 & & Pr & Re & F1 & & Avg. \\
\midrule
\#1  & Multi-WaSR &   14.8 &  97.9 &    &  96.0 &  92.6 &  \textbf{94.3} &     &  \textbf{90.4} &  95.2 &  92.7 & & \textbf{93.5} \\
\#3  & MariFormer &   \textbf{10.4} &  \textbf{98.6} &    &  \textbf{97.5} &  89.7 &  93.4 &     &  \textbf{90.4} &  95.5 &  \textbf{92.9} & & 93.2 \\
\#13 & RevDeep    &   14.1 &  98.0 &    &  95.7 &  91.8 &  93.7 &     &  85.3 &  94.3 &  89.6 & & 91.6 \\
\#15 & WaSR       &   16.5 &  97.7 &    &  95.6 &  \textbf{92.7} &  94.1 &     &  82.9 &  94.7 &  88.4 & & 91.3 \\
\#16 & APTX003    &   33.7 &  94.8 &    &  93.8 &  92.1 &  92.9 &     &  79.4 &  \textbf{96.0} &  86.9 & & 89.9\\
\#17 & HRNet-OCR  &   11.4 &  98.3 &    &  95.5 &  91.8 &  93.6 &     &  77.5 &  95.3 &  85.5 & & 89.6 \\
\#18 & DeepLabv3  &   17.1 &  97.6 &    &  93.7 &  89.4 &  91.5 &     &  81.6 &  94.4 &  87.6 & & 89.5 \\
% Lightnet   &  430.4 &  21.1 &    &  48.8 &  91.3 &  63.6 &     &   4.8 &  65.1 &   8.9 \\
\bottomrule
\end{tabular}
\end{table*}

\begin{figure*}
    \centering
        \includegraphics[width=\textwidth]{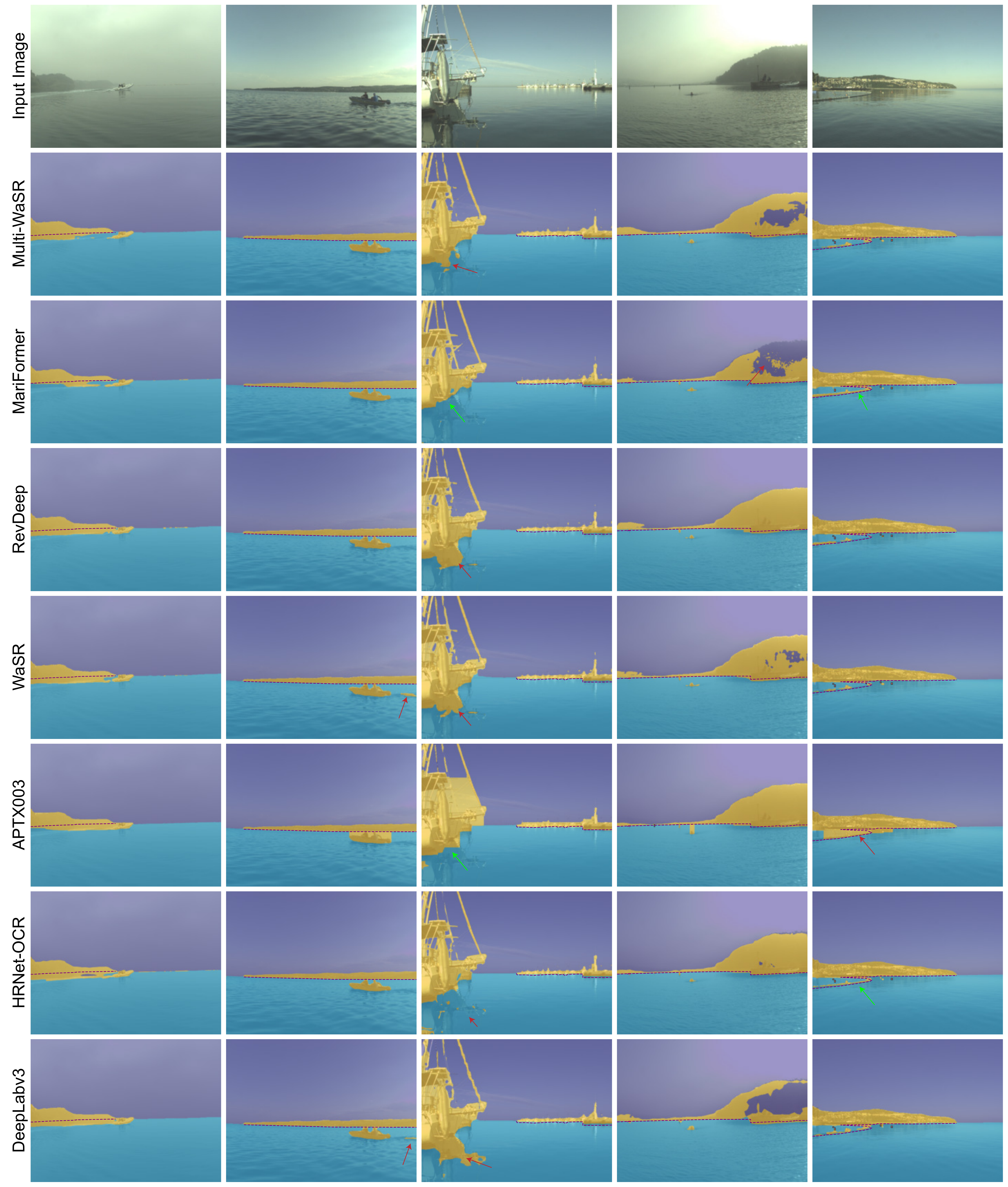}
    \caption{Qualitative comparison of methods for USV obstacle segmentation. Examples of errors and good behaviour are highlighted.}
\label{fig:usvseg-qualitative}
\end{figure*}

We have received 26 submissions from 5 different teams. This includes two baselines provided by the MaCVi2023 committee, DeepLabv3 (\ref{usv-seg:deeplab-baseline}) and WaSR (\ref{usv-seg:wasr-baseline}). We have grouped the methods by teams and only analyse the best method by each team. Table~\ref{tab:usvseg-overview} presents the overview of the best submitted methods by individual teams. One of the teams submitted two reports of very different methods by different authors, thus we decided to include both. In this analysis we will focus on all the methods that have beaten the DeepLabv3 baseline in terms of the average score. In the following we will refer to the methods by their ranking on the leaderboard with the notation \rankn{$n$}, where $n$ is the rank of the method.

Overall, all models except \rankn{3} MariFormer use convolutional neural networks as the base. Models \rankn{13} RevDeep, \rankn{16} APTX003 and \rankn{18} DeepLabv3 are based on the DeepLabv3 family of models, \rankn{1} Multi-WaSR and \rankn{15} WaSR are based on the recent maritime model WaSR~\cite{Bovcon2021}, and \rankn{17} uses HRNet with an additional Transformer-based OCR module~\cite{HRNet-OCR}. Model \rankn{3} MariFormer on the other hand derived from a recent Transformer-based method SegFormer~\cite{xie2021segformer}. 

Authors make several tweaks and changes to the architecture or methodology to increase performance on this task. \rankn{1} Multi-WaSR extends the original WaSR architecture by replacing the Attention Refinement Modules (ARM) with two Transformer blocks. While this model does not achieve the best results \rankn{14}, authors train several models, each with their own strengths and weaknesses, and then use an ensemble approach to make predictions by combining the votes of several models. The final ensemble model achieves the 1st place on the leaderboard. This is also the only entry in this analysis that uses an ensemble approach.

Authors of \rankn{3} MariFormer remove the boundaries of the camera housing, that is visible in several sequences of MODS, to prevent its influence. \rankn{13} RevDeep employs label smoothing as regularization. \rankn{16} APTX003 uses conditional random fields (CRF) to refine the output segmentation maps, and morphological post-processing to fill holes in obstacle segmentations.

Almost all methods have been trained exclusively on the suggested MaSTr1325 dataset~\cite{bovcon2019mastr}. The exception is \rankn{13} RevDeep, which also utilizes the additional 153 images of MaSTr1478~\cite{Zust2022Temporal}. A majority of approaches also employs various image augmentations, such as color transformations, addition of noise and geometric transformations.

The detailed performace of the different methods is reported in Table~\ref{tab:usvseg-results}. The methods can be roughly grouped into three categories based on their performance: 1) state-of-the-art, 2) WaSR-like performance and 3) DeepLabv3-like performance. The 1st and 2nd overall methods \rankn{1} Multi-WaSR and \rankn{3} MariFormer achieve very similar performance and significantly outperform the WaSR baseline (+2.2\% and +1.9\% average F1). Multi-WaSR is slightly better overall (+0.9\% F1), while MariFormer performs slightly better inside the danger-zone (+0.2\% F1). \rankn{13} RevDeep performs on par with the WaSR baseline, outperforming it by 0.3\% in average F1 score, and outperforming the DeepLab baseline by 2.1\% in average F1. \rankn{17} APTX003 and \rankn{17} HRNet-OCR perform close to the DeepLabv3 baseline, outperforming it slightly (+0.4\% and +0.1\% average F1). 

Note that the largest differences between methods seem to be dictated by the performance inside the danger zone, where the precision varies widely. The danger zone is a common place for maritime visual artefacts such as sun glitter, reflections or foam which are often a source of FP detections. The precision (Pr) of the methods is thus largely determined by their robustness to such artefacts.

In terms of water-edge localization, \rankn{3} MariFormer achieves the best results, followed closely by \rankn{17} HRNet-OCR (+1.0 $\mu_A$). Both these methods outperform other methods by a large margin in this aspect, which suggests higher segmentation accuracy. We also observe this in the qualitative analysis (see Figure~\ref{fig:usvseg-qualitative}), where more accurate segmentation of thin objects such as ropes and water barriers is apparent. These two methods operate at a higher resolution than other approaches and use transformers, which might both contribute to this result. 

\textbf{Detection by obstacle size:} For a better insight into the strengths and weaknesses of different methods we also perform an analysis of the dynamic obstacle detection performance based on the obstacle sizes. To do this we group GT obstacles (and FP detections) by covered area (in pixels) into 12 equally populated bins and compute the F1 metric within each group. The results are presented in Figure~\ref{fig:usvseg-sizes}. The most significant difference between methods occurs on small obstacles. This is where \rankn{1} Multi-WaSR and \rankn{15} WaSR perform the best, while \rankn{3} MariFormer and \rankn{18} DeepLabv3 are the worst performing in this category. However, the performance of \rankn{3} MariFormer increases significantly with obstacle size and it achieves the best performance on large obstacles.

\subsubsection{Discussion and Challenge Winners}

We have received a lot of interesting entries into the challenge. 
Authors have explored various architectures, data augmentations and post-processing techniques. Tthe overall winners of the USV Obstacle Segmentation Challenge are:
\begin{description}
    \item[1\textsuperscript{st} place:] Beijing University of Posts and Telecommunications (BUPT) with Multi-WaSR, and
    \item[2\textsuperscript{nd} place:] Hong Kong University of Science and Technology (HKUST) with MariFormer.
\end{description}
The best performing method demonstrated that ensemble techniques can be effectively used to increase the robustness in this domain. The second best approach closely matches the detection performance of the winning method and demonstrates outstanding segmentation accuracy by using a transformer architecture and a higher output resolution. However, this comes at a large cost in inference speed. Exploring efficient ways to incorporate these techniques is an interesting direction for future work.

% \matejcomment{Lojze, can you summarize breifly the results here in a few sentences and state the winner of this challenge?}

\subsection{USV-based Obstacle Detection Challenge}
\label{sec:usvobstacledetectionchallenge}

\subsubsection{Evaluation Protocol}

To evaluate obstacle detection predictions, we employ the detection evaluation protocol of MODS~\cite{MODSBenchmark2022}. 
All competing algorithms were required to output detections of all waterborne objects of the MODS semantic classes: \emph{vessel}, \emph{person} and \emph{others} with rectangular axis-aligned bounding boxes.

We followed the standard COCO/LVIS object detection evaluation protocol from~\cite{lin2014coco,gupta2019lvis}, which is based on the Jaccard index, i.e., an intersection-over-union (IoU) between ground truth and detected bounding boxes. A detection counts as a true positive (TP) if its respective IoU exceeds a predefined threshold, otherwise it is counted as a false positive (FP).  Because the precise localization of waterborne obstacles is difficult, especially if the objects are small, we used IoU=0.3 for the detection threshold. Precision and recall are calculated over all the images in the dataset and the F1 score is reported as the primary performance measure.

In order to focus only on the dynamic obstacles and avoid detections of people and boats on land, we use the water edge annotations to exclude detections above the water edge, unless there exists an overlap with a ground truth annotation. Thus, a detector not reporting objects above the water edge does not count as a false negative. 
As per the LVIS protocol~\cite{gupta2019lvis}, false positives are not counted in the images that are labelled as \textit{not exhaustively annotated}. %\cmnt{A Python implementation of the evaluation routines is available from the project page.}
%at \url{vicos.si/Downloads}.}

The final score is composed of three different metrics:

\begin{itemize}
\item Average F1 score, $F1_1$ when taking into account the class of the ground truth and the prediction.
\item Average F1 score, $F1_3$ where the class information is ignored.
\item Average F1 score, $F1_3$ of objects within a 15m large radial area in front of the boat (i.e. danger zone). The ground truth and the detection bounding boxes are considered as within the danger zone if at least 50\% of the area lies within the danger zone.

\end{itemize}

To determine the winner of the challenge, the average of the above three F1 scores, $F1_{avg}$ was used as an overall measure of quality of the method. 

\begin{table*}[t]
\centering
\caption{Overview of the submissions for the USV Obstacle Detection challenge, with results. Ranking of the method on the leaderboard as well as the final placement of the teams are indicated. We also include the self-reported inference speeds, used hardware, and whether any other datasets were used in the training.}
\label{tab:usvdet-overview}
% \vspace{-.2cm}
\begin{tabular}{lccccccccccc}
\toprule
Place & Rank & Team & Model name & Section & Other & FPS & Hardware & $F1_{avg}$ & $F1_1$ & $F1_2$ & $F1_3$  \\
\midrule
\rowcolor{gold(bg)}
1st & 1 & \makecell{Fraunhofer\\IOSB} & DetectoRS & \ref{usv-det:detectors} & \checkmark & 5 & Tesla V100 & 0.546 & 0.265 & 0.400 & 0.973\\
\rowcolor{silver}
2nd & 2 & \makecell{Nvlab x\\Acvlab \hspace{4.7mm} } & PRBNet & \ref{usv-det:prbnet} & \checkmark & 6 & Tesla V100 & 0.514 & 0.236 & 0.328 & 0.980\\
 & 3 & \makecell{Nvlab x\\Acvlab } & Yolo v7 & - & \checkmark & 6 & Tesla V100 &  0.513 & 0.260 & 0.296 & 0.984\\
  & 4 & \makecell{Fraunhofer\\IOSB} & FIOSB KA & - & \checkmark & 17 & Tesla V100 & 0.509 & 0.223 & 0.328 & 0.976\\
\rowcolor{bronze(bg)}
3rd & 5 & Ocean U. & Ocean U. & \ref{usv-det:oceanu} & \checkmark & 0.5 & i7 CPU & 0.492 & 0.223 & 0.283 & 0.970\\
  & 6 & \makecell{Nvlab  x\\Acvlab } & \makecell{PRBNet\\Yolo v7}& - &  & 6 & Tesla V100 & 0.485 & 0.216 & 0.260 & 0.980\\
  & 7 & nutn & pcb & - &  & 10 & RTX 2080 &  0.457 & 0.187 &	0.218 & 0.965\\
  & 8 & ? & Yolo v7 & - &  & 20 & 1080ti & 0.443 & 0.156 & 0.228 & 0.944\\
  & 9 & NCKU & YOLO & - &  & 10 & RTX 2080 & 0.436 & 0.162 & 0.166 & 0.980\\  
\midrule
Baseline & - & UL & \makecell{Mask\\R-CNN} & - &   & 10 & RTX2080 &  	0.419 & 0.122 & 0.172 & 0.964\\
\bottomrule
\end{tabular}
\end{table*}

\subsubsection{Submissions, Analysis and Trends}
We received 9 submissions from five different teams (in one case, team/institution name was not provided). Submissions are listed in Table~\ref{tab:usvdet-overview}. Sorted by $F1_{avg}$ metric, the top of the list is dominated by two teams: Fraunhofer IOSB and Nvlab x Acvlab, whose methods ranked from first to the fourth. The best method of the single team determined their final place in the challenge, and therefore the third place went to Ocean U. team, with method ranked the fifth overall. All the submissions outperformed the baseline method, Mask R-CNN. Teams were invited to submit their technical reports, but we received only the reports from the Fraunhofer IOSB, Nvlab x Acvlab and Ocean U, which are provided in sections \ref{usv-det:detectors}, \ref{usv-det:prbnet} and \ref{usv-det:oceanu}, respectively.

Fraounhofer IOSB's wining submission is based on DetectorRS \cite{qiao2021detectors} architecture, with tweaks allowing it to detect smaller objects, and was extensively trained on several different datasets featuring water-borne environment. It is interesting that while it achieved the first rank according to the decisive $F1_{avg}$ metric, it fared poorly when observing only objects in danger zone (using the $F1_3$ metric), that is, in the 15 meter radius in front of the USV. Observing only $F1_3$, the method is ranked only fifth, but this is compensated with distinctively higher $F1_1$, which requires proper class information in addition to obstacle detection.

Nvlab x Acvlab's top submission is based on PRBNet~\cite{chen2021parallel}, trained on MS COCO, with extensive postprocessing to reduce the number of false positives. Dataset metadata (shore information) is also used for this purpose. It should be noted that this method outperforms the first ranked DetectorRS in the danger zone evaluation, using the $F1_3$ metric. 

Ocean U's submission, which was awarded the third place is based on Yolo v7~\cite{yolov7github} with modified computational block. The network was trained on MS COCO dataset, and the only adaptation to the marine domain by selecting marine-relevant categories, and merging all other categories into MODS-stipulated \emph{others} category.

\begin{figure*}
    \centering
    \begin{tabular}{ccc}
        \includegraphics[width=0.33\textwidth]{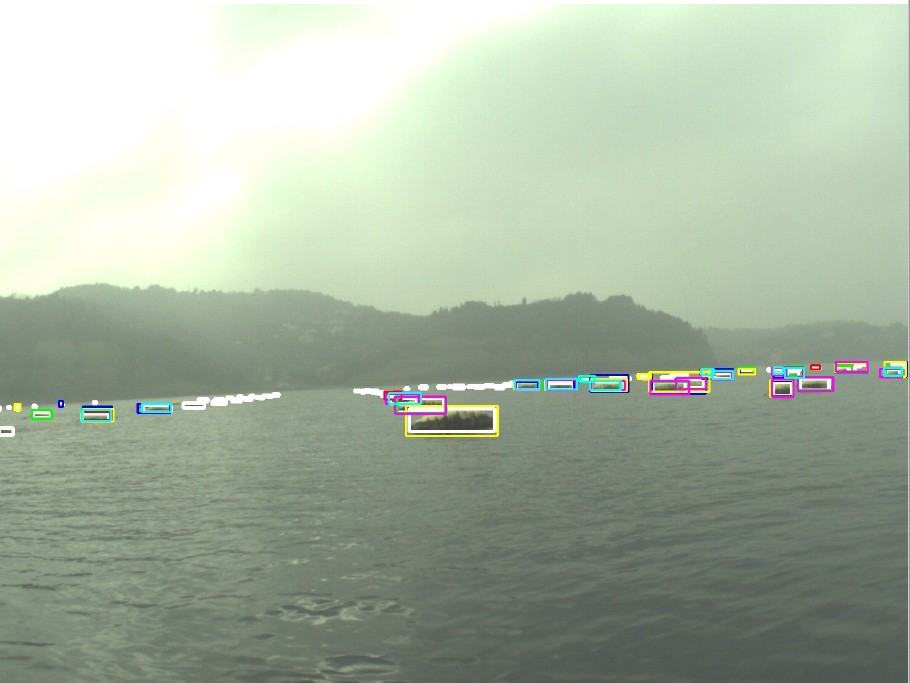}
        &
        \includegraphics[width=0.33\textwidth]{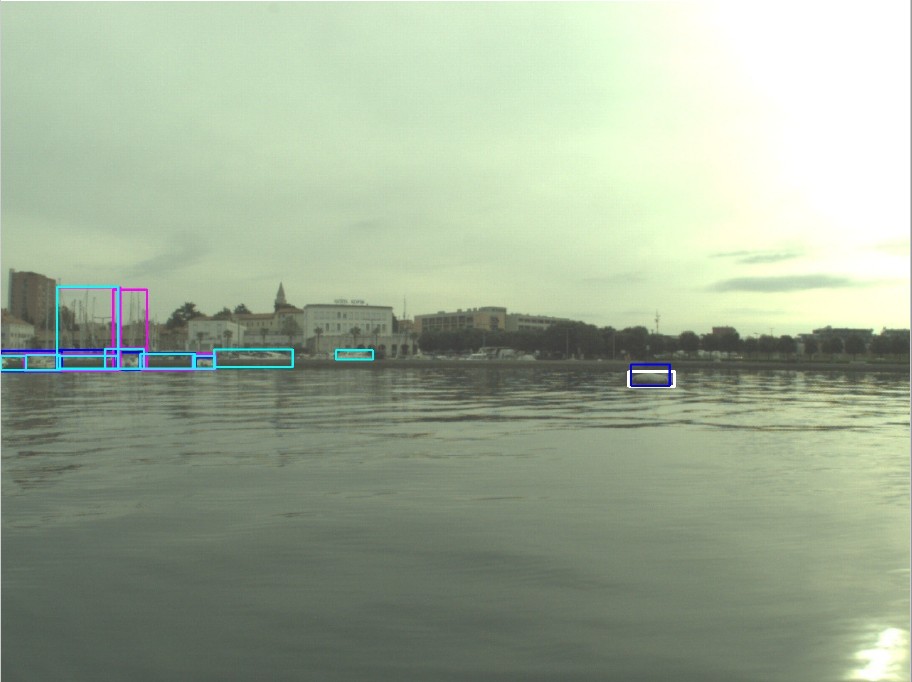}
        &
        \includegraphics[width=0.33\textwidth]{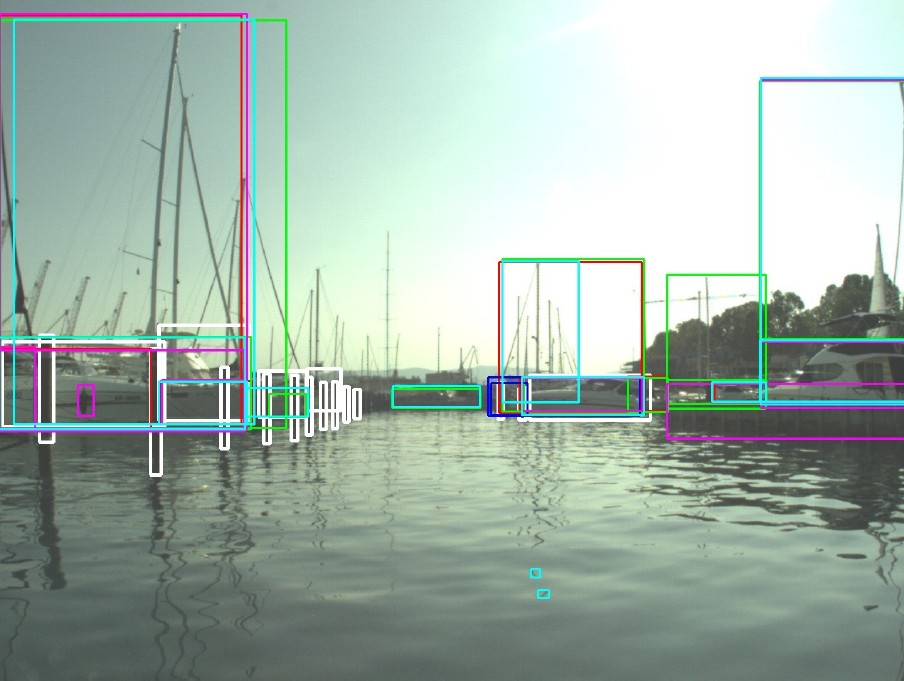}\\
        \includegraphics[width=0.33\textwidth]{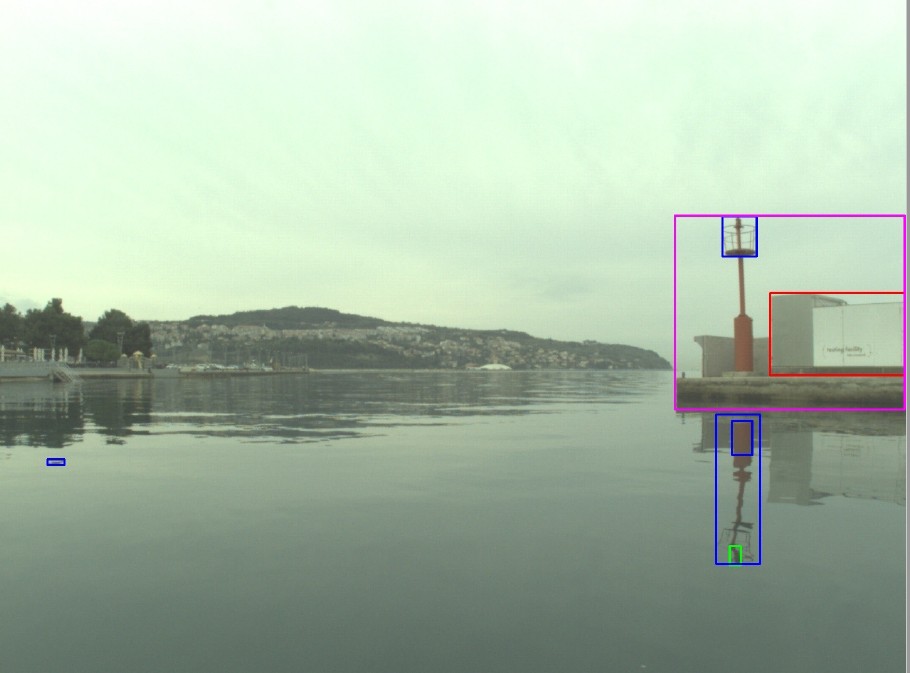}
        &
        \includegraphics[width=0.33\textwidth]{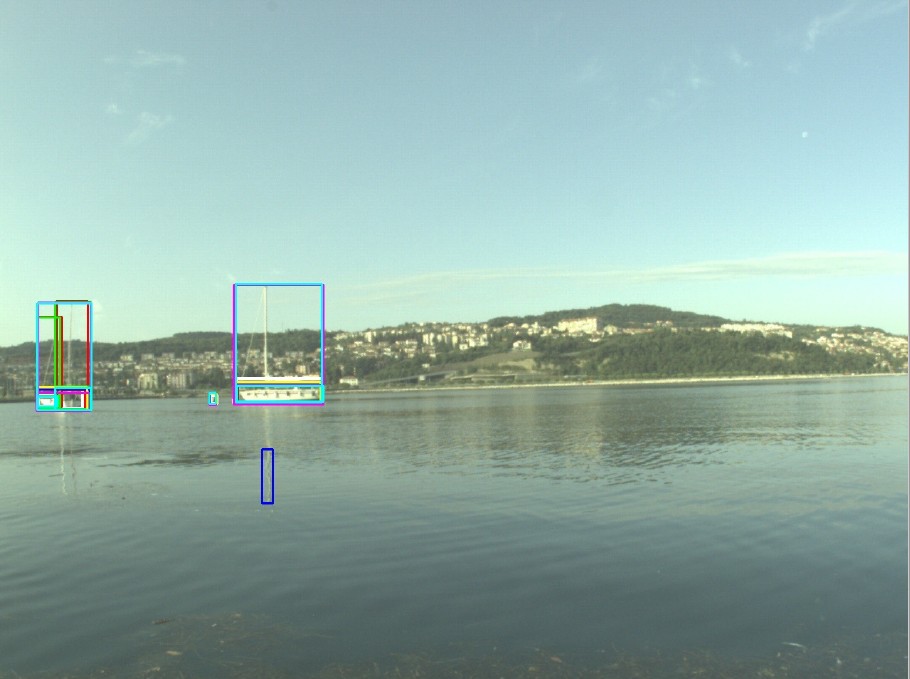}
        &
        \includegraphics[width=0.33\textwidth]{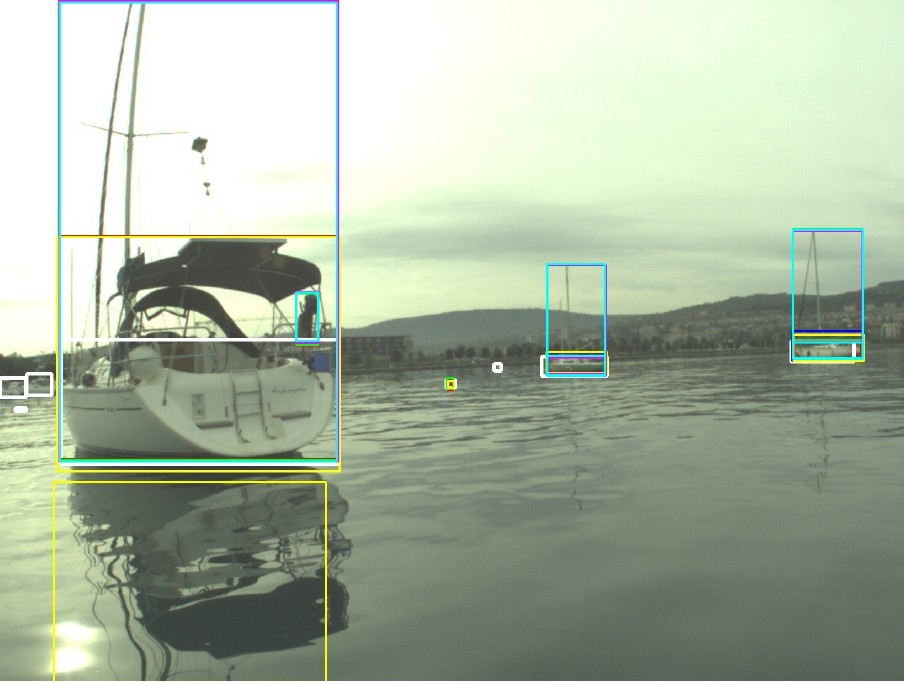}\\  
        \includegraphics[width=0.33\textwidth]{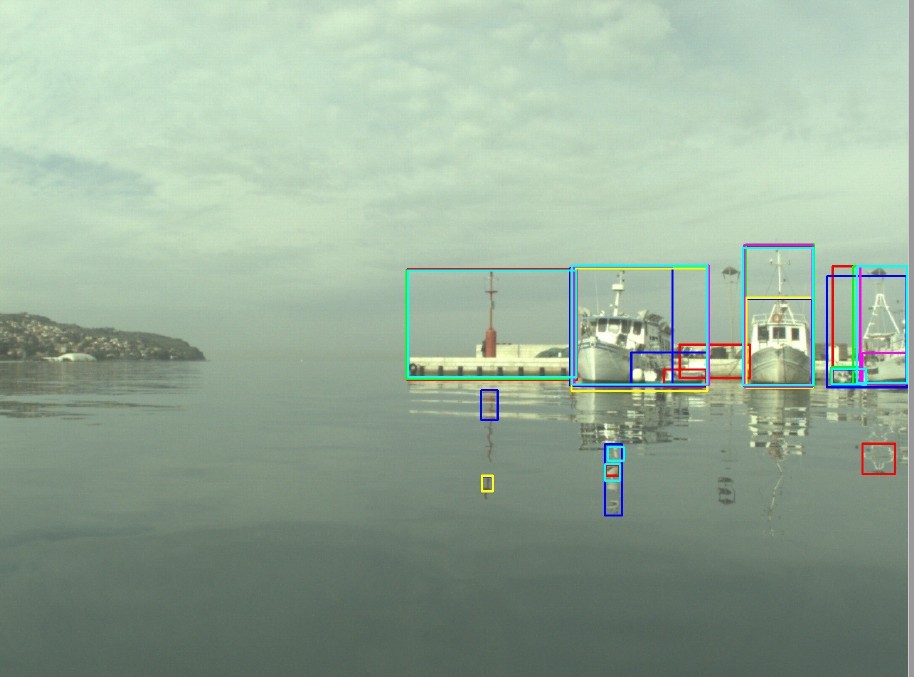}
        &
        \includegraphics[width=0.33\textwidth]{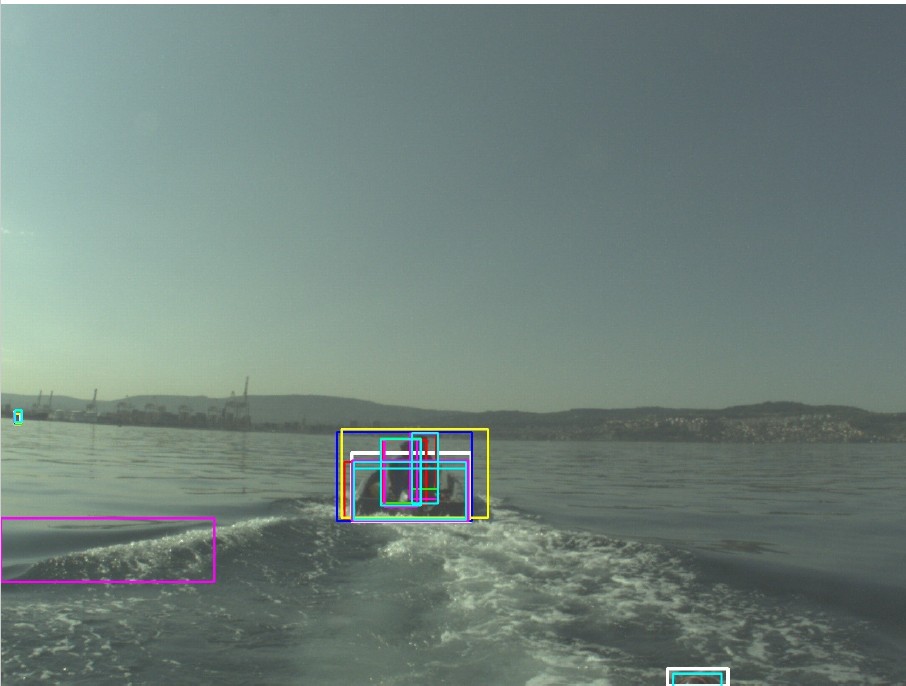}
        &
    \end{tabular}
    \caption{Qualitative comparison of methods for USV obstacle detection. Results of the individual methods are shown with colored bounding boxes as follows: white = ground truth, blue = DetectoRS, red = PRBNet, green = Nvlab x Acvlab's Yolo v7, yellow = FIOSB KA, cyan = Ocean U, magenta = Nvlab x Acvlab's PRBNet+Yolo v7.}
\label{fig:usvdet-qualitative}
\end{figure*}

Qualitative evaluation provides some further insighs into the performance of the methods, competing in the USV object detection challenge. The main conclusions are illustrated in Fig.~\ref{fig:usvdet-qualitative}.

In the first row of Fig.~\ref{fig:usvdet-qualitative} we can see the typical examples of failed detections (white ground truth bounding boxes with few or no detections). Small objects, such as faraway buoys are not detected by any of the methods. Low contrast objects are often detected only by DetectoRS. Finally, atypical objects (rarely found in object databases, in our example mooring posts) are missed by all methods, regardless of their apparent size.

In the second row of Fig.~\ref{fig:usvdet-qualitative} we can see the typical examples of false positive detections. Most often, these are reflections on the water surface, and most often, the algorithm that fails in this case is DetectoRS, which could be seen as the flip side of DetectoRS being able to detect low-constrast objects in marine environment.

The third row of Fig.~\ref{fig:usvdet-qualitative} shows further effect of reflections on the water surface in the first image, and the effect of waves (false positives) in the second image. Second image shows fragmentation of detection that plagues multiple models, but not DetectorRS, which may explain its good performance in the framework of IoU-based evaluation.

\subsubsection{Discussion and Challenge Winners}

Authors of the submitted methods tried to address various challenges of maritime obstacle detection, such as the large number of small objects and sensitivity to FP detections.
Overall, the winners of the USV object detection challenge are as follows:
\begin{description}
    \item[1\textsuperscript{st} place:]{Fraunhofer IOSB with DetectoRS,}
    \item[2\textsuperscript{nd} place:]{Nvlab x Acvlab with PRBNet,}
    \item[3\textsuperscript{rd} place:]{Ocean U. with their Ocean U. approach.}
\end{description}

In the analysis of the methods, we observed notable differences in the detection of the \emph{well known} object categories, which are included in many standard datasets (\eg person, boat) and more specialized objects, that can only be seen in the marine domain, such as mooring piers. This could be a natural consequence of the use of the common object detection datasets for training marine domain detection methods. This implies limited domain understanding of the environment and should be countered with emphasis on collecting visual data that contains a healthy proportion of marine-environment-only objects and obstacles.

\section{Conclusion}

% \janezcomment{I believe this is TODO for all the authors, overall conclusion to the paper?}
% \benjamincomment{Yes. Feel free to add stuff.}

In this summary work, we analyzed the challenges as part of the 1\textsuperscript{st} Workshop on Maritime Computer Vision. We looked at the aerial and surface domain and the challenges they pose. We worked out the advantages and limitations of submitted methods in these domains. For all challenge tracks, the need for real-time models has been raised and in future iterations, this could be a focus. 

Winners of the \emph{UAV-based Object Detection challenge} were (1\textsuperscript{st}) Maritime-VSA from the The University of Sydney, (2\textsuperscript{nd}) DetectoRS from Fraunhofer IOSB in Karlsruhe, and (3\textsuperscript{th}) YOLOv7-Sea from the Beijing University of Posts and Telecommunications. Each of these teams offered a distinct solution to the task, that is either transformer-based, a two-stage detector or a one-stage detector.

Winners of the \emph{UAV-based Multi-Object Tracking challenge} were (1\textsuperscript{st}) MoveSORT from the Beijing University of Posts and
Telecommunications, (2\textsuperscript{nd}) byteTracker from the National University of Defense Technology, and (3\textsuperscript{th}) StrongerSORT from EPFL. Each of these teams reaches top 3 performance with different backbones, but performances degrade for all methods if there are quick camera movements.

Winners of the \emph{USV Obstacle Segmentation Challenge} were (1\textsuperscript{st}) Multi-WaSR from Beijing University of Posts and Telecommunications, and (2\textsuperscript{nd}) MariFormer from Hong Kong University of Science and Technology. Both methods achieve similar overall performance, the former using an ensemble of weaker models, and the latter using transformers with a higher output resolution. The 2\textsuperscript{nd} placing method also demonstrates outstanding water-edge segmentation accuracy, however at a large cost in inference speed, which is critical for real-life deployment.

Winners of the \emph{USV Obstacle Detection Challenge} are (1\textsuperscript{st}) DetectoRS from Fraunhofer IOSB, (2\textsuperscript{nd}) PRBNet from Nvlab x Acvlab, and (3\textsuperscript{rd}) Ocean U. from Ocean U.
We observed notable differences in the detection of \emph{well known} object categories and maritime-specific objects, which could be a consequence of training with common object detection datasets. We thus believe more effort should be put into the collection and annotation of diverse maritime datasets.

% \janeze{In the USV domain we observed the notable difference in detection of the \emph{well known} object categories, which are included in many standard datasets (e.g. person, boat) and more specialized objects, that can be only seen in marine domain, such as mooring piers. This could be natural consequence of use of the common object detection datasets for training marine domain detection methods. This implies closed domain understanding of the environment and should be countered with emphasis on collecting visual dataset data that contains a healthy proportion of marine-enviroment-only objects and obstacles.}

Importantly, to obtain more significant challenge results, there needs to be a shift to sequestered test sets or at least hidden test set performances during submission phase. Lastly, the maritime domain brings up many related tasks and use-cases, such as maritime anomaly detection, which should be looked at in future iterations of MaCVi.

\vspace{.2cm}
\noindent\textbf{Acknowledgments.}
This work was supported by
% Fabio P.
the SHIELD project under the European Union’s Joint Programming Initiative – Cultural Heritage,
Conservation, Protection and Use joint call,
% Janez P.
Slovenian Research Agency (ARRS) project J2-2506 and programs P2-0214 and P2-0095,
% Benjamin
the German Ministry for Economic Affairs and Energy, Project Avalon, FKZ: 03SX481B and
Sentient Vision Systems for sponsoring prizes for the UAV-based Object Detection v2 challenge. Mr. Qiming Zhang, Mr. Yufei Xu, and Dr. Jing Zhang are supported by ARC FL-170100117.

% TODO: mention DAViMaR

\newpage
\appendix
\addcontentsline{toc}{section}{Submitted Methods}
\section*{Submitted Methods}

%detection technical reports

\section{UAV-based Detection}

\subsection{Maritime-VSA}
\label{trod:Maritime-VSA}
\emph{Qiming Zhang, Yufei Xu, Jing Zhang, Dacheng Tao}\\
\emph{\{qzha2506,yuxu7116\}@uni.sydney.edu.au,\\jing.zhang1@sydney.edu.au, dacheng.tao@gmail.com}\\
\\

This technical report describes the solutions to the MaCVi Object Detection v2 Challenge. Our team is with the `USYD' Institution. We obtain \textbf{the first place} in the leaderboards of both tracks, i.e., 61.52 mAP in \textbf{Object Detection v2} and 55.83 mAP in \textbf{Binary Object Detection v2}, and outperform the second participant by 1.9 mAP and 1.0 mAP in the two tracks, respectively. We use a single model without model ensemble. This technical report introduces our solutions to the challenge in detail.

\noindent \textbf{Overall architecture: }
The model architecture is based on our recent work \href{https://github.com/ViTAE-Transformer/ViTAE-VSA}{VSA} \cite{zhang2022vsa} with several backbone augmentations, i.e., \href{https://github.com/VDIGPKU/CBNetV2}{CBNetv2} \cite{liang2022cbnet} with \href{https://github.com/SwinTransformer/Swin-Transformer-Object-Detection}{Swin Transformer} \cite{liu2021swin}. Specifically, we use varied-size window attention (VSA) for the attention in Vision Transformer and select DB-Swin-S in CBNetv2 as the base model, which uses two Swin-S models in sequential to enhance the feature representations. It should be noted that the hand-crafted fixed window design in current works \cite{liu2021swin,zhang2022vitaev2} restricts the model's capacity to model long-term dependencies and adapt to objects of different sizes. VSA is better at processing images with large resolutions. Regarding the image resolutions in the SeaDronesSee v2 dataset are 3840x2160, 5456x3632, and 1229x934, our proposed VSA is suitable in this case. It can adapt the windows to various resolutions for the detection task in SeaDronesSee v2 by learning the window scales and shifts as adaptive window configurations from data and conducting self-attention within the learned windows. It can thus learn large window scales from high-resolution images in SeaDronesSee v2, model long-term dependencies, capture rich context from diverse windows, and extract better feature representations to improve detection performance. Besides, it is an easy-to-implementation module with minor modifications and negligible extra computational cost for window attention while improving the performance by a large margin. We use the popular Cascade R-CNN as the detection head.
We obtain 61.2 mAP for DB-Swin-S and 61.7 mAP for DB-Swin-S-VSA on the validation set in SeaDronesSee v2. DB-Swin-S-VSA obtains 60.62 mAP and 55.17 mAP on the test set in SeaDronesSee v2 and Binary SeaDroneSee v2, respectively. With test time augmentation (TTA), the results further increase to 61.52 mAP and 55.83 mAP.

\noindent \textbf{Training methods: }
We use MMDetection Toolbox \cite{chen2019mmdetection} and the default training settings for COCO detection \cite{lin2014microsoft}, such as an Adam optimizer and image augmentation techniques like normalizing, resizing, and flipping. We calculate the $mean$ and $std$ values of the images based on the training set to normalize the inputs. After pretraining on the training sets in ImageNet-22k and MS COCO, the model is finetuned with SeaDronesSee v2 for 12 epochs, resulting in three datasets in total, as described in the leaderboard. We use NVIDIA A100 GPUs for the experiments, and the inference speed is roughly 1.5 images per second per GPU with batch size 1 and image resolutions of $3840\times2160$.

\subsection{DetectoRS}
\label{trod:DetectoRS}
\emph{Lars Sommer, Raphael Spraul}\\
\emph{\{lars.sommer, raphael.spraul\}@iosb.fraunhofer.de}\\
\\
To generate our detections, we used DetectoRS \cite{qiao2021detectors} with
Cascade R-CNN and ResNet-50. For initialization, we used
weights pre-trained on MS COCO. To account for small object dimensions, we set the “scales” parameter to 3, yielding
smaller anchor boxes. The ”ratios” parameter was set to 0.5,
0.7, 1.0, 1.4 and 2.0 to increase the number of anchor boxes
All other parameters remained unchanged. SGD was used
as optimizer with an initial learning rate of 0.02, a momentum of 0.9 and a weight decay of 0.0001. The model was
trained for 12 epochs.
We employed the SeaDronesSee Object Detection v2
train and validation set as training data. For images
with dimensions less than 3840x2160 pixels, we used
multiple scales (1920x1080, 2376x 1296, 2688x1512 and
3360x1890 pixels). Otherwise, we set the input scale to
3360x1890 pixels. For inference, we applied multiscale
testing (2688x1512, 3360x1890 and 4032x2268 pixels). We
considered all five classes during training and inference.
The implementation provided by MMDetection \cite{chen2019mmdetection} - an
open source object detection toolbox based on PyTorch –
was used to train our detector. We used 2 Tesla V100 GPUS
(CPU: Intel Xeon E5-2698 v4 @ 2.20GHz). The inference
speed of the detector was about 1 FPS.
We tried several other baselines. To avoid redundant information of adjacent frames, we reduced the number of
images (using every 2nd or every 3rd frame), which yield
slightly worse AP values. Using only the train set as training data, yielded clearly worse AP values.

\subsection{YOLOv7-Sea}
\label{trod:YOLOv7Sea}
\emph{Hangyue Zhao, Hongpu Zhang, Yanyun Zhao}\\
\emph{\{zhaohy21315, zhp, zyy\}@bupt.edu.cn}\\
\\
Our method is mainly based on YOLOv7 for improvement. The whole architecture consists of three parts. First, the ELAN backbone from YOLOv7 is employed to extract feature maps. To make the network better learn useful information, we introduce SimAM attention module. In this way, the key target features contained in the shallow network can be highlighted, the irrelevant information can be weakened, and the detection performance of the algorithm on small targets can be improved. Since the SeadroneSee dataset contains many very small instances, we add the predicted head to the neck and head parts. Finally, other effective techniques are employed to achieve better accuracy and robustness, including Test Time Augmentation (TTA) and Weighted Box Fusion (WBF).

Training:
Augmentation: Mosaic, Mixup

Dataset:
SeedroneSee dataset for training (only used trainset part);
Pretrained on COCO dataset.

Device:
NVIDIA Tesla V100 GPUs
Time: about 1 fps
\\

See a more thorough explanation in our paper \cite{Zhao_2023_WACV}.

\subsection{DyHead}
\label{trod:DyHead}
\emph{Jan Lukas Augustin}\\
\emph{augustin@hsu-hh.de}\\
\\
{\bf Method:} We chose to use the Dynamic Head \cite{dai2021dynamic} framework combined with a powerful Swin-L \cite{liu2021swin} backbone. Considering the high amount of small objects in the SeaDronesSee
dataset, Dynamic Head seemed promising due to its scale-awareness and the excellent results in terms of APS on
the COCO test-dev dataset. MMDetection \cite{chen2019mmdetection} served as
a powerful toolbox to modify proven pipelines and tune
pretrained models and backbones.\\
{\bf Backbone:} Swin-L pretrained on ImageNet22k 384x384 \\
{\bf Neck:} Feature Pyramid Network \cite{lin2017feature} (3 scales)\\
{\bf Head:} Dynamic Head (6 blocks)\\
{\bf Box:} Adaptive Training Sample Selection \cite{zhang2020bridging}\\
{\bf Training:}\\
{\bf Optimizer:} AdamW, learning rate 5e-05, decay 0.05\\
{\bf Schedule:} 7 epochs, learning rate steps at epochs 5 and 7\\
{\bf Augmentations:} Multiscale resize in range 1400 to 2000\\
{\bf Datasets used:}\\
{\bf Backbone pretraining:} ImageNet22k\\
{\bf Model pretraining:} COCO 2017\\
{\bf Finetuning:} SeaDronesSee train
\begin{tightitemize}
    \item The best submission was trained with the all model parameters being unfrozen. Partially frozen experiments showed similar but slightly inferior results.
    \item The SeaDronesSee dataset was left unchanged training only on the train split assuming training on the validation set would be against challenge rules.
\end{tightitemize}

{\bf Hardware:} \\
We used a A100 40GB for training to allow for a powerful
backbone and larger training input size. Test time augmentations were used to improve prediction performance
on the test set at the cost of inference speed. Augmentations of the best submission included multiple scales
(4096, 2048 and 1280 pixels) and horizontal flipping at
each scale. Inference time including test time augmentations was 4.83s per image. The gains compared to a single
forward pass are marginal (0.55 vs. 0.57 AP) and so in
practice at an input size such as 2000 pixels would be used
and result in an inference time of 0.38s.\\
{\bf Adaptations considered:}
\begin{tightitemize}
    \item Augmentations such as color jitter to improve robustness to different light conditions were considered but not evaluated.
    \item Giving the model meta data information to improve scale-awareness was considered but not implemented due to time constraints.
\end{tightitemize}
{\bf Observations:}
\begin{tightitemize}
    \item Wrong annotations in the dataset were noticed but left
unchanged assuming changing them or leaving them out
would be against challenge rules. Analyses on the validation set suggest that cleaning the dataset may have
helped significantly. This assumption is based on the
observation that the model learned to predict bounding
boxes with offsets resembling the offsets of misplaced
annotations. Early stopping mitigated the problem at the
cost of poorer classification performance.
    \item Increasing the input size played a significant role. This
way even the model pretrained on COCO was already
able to detect tiny ships on the horizon in the largest
images. Only replacing the classification head did not
work better than tuning the entire model. Boxes were
well placed and scaled, but bright colors would always
be linked to all bright classes (life vests, jetskis, buoys).
It may be a good option for the binary case, especially
when the dataset is smaller.
\end{tightitemize}

\subsection{YOLOv7-X}
\label{trod:YOLOv7X}
\emph{Eui-ik Jeon, Impyeong Lee}\\
\emph{\{euiik0323,iplee\}@uos.ac.kr}\\
\\
We used 6 models provided by the official YOLOv7 Github \cite{yolov7github}. The hyperparameter was hyp.scratch.p5.yaml provided by yolov7. In the learning process, weights pretrained with cocodataset were used, and at this time, the optimization algorithm and data augmentation used ADAM, flip left-right, mosaic, mixup, and paste-in, respectively. We thought a flip up-down would not be necessary for data augmentation, but we now believe that that idea is wrong. We used the given object detection v2 dataset without change, and no additional dataset was used. In fact, we recently started
a research project related to search and rescue. So, at the end of September, I found out that this challenge was going on in the process of investigating prior research. Unfortunately, we did not consider using other datasets due to lack of time. In
the paper of YOLOv7, YOLOv7-E6E had 151.7M parameters, and mAP50 was the highest at 74.4. We thought that small object detection was important in this challenge. So, in order to enlarge the size of the feature map, the size of the image was simply enlarged rather than changing the structure of the model. As a result of changing the image size from 640 to 1920, it was found that mAP50 continuously increased. However, the image size change experiment was applied up to 1920 only in the basic model of YOLOv7 and X due to the limitations of the GPU, and up to 1600 in the rest of the models. As a
result, YOLOv7-X with image size set to 1920 showed slightly higher mAP than YOLOv7-E6E with image size 1600. And we did an experiment using YOLOv5. YOLOv5 provides 10 models according to the backbone structure. In YOLOv5, we experimented with the same input image size (640x640) as a model with the same bottleneck structure (eg s, s6). As a result, s6, where the size of the feature map becomes smaller due to a deeper backbone, showed lower accuracy. The reason we did not use YOLOV5 for the challenge is that the mAP of the validation dataset was lower than that of YOLOv7. Finally we used
NVIDIA’s RTX 3090 24GB, YOLOv7-X consumed 15 fps to inference one picture.

\subsection{YOLO-CNS}
\label{trod:YOLOCNS}
\emph{Luca Zedda, Andrea Loddo, Cecilia Di Ruberto}\\
\emph{l.zedda12@studenti.unica.it, \{andrea.loddo,dirubert\}@unica.it}\\
\\
For this challenge, we propose a novel and innovative architecture based on YOLOv5, YOLO-CNS. It
stands for You Only Look Once CBAM NAM SwinTransformer and has the following characteristics:
\begin{tightitemize}
    \item the architecture’s neck and backbone contain several Convolution Block Attention Modules (CBAM)
    \item the features of the last C3 module of each head with a set of 3 sequential Swin transformer blocks were
merged to create a custom set of heads
    \item the final layer is a Normalization-based Attention Module (NAM), projected to give more importance
to the best features.
\end{tightitemize}
Because of the large amount of small objects in the challenge dataset, a YOLO head specialized for small
objects has been employed. It work with features retrieved from the first layers of the backbone.\\
The hyperparameters employed are described as follows: number of epochs: 150; input image size 1280 ×1280 pixels; IOU threshold: 0.2; confidence-threshold 0.01.\\
The remaining hyperparameters were left as defaults and are those defined by the authors of YOLOv5.
They can be found at \cite{yolov5github}.\\
The model architecture, pretrained on the COCO dataset, was trained on the SeaDronesSee Object Detection v2 Dataset. All the experiments have been conducted on the same machine with the following configuration: Intel(R) Xeon(R) Gold 6136 CPU @ 3.00GHz CPU and Tesla P6 16 GB GPU.\\
Our team has studied a similar model for a malaria parasite detection task \cite{zedda2022deep}, which shares many technical difficulties with this challenge’s track, such as the presence of tiny objects. Our different submissions are related to the current epoch of the training process. We decided to validate our model every 25/30 epochs
to recognize possible overfitting issues or incremental improvements of the model.

\subsection{YOLOv7-W6}
\label{trod:YOLOv7W6}
\emph{Sagar Verma, Siddharth Gupta}\\
\emph{\{sagar, sid\}@granular.ai}\\
\\
{\bf 1. Solution}\\
This submission is from YOLOv7-W6 \cite{wang2022yolov7} network
trained directly on the training set of the dataset. We use the
network as it is. During training, we used images with an input size of 1280 while maintaining the aspect ratio. We train
the network on 4xV100 NVIDIA GPUs using PyTorch data
parallelism. We manage our experiments on GeoEngine
platform \cite{verma2022geoengine,9884440}.\\
An initial learning rate of 0.01, a momentum of 0.937,
and a weight decay of 0.0005 have been used. The following gain parameters in the loss function have been used:
box loss gain is 0.05, class loss gain is 0.3, object loss gain
is 0.7, IoU threshold is 0.20, and anchor multiple threshold is 4.0. Following augmentations have been used: HSV-
Hue (0.015), HSV-Saturation (0.7), HSV-Value (0.4), rotation (+/- 0.25 degrees), translate (+/- 0.2), scale (+/- 0.5),
shear (+/- 0.1), horizontal flips (0.1 probability), mixup
(0.1), copy paste (0.1) and paste in (0.1).\\
We also tried training on the synthetic dataset and then
fine-tuning it on the real dataset. We observed that this did
not improve results that much. In the later analysis, we
found out that classes are balanced in the synthetic dataset
and imbalanced in the real dataset. Also, the synthetic
dataset has a huge variance in object sizes for a given class.
For inference, we used a single RTX 3090 GPU and used
input image size to be 2560 while maintaining the aspect
ratio. We infer one image at a time and found that 10 images can be processed in 1 second. This can be sped up by using batch during inference.\\
{\bf 2. Solution}\\
The main problem we found with the above submission
is that YOLOv7-W6 is quite big and unsuitable for search
and rescue applications. It is not a practical solution if the
search and rescue will be performed in a remote area and the
model requires a huge GPU or a network connection to a remote GPU server. Also, if search and rescue happen over a large lake or ocean, the model will see the ocean most of the time. Using this information, we can use smaller networks
like YOLOv7-Tiny to detect the presence of an object, and
if something exists, then we can use EfficientNet-B0 to classify the object (extracted patch) into one of the five classes.\\
We trained YOLOv7-Tiny using the same strategy as
used for YOLOv7-W6. We also extract 256x256 patches
for all the objects such that the object is in the center of
the image and train an EfficientNet-B0 classifier. We then
join both networks and create a two-stage pipeline in which
the classifier is triggered only one YOLOv7-Tiny detects
objects. This strategy is good in terms of speed (88 FPS)
but not accuracy. We found out that the binary object detection performance was good, but the classifier was confused due to class imbalance. We tried to solve this problem using copy-paste, mix-up augmentations, and label smoothing but did not significantly improve results. We believe this could work if done correctly, which we failed to manage as the deadline was almost there.

\subsection{M10}
\label{trod:M10}
\emph{Shishir Muralidhara, Niharika Hegde}\\
\emph{\{s\_muralidhara21, n\_hegde21\}@cs.uni-kl.de}\\
\\
In our work, VarifocalNet is used as the architecture, with ResNeXt-101 as the backbone. We use the model provided by PyTorch pretrained on ImageNet, without any additional datasets or explicit pretraining. VarifocalNet is a single-stage object detector, with a custom varifocal loss which treats the positive examples and the negative examples asymmetrically, that is, a higher emphasis is placed on positive instances. This is significant in case of aerial images which are large and suffer from sparseness resulting in an imbalance between the positive and negative instances. Another challenge associated with large images is that the defining characteristics of objects are often indiscernible since the objects of interest are extremely small. Addressing this, we experimented with different scales of input data.\\
All the models were trained with identical hyperparameters and for 24 epochs on a single RTX3090 GPU. First, we set the baseline with the model trained on the full-sized raw image, using the train-val split. We also experimented with a tile-based approach, where each image is split into tiles of 512x512 by a sliding window approach with overlap and only retaining tiles containing a positive instance of an object. The tile-based approach ensures that representations of an object are feature rich. We observed that using tiles in conjunction with multi-scale representation of the full-sized image improved the performance significantly. As part of data augmentation, we implement normalisation, random flipping and cropping. Following the tile-based approach used during training, we utilised Slicing Aided Hyper Inference (SAHI) for testing, which again divides the image into tiles and performs object detection in each tile. We also tested the model using the full image at multiple scales and with test-time augmentation. The latter improved the results, achieving our best mAP of 52.54, with an inference speed of one image/second. 

\subsection{YOLOv7-NYU}
\label{trod:YOLOv7NYU}
\emph{Daitao Xing, Nikolaos Evangeliou, Anthony Tzes}\\
\emph{\{daitao.xing, nikolaos.evangeliou, anthony.tzes\}@nyu.edu}\\
\\
In this challenge, our team implemented a YOLO-based
object detection method. Specifically, we employ the recently released YOLOv7 \cite{wang2022yolov7} as our main detector, considering its good balance on accuracy and inference speed.
The YOLOv7-X with modified E-ELAN network has 120 stacked layers and predicts bounding boxes on P3 to P5 layers, which correspond to 8 1 to 32
feature maps. The SeaDroneSee \cite{varga2022seadronessee} dataset consists of images with ultra-high resolution and tiny bounding boxes annotations from swimmers
and life-saving applications. However, YOLOv7-X takes
a resized image of 640 × 640 as input in default settings,
which is inadequate for object detection in 4K images. To
address this problem, we modified the anchor size based
on YOLOv7-tiny settings. During training, we randomly
crop the patches of size 640 × 640 from the input images.
We also employ the crop-and-paste method to increase the
number of instances in patches. Other augmentations including color jitter, random rotation, scaling, flip, and mosaic are applied to improve the model’s robustness. We only use the provided dataset for training and validation purposes
without pre-training on the additional dataset. The network is
trained on a server with 4 GPUs for 300 epochs. The batch
size is set to 32 and the learning rate is 0.01. During inference, we use SAHI \cite{akyon2022slicing} to slice the input into overlapped
patches of size 640. After slicing, the patches are fed into
the network in a batch way. We noticed that even though
the patches are extracted for best inference preference, the
objects are still too tiny to be detected especially when the
drones are hovering at high altitudes. So during inference,
we apply an off-the-shelf super-resolution network \cite{ledig2017photo} to
boost the image qualities. Specifically, for bounding boxes
smaller than 200 pixels, we first cropped patches of size
60 × 60 centered at the bounding boxes. We then apply the
super-resolution network on those patches and fed the output into the detector again. We observe a significant performance jump (about 1.3\% point on AR score) after applying SR techniques. The performance can be further improved
if the super-resolution network is fine-tuned on the training dataset. Overall, the inference speed on a 2060 GPU is around 1 FpS and the best AP score is 0.5193.

\subsection{YOLOv7-FIT}
\label{trod:YOLOv7FIT}
\emph{Vojtěch Bartl, Jakub Špaňhel, Adam Herout}\\
\emph{\{ibartl, ispanhel, herout\}@fit.vutbr.cz}\\
\\
We participated at 1st Workshop on Maritime Computer Vision (MaCVi) in task Object Detection v2. The task was to detect objects in sea drones images. All our experiments were done on a personal computer with the following setup:
\begin{tightitemize}
    \item System: Ubuntu 20.04.5
    \item CPU: Intel Core i7-11700K
    \item GPU: Nvidia RTX3090
    \item RAM: 128 GB
\end{tightitemize}
We experimented with 4 different methods and achieved results as described in Table \ref{tab:YOLOv7FIT}.
\begin{table}[]
\centering
\begin{tabular}{p{1.7cm}ccccc}
\toprule
                \bf Method & AP & AP$_{50}$ & AP$_{75}$  & AR$_1$ & AR$_{10}$    \\
\midrule
                 YOLOv7 & 0.517 & 0.801 & 0.551 & 0.421 & 0.580 \\
                 TOOD & 0.471 & 0.732 &  0.492 & 0.396 & 0.536 \\
                 Pix2seq & 0.419 & 0.749 & 0.406 & 0.394 & 0.538 \\
                 DETR & 0.350 & 0.676 & 0.334 & 0.321 & 0.439 \\

\bottomrule
\end{tabular}
\caption{Results of tested methods.}
\label{tab:YOLOv7FIT}
\end{table}
In all cases we used models pretrained on COCO dataset
and fine-tuned them on SeaDronesSee dataset provided by
the challenge authors. If not mentioned all training parameters were same as in training scripts provided in relevant
repositories. Only YOLOv7 was able to run about $\approx$10
FPS; all other methods run about $\approx$1.5 FPS.\\
{\bf YOLOv7:}
We used YOLOv7 repository \cite{wang2022yolov7} with prepared fine-tuning
scripts — our variant was YOLOv7-E6 with image size
1280 × 1280, batch size 4, and training for 300 epochs.\\
{\bf TOOD:}
Another tested method was TOOD \cite{feng2021tood}. We used implementation provided in MMDetection toolbox \cite{chen2019mmdetection} — our variant was ResNet101 backbone with DCNv2 (R-101-dcnv2).
Model was learned for 24 epochs with batch size 6.\\
{\bf Pix2seq:}
Next tested method was Pix2seq \cite{chen2021pix2seq} which also provide prepared scripts for fine-tuning — we used ResNet50 backbone
with image size 1333 × 1333. The model was trained for 40
epochs with batch size 4.\\
{\bf DETR:}
We also tried DETR \cite{carion2020end} model. Similarly to TOOD we used
implementation provided in MMDetection toolbox. Backbone was ResNet50 and model was trained for 80 epochs with batch size 6.\\
{\bf Observations:}
We tried to fine-tune two “classical” CNN models
(YOLOv7, TOOD) and also two models based on transformers (Pix2seq, DETR) on SeaDronesSee dataset. Our main observation is that “classical” CNN models still reach comparable results and transformers do not provide significant result improvement.

\subsection{DurObj}
\label{trod:DurObj}
\emph{Neelanjan Bhowmik, Toby P. Breckon}\\
\emph{\{neelanjan.bhowmik, toby.breckon\}@durham.ac.uk}\\
\\
In the context of object detection tasks, most efforts have
focused on detecting objects-of-interest in standard colour
imagery by using Convolutional Neural Networks based object detection architecture of diverse characteristics such as, singe-stage \cite{redmon2018yolov3,feng2021tood}, multi-stage \cite{he2017mask,cai2019cascade}, and transformer-based \cite{carion2020end,zhu2020deformable}. In this challenge, we employ a transfer learning approach where the object detector (TOOD \cite{feng2021tood}) is initialised with task-specific pre-trained weight to obtain the knowledge and transferred to the target domain (e.g. SeaDronesSee v2 dataset).\\
{\bf TOOD \cite{feng2021tood}}. Recent one-stage detectors \cite{redmon2017yolo9000} predict two separate outputs by deploying two sub-networks to deal
with two sub-tasks, classification and localisation respectively. However, there is a degree of misalignment when
two separate branches are used to make predictions. To
overcome this misalignment, a sample assignment scheme
and a task-aligned loss function are designed in Task-aligned One-stage Object Detection (TOOD) by explicitly
aligning the two tasks (object classification and localisation) in a learning-based manner by utilising novel task-
aligned head (T-Head) which offers a better balance between learning task-interactive and task-specific features and task alignment learning via a designed sample assignment scheme and a task-aligned loss.\\
Model Initialisation. The object detection method relies
heavily on architectures that have been trained on large-scale colour imagery datasets. We incorporate a task-specific model initialisation or transfer learning strategy in
this challenge. In our prior work of \cite{bhowmik2022lost}, we analysed the impact of object area on the performance of object detectors.
As the target dataset (SeaDronesSee v2) primarily consists
of aerial imagery (air-to-ground) with small-size object area
(object-area $<$ 32 × 32 pixel \cite{bhowmik2022lost}), we initialised our model
with pre-trained on VisDrone \cite{visdroneworkshop} (aerial imagery dataset)
instead of the commonly used ImageNet \cite{deng2009imagenet} or MS-COCO
\cite{lin2014microsoft} datasets.\\
Implementation Details. TOOD is implemented using the MMDetection framework \cite{chen2019mmdetection} and initialised with weights pre-trained on the VisDrone dataset. Our model is trained only on the train split of SeaDronesSee
v2 dataset using a ResNet 101 \cite{he2016deep} backbone with the following training configuration: backpropagation optimisation is performed via Stochastic Gradient Descent (SGD), with initial learning rates of $10^{-2}$, trained for 40 epochs. Standard data augmentation techniques, such as Random Crop, Random Flip, have applied during model training with an application probability of 0.5. The model inference is carried out on the NVIDIA TITAN RTX GPU, with an achieved
inference speed of 4 FPS.

\subsection{APX}
\label{trod:APX}
\emph{Shivanand Kundargi*, Tejas Anvekar*, Ramesh Ashok Tabib, Uma Mudengudi (*equal contribution)}\\
\emph{\{shivanandkundargi992,anvekartejas\}@gmail.com, \{ramesh\_t, uma\}@kletech.ac.in}\\
\\
At sea, rescue operations are carried out to save stranded sailors and passengers, as well as survivors of crashed aircraft. Climate change and migration across major oceans have increased the importance of maritime search and rescue. In this report, we propose \textbf{APX}: Adaptive Pixel Clustering for seamless marine object detection for quicker convergence and a faster train-to-deployment pipeline for marine drones.
\begin{figure}[h]
    \centering
    \includegraphics[width=1\linewidth]{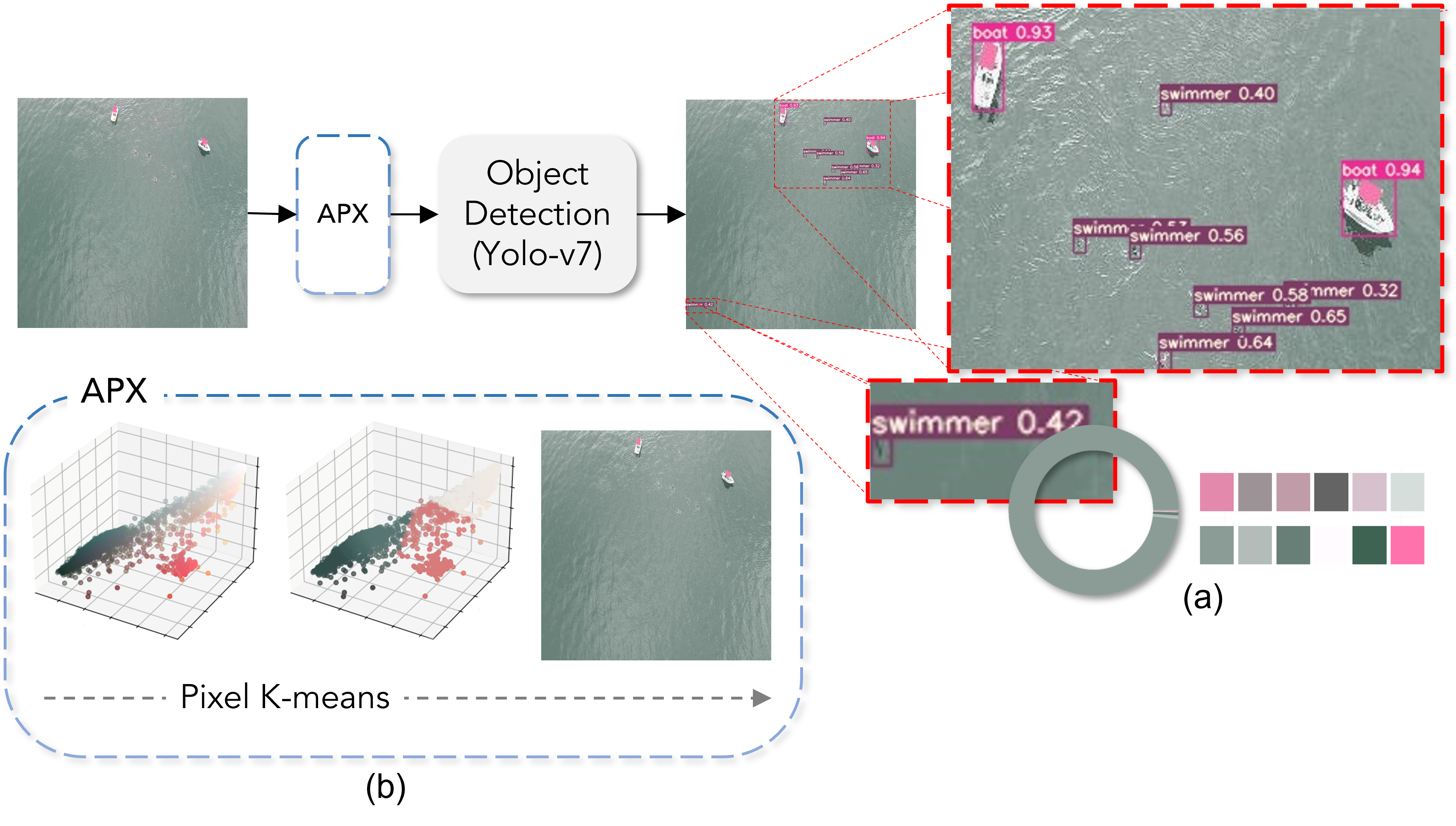}
    \caption{\textbf{APX} is an immanent method to perform Adaptive Pixel Clustering for seamless marine object detection. \textbf{(a)} depicts Most Dominant colour distribution of a sample image in SeaDronesSee Object Detection v2 Dataset. \textbf{(b)} depicts Adaptive Pixel Clustering with Elbow-method for setting no of cluster \textbf{K}.}
    \label{trod:APX_img}
    \vspace{-5mm}
\end{figure}
\\
We observe that, majority of the colour-variance in the SeaDronesSee dataset are in blue-green tones, as the dataset is a drone-view image of oceans. Almost 80-90\% of image consists blue-green tones. An object-detection model can be aided by simply binning the colour variance such that true objects are implicitly emphasised. As shown in Figure \ref{trod:APX_img}, we propose \textbf{APX}: Adaptive Pixel Clustering to facilitate object-detection in Marine. Our method is adaptive as we use Kmeans \cite{KMEANS} + Elbow method to choose the no of dominant colours in a given image. We observe that for training set, Elbow method yields K = 28 $\pm$ 4 and we achieve best mAP on validation set for K = 32.\\
{\bf Experimental Setup:}
We train the proposed pipeline on compressed version of SeaDronesSee object detection v2 dataset, the resolution of the image is resized to 640 $\times$ 640 for quicker convergence, better inference time and resource management. We train the network on RTX-3090 for 50 epochs, batch-size: 32 and the other hyperparameters are the same as those listed in YoloV7 \cite{wang2022yolov7,yolov7github}.\\
{\bf Results and Discussions:}
The results on test set as per the leader-board of SeaDronesSee \& MODS challenge, we obtain the following results, AP: \textbf{0.4968}, AP50: 0.8326, AP75: 0.5004, AR1: 0.4131, AR10: 0.5797.\\
On the validation set, YoloV5 trained for 100 epochs achieves mAP @ 0.5:0.95 = 0.2998,
YoloV7 trained for 100 epochs achieves mAP @ 0.5:0.95 = 0.4632,
\textbf{Ours} (YoloV5 + APX) trained for 50 epochs achieves mAP @ 0.5:0.95 = \textbf{0.3122},
\textbf{Ours} (YoloV7 + APX) trained for 50 epochs achieves mAP @ 0.5:0.95 = \textbf{0.4808}.

\subsection{YOLOv7-TILE}
\label{trod:YOLOv7TILE}
\emph{Arpita Vats}\\
\emph{avats@scu.edu}\\
\\
The dataset provided was unbalanced, so the first
step was to balance the dataset using PyTorch library
\texttt{WeightedRandomSampler}, which is responsible for
making sure that the model sees the minority classes more
while training. We used different datasets provided for
training, validation, and testing We used high-resolution
uncompressed images for training and Inference. We executed all our experiments on a system running Ubuntu Linux version 20.04 and equipped with a 12-core Intel(R)
Core(TM) i9-7920X CPU @ 2.90GHz, 128 GB RAM, and
2 NVIDIA RTX 3090 24G GPUs. it took approximately
5-6 hours for training and around 2-3 hour for inference
However, while portions of our method take advantage of
multi-threaded CPU-based processing, our method uses two
GPUs for training and one GPU for inference. In our proposed model, we used the \texttt{yolov7} model as the backbone
for object detection. Since the actual object to be detected
was very tiny, we tried different approaches to make sure the
tiny objects are detected correctly. One of the approaches
that we considered is \texttt{Oriented-RCNN}, which proposes
an effective and simple oriented object detection framework, termed Oriented R-CNN,which is a general two-stage
oriented detector with promising accuracy and efficiency.
To be specific, in the first stage, they proposed an oriented
Region Proposal Network (oriented RPN) that directly generates high-quality-oriented proposals in a nearly cost-free
manner. The second stage is oriented R-CNN head for refining oriented Regions of Interest (oriented ROIs) and recognizing them \cite{xie2021oriented}, this model was not able to perform on the given dataset. \texttt{SAHI} \cite{akyon2022slicing} Inference model based on image tiling, which currently supports yolov7 and many others. \texttt{SAHI} provided us with motivation to try the approach
of Image Tiling \cite{ozge2019power}, which divides the image into tiles for
training and testing, so for tilling training images, it uses
the bounding box region given in the labels for training images for tilling and these tiled images are used for training
using Yolov7 pre-trained weights. For inference, we used
the testing images provided and tiled those testing images
to use them for inference, and after the inference is completed prediction on tile images is stitched back to the original images, to get the final prediction in the required format
(COCO format). We were able to achieve an AP of 0.42315.

\subsection{YOLOv7 Baseline}
\label{trod:yolov7baseline}
\emph{MaCVi Organizers}\\

We trained a pre-trained (on COCO) YOLOv7 model \cite{wang2022yolov7} for eight epochs on the SDS ODv2 train data set with default configurations.

\subsection{Faster R-CNN ResNet-18 Baseline}
\label{trod:rcnnbaseline}
\emph{MaCVi Organizers}\\

We trained a pre-trained (on COCO) Faster R-CNN model \cite{wang2022yolov7} with ResNet-18 backbone on the SDS ODv2 train data set with adapted anchor sizes.

\section{UAV-based Tracking}
%\appendix
%\section{Submitted Methods}

%MOT
\subsection{DeepSORT with ECC (MoveSORT)}
\label{tr:MoveSORT}
\emph{Yang Song, Delong Liu}\\
\emph{\{sy12138,liudelong\}@bupt.edu.cn}\\
\\
We propose the MoveSORT algorithm for Multiple
Object Tracking tasks in the maritime UAV view. The
MoveSORT algorithm makes use of some advanced modules and inference tricks to effectively improve the
DeepSORT \cite{wojke2017simple} algorithm.
Detector: DeepSORT uses the optimized Faster R- CNN
presented in as the detector. Instead, we replace the detector
with YOLOv7, which holds excellent trade-off performance
between accuracy and speed.\\
{\bf ECC:} Enhanced Correlation Coefficient Maximization
(ECC \cite{evangelidis2008parametric}) is a technique for parametric image alignment, which can estimate the global rotation and translation between adjacent frames. Motion scenes may cause the linear
Kalman filter of DeepSORT algorithm to fail, so we use
ECC to compensate for the non-linear variations caused by
the UAV camera.\\
{\bf IOU:} We use the IOU distance instead of the Mahalanobis distance in DeepSORT cascade matching and
choose a larger IOU threshold, resulting in cascade matching that focuses more on matching in terms of appearance
features and improving the robustness of algorithms in mobile scenes.\\
{\bf NSA Kalman:} The Kalman filter is susceptible to low-quality detection and ignores detection noise. To solve
this problem,we borrow the NSA Kalman algorithm from
GIAOTracker \cite{du2021giaotracker}, which adaptively calculates the noise covariance $\tilde{R}_k$ based on the detection confidence score:
\begin{equation*}
    \tilde{R}_k = (1-c_k)R_k
\end{equation*}
where $R_k$ is the preset constant measurement noise covariance and $c_k$ is the detection confidence score at state
k. Though its simplicity, it can help improve the accuracy of
updated states.
Our model was run on a Tesla T4 with FPS of 10
frames per second. We only use the official dataset and have
achieved excellent results.

\begin{table}[h]
\centering
\begin{tabular}{p{1.7cm}ccccc}
\toprule
                \bf Method & \bf ECC & \bf IOU & \bf NSA & \bf HOTA & \bf MOTA     \\
\midrule
                 DeepSORT & & & & 62.7 & 77.2 \\
                 MoveSORTv1 & \checkmark & & &63.6 & 78.3 \\
                 MoveSORTv2 & \checkmark & \checkmark & & 64.7 & 78.4 \\
                 MoveSORTv3 & \checkmark & \checkmark & \checkmark & \bf 66.6 & \bf 80.0  \\

\bottomrule
\end{tabular}
\caption{Ablation study of MoveSORT on the SeaDronesSee-MOT set. HOTA and MOTA in \%.}
\label{tab:MoveSORT}
\end{table}

\subsection{byteTracker}
\label{tr:byteTracker}
\emph{Yonglin Li, Shuman Li, Chenhao Tan, Long Lan}\\
\emph{\{liyonglin12345, lishuman13\}@nudt.edu.cn,\\2826554153@qq.com, long.lan@nudt.edu.cn,
}\\
\\
We use the ByteTrack as our method. Specially, the
YOLOX is selected as backbone network to detect and the
BYTE implements the association. In our method, we
remove the vertical bbox restrict in Byte to satisfy marine scene. Compared to the pedestrian tracking task in
MOT challenge \cite{MOT16}, we decrease detection threshold from 0.7
to 0.6 and tracking threshold from 0.6 to 0.5 to adapt
smaller and sparser objects. ByteTrack public code are re-leased at \cite{bytetrackrepo}. YOLOX
public code are released at \cite{yoloxrepo}.\\
{\bf Implementation Details:}
The detector is YOLOX with YOLOX-x as the backbone
and the COCO-pretrained model is introduced as the initialized weights. For SeaDronesSee MOT, the training schedule is 120 epochs on train dataset. The input size is 1440
× 800. The shortest side ranges from 576 to 1024 during
multi-scale training. The data augmentation we used includes Mosaic and Mixup. The used model is trained on 8
NVIDIA Tesla A100 GPU with batch size of 48. The optimizer is SGD with weight decay of $510^{-4}$ and momentum
of $0.9$. The initial learning rate is $10^{-3}$ with $1$ epoch warm-up and cosine annealing schedule.
In inference, the default detection score threshold is 0.5,
unless otherwise specified. We only use IoU as the similarity metrics. In the linear assignment step, if the IoU between
the detection box and the tracklet box is smaller than 0.2, we
reject the matching. For the lost tracklets, we keep it for
30 frames in case it appears again. FPS is measured with
FP16-precision and batch size of 1 on a single GPU. The
FPS on our machine is about 6 FPS.
Compared to the pedestrian tracking in MOT, SeaDronesSee MOT has more sparser scenes and smaller objects.
Based on it, We decrease detection threshold and tracking
threshold for adapting it. Meanwhile, as the overlap objects in the dataset are rarely, we increase nms threshold to
remove potential false association and decrease num of ID
Switch.

\subsection{StrongerSORT and STI-StrongSORT}
\label{tr:StrongerSORT}
\emph{Vladimir Somers, Christophe De Vleeschouwer, Alexandre Alahi
}\\
\emph{\{vladimir.somers, alexandre.alahi\}@epfl.ch,\\christophe.devleeschouwer@uclouvain.be
}\\
\\
To address the SeaDronesSee Multi-Object Tracking challenge, we focus on the data association problem and propose two methods using the provided public detections. 
We first detail both methods below and then provide some insights about their performance.

\textbf{StrongSORT with PCB+GiLt}: The first method is fully online and adapted from StrongSORT (implementation provided at \cite{strongsortgithub}) \cite{Du2022}, with the two offline modules GSI and AFLink removed.
For the ReID model, we employed a PCB \cite{PCB} part-based re-identification model, pre-trained on Market1501 \cite{market1501} with a GiLt (Github: \cite{giltloss}) loss \cite{bpbreid}.
This method achieves 10fps.

\textbf{Spatio-temporal StrongSORT with Interpolation}: 
On the SeaDronesSee MOT dataset, targets (boats and swimmers) have very similar appearance, occlusions are rare because of the aerial nature of the shots and tracking targets that go out and back in view is not necessary according to challenge rules. 
For these reasons, we argue spatio-temporal information is more valuable to solve the data association, compared to appearance information from a re-identification model.
Based on these observations, we introduce our second method that rely solely on spatio-temporal information.
We changed two components to the above method for our second offline method.
We first use a new two stage association step using only spatio-temporal information: (1) tracklets are matched with current detections using the IOU score with a minimum threshold of 0.1, and (2) remaining detections are matched according to their bbox centers distance with a max threshold of 60 pixels.
The goal of this second step is to address large displacements caused by fast camera motion from UAV rotations.
Finally, to address the false negatives that are predominant in the public detections, we adopt a post-processing strategy to linearly interpolate missing detections within each final tracklet.
We also delete tracklets that do not span at least 20 frames. 
This method runs at 30fps.

\textbf{Analysis}: Both methods ranked third in the final leaderboard.
The second method induces big improvement regarding all metrics related to data association, with a much lower fragmentation and five times decreased id switches: the vast majority of targets are now tracked from the beginning to the end of each video.
However, the interpolation module is more sensitive to false positive detections, which induces a drop in detection accuracy and therefore in MOTA and HOTA metrics.
Moreover, the bbox centers distance matching step is still sensitive to false positives/negatives and would strongly benefit from a better object detector.
Just like most other MOT works, there's a big detection/association accuracy trade-off between our two methods, the second one achieving better association.

\subsection{MOT}
\label{tr:MOT}
\emph{Lars Sommer, Raphael Spraul}\\
\emph{\{lars.sommer, raphael.spraul\}@iosb.fraunhofer.de}\\
\\
Our approach is comprised of an initial object detector
followed by multiple object tracker.
\\
{\bf Detector:}
To generate our detections, we used DetectoRS \cite{qiao2021detectors} with
Cascade R-CNN and ResNet-50. For initialization, we used
weights pre-trained on MS COCO. To account for small object dimensions, we set the “scales” parameter to 4, yielding
smaller anchor boxes. All other parameters remained unchanged. SGD was used as optimizer with an initial learning rate of 0.02, a momentum of 0.9 and a weight decay of
0.0001. The model was trained for 12 epochs.
We employed the SeaDronesSee Object Detection v2
train and validation set as training data. For images
with dimensions less than 3840x2160 pixels, we used
multiple scales (1920x1080, 2376x 1296, 2688x1512 and
3360x1890 pixels). Otherwise, we set the input scale to
3360x1890 pixels. For inference, we applied multiscale
testing (2688x1512, 3360x1890 and 4032x2268 pixels). We
considered all five classes during training and inference.
The implementation provided by MMDetection \cite{chen2019mmdetection} - an
open source object detection toolbox based on PyTorch –
was used to train our detector. We used 2 Tesla V100 GPUS
(CPU: Intel Xeon E5-2698 v4 @ 2.20GHz). The inference
speed of the detector was about 1 FPS.
\\
{\bf Tracker:}
For tracking we use the DeepSORT \cite{wojke2017simple} tracking approach. Since our detector output does not provide any appearance features for the individual boxes, we chose constant feature values as input for the tracker. Tracks were
only initialized on detection boxes with a confidence value
greater than 0.3. For updating tracks we considered every
detection with detection score greater then 0.0. Since we
didn’t expect large occlusions in the test scenarios, we set
the max age value to 1. max age is the maximum number
of consecutive misses before the track is deleted. The remaining parameters were left at the default settings. The inference speed of the tracker was about 500 FPS on an AMD
Ryzen 9 3900X @ 3.8GHz.

\subsection{OCSORT}
\label{tr:OCSORT}
\emph{Hsiang-Wei Huang, Cheng-Yen Yang, Jenq-Neng Hwang, Pyong-Kun Kim, Kwangju Kim, Kyoungoh Lee}\\
\emph{\{hwhuang, cycyang, hwang\}@uw.edu, \\ \{iros, kwangju, longweek7\}@etri.re.kr}\\
\\
We use YOLOX-XL as our detector, we dump all the five different classes into
one class and train the model for 20 epochs. The model is pretrained on the MaCVi
object detection dataset for 20 epochs and then finetune with MaCVi object
tracking training and validation dataset. The tracking algorithm we used is OCSORT.
\\
{\bf Device:}
We use V100 for model training and inference. The inference speed is roughly 20
FPS. \\
{\bf Adaptation:}
Due to the different drone heights of the testing video, the size of the target can
vary significantly. For several videos with lower height, we use the detector trained
on MaCVi object detection dataset to conduct object tracking. But the performance
is not improved significantly.
4. Result
We achieve 0.608 in HOTA, 0.724 in MOTA and 0.692 in IDF1.

\subsection{Tracktor Baseline}
\label{tr:Tracktor Baseline}
\emph{MaCVi Organizers}\\

The baseline we provided is a
Tracktor-based tracker using Camera Motion Compensation with a Faster R-CNN ResNet-50 detector. We used the tracking implementations from
mmdetection \cite{chen2019mmdetection} with default hyperparameters.

%USV-related technical reports
\section{USV-based Obstacle Segmentation}

\subsection{Multi-WaSR}
\label{usv-seg:multi-wasr}
\noindent
\emph{Shuai Jiang, Haiwen Li}\\
\texttt{\small{\{js, lihaiwen52\}@bupt.edu.cn,
}}\\
\emph{Beijing University of Posts and Telecommunications (BUPT)}\\

\begin{figure}[t]
    \centering
    \includegraphics[width=1\linewidth]{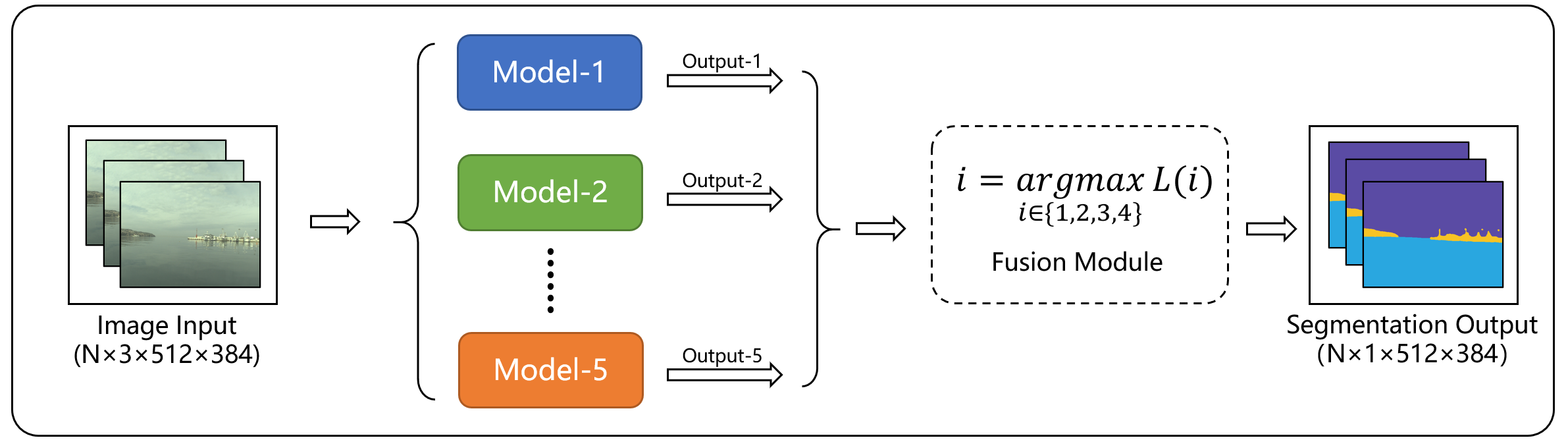}
    \caption{Multi-WaSR: The process for the fusion method.}
    \label{fig:techreports/multi-wasr}
\end{figure}

Inspired by~\cite{Bovcon2021}, we use the WaSR network as our base- line model. During the competition, we tried a total of four network architectures. Combining different data augmentation methods, we got five models and trained them separately. Model-1 and model-2 both use WaSR-Res101~\cite{Bovcon2021}, but their augmentation methods are different. Model-3 uses WaSR-Res50~\cite{Bovcon2021} as its architecture, while model-4 uses DeepLab~\cite{chen2017rethinking}. Model-5 modifies the WaSR framework to improve its effect on Water-edge segmentation. Specifically, we replace ARM1 and ARM2 with two Transformers~\cite{zamir2022restormer} named TRF1 and TRF2. Unlike the WaSR network which concatenates one channel of IMU feature, model-5 concatenates four channels of IMU feature. Each of these five models has its own strengths. Thus, we trained the above models separately and perform inference. Then we propose a module to fuse the segmentation outputs by the five models, which leads to our first ranking on the leaderboard. The fusion criterion is to vote on each pixel and the class with the highest votes is used as the classification result for that pixel. The models involved in our experiments are shown in Table~\ref{tab:techreports/multi-wasr} (Multi-WaSR shows the results after fusion). The fusion process is shown in Figure~\ref{fig:techreports/multi-wasr}.

\begin{table}[h]
\centering
\begin{tabular}{cccc}
\toprule
    \bf Model & \bf Backbone & \bf Ranking & \bf Avg. Score    \\
\midrule
    1    & WaSR-Res101     & \#       & 92.5      \\
    2    & WaSR-Res101     & 8th      & 92.3      \\
    3    & WaSR-Res50      & \#       & 88.8      \\
    4    & Deeplab         & 12th     & 92.0      \\
    5    & WaSR-TRF-Res101 & 14th     & 91.4      \\
\midrule
  \bf 6  & \bf Multi-WaSR  & \#       & 93.5      \\
\bottomrule
\end{tabular}
\caption{Multi-WaSR: The models correspond to the submissions in the leaderboard (\# represents not submitted to the leaderboard).}
\label{tab:techreports/multi-wasr}
\end{table}

The training dataset is MaSTr1325~\cite{bovcon2019mastr} and the test dataset is MODS~\cite{MODSBenchmark2022}. Considering that light spots reflected from sea level may be considered as false positives, we add two augmentation methods, namely Random-Brightness and Random-Contrast, which can also improve the generalization performance. We modify epoch to 50, the rest of the hyperparameters and the loss function are the same as in reference~\cite{Bovcon2021}. The device we take is a Tesla V100 with an inference speed of approximately 12 FPS.

% TODO: table

\subsection{MariFormer \& RevDeep}
\noindent
\emph{Zheng Ziqiang, Tuan-Anh Vu, Hai Nguyen-Truong, Sai-Kit Yeung}\\
\texttt{\small{\{zzhengaw,tavu,thnguyenab\}@connect.ust.hk, saikit@ust.hk}}\\
\emph{Hong Kong University of Science and Technology (HKUST)}\\

\subsubsection{MariFormer}\label{usv-seg:mari-former}
\textbf{Algorithm Outline}: The proposed maritime obstacle segmentation is based on the transformer-based neural architecture SegFormer~\cite{xie2021segformer}. SegFormer is a simple, efficient yet powerful semantic segmentation framework, which introduces Transformer architecture into the semantic segmentation. The lightweight multi-layer perception (MLP) decoders are utilized for generating high-quality segmentation map based on multi-scale feature representations from a novel hierarchically structured Transformer encoder. The MLP decoder could aggregate information from different layers to generate more powerful representations. Also, the SegFormer could support the precise semantic segmentation on the high resolution images.

We perform the semantic segmentation based on MMSegmentation~\cite{mmsegrepo}. The training configuration is ``segformer\_mit-b5\_8x1\_1024x1024\_80k''. The image resolution is set to $1024\times 1024$ and the training schedule is set to ``schedule\_80k''. The optimizer is SGD optimizer and learning rate is 0.01. The momentum is 0.9 and the weight decay is 0.0005. The data augmentation includes random crop and resize, random flipping, and normalization. 

We only use the MaSTr1325 dataset~\cite{bovcon2019mastr} for training and no other additional images are used. The training device is RTX 3090 with 24G memory. The inference speed is about 3.8 frames per second under the image resolution $1278\times 758$.

\textbf{Special processing}: For sequence ``kope100-00006790'', we remove the boundary according to the given mask as shown in following Figure~\ref{fig:techreports/mari-former} to remove the influence of the boundary.
\begin{figure}[t]
  \begin{center}
    \includegraphics[width=\linewidth]{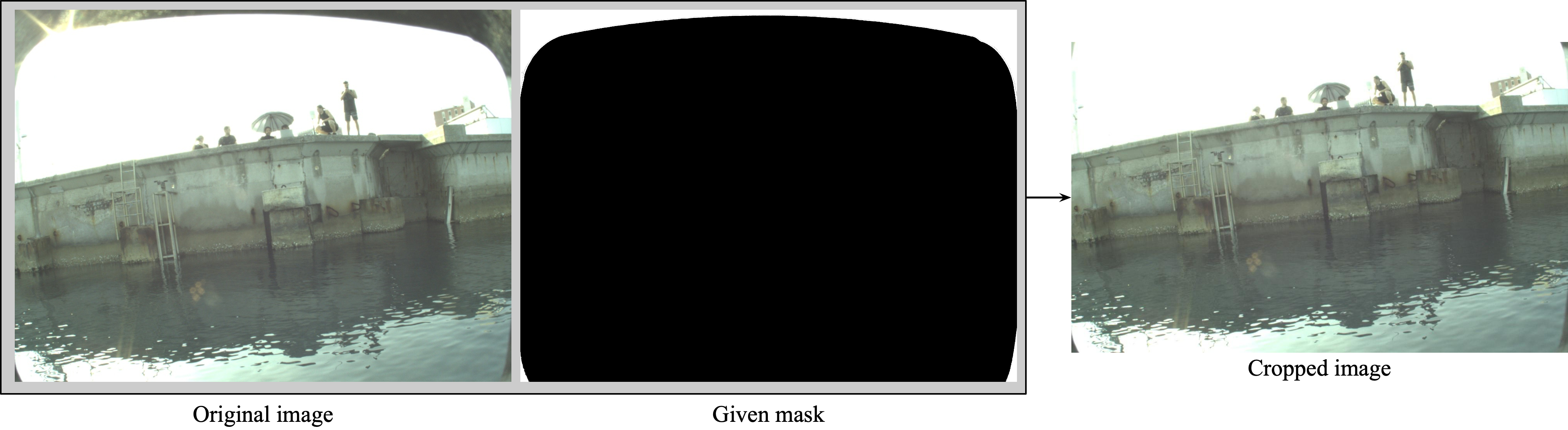}
  \end{center}
  \caption{MariFormer: The illustration of generating the cropped image based on the given mask for sequence ``kope100-00006790''.}
  \label{fig:techreports/mari-former}
\end{figure}

\textbf{Limitation}: we observe that the proposed method show limited performance for the segmentation of the obstacles that will cause danger. There are many wrongly detection for this kind of object category. To alleviate this, we may introduce the other cross-modal supervision (from IMU information) to formulate additional constraint, which may lead to further improvement. 

\textbf{Best submission}: The best submission of this method achieved the average score 93.2 shown in Table~\ref{tab:techreports/mari-former}. 

\begin{table}
    \caption{The best submission result of our method.\label{tab:techreports/mari-former}}
	\centering
\resizebox{\linewidth}{!}{
\begin{tabular}{cccccccccc}
    \toprule
    Methods & Average & $\mu_{A}$ & $\mu_{R}$ & Pr & Re & F1 & Pr$_D$ & Re$_D$ & F1$_D$
    \\ \midrule
    MariFormer & 93.2 &10.47	&98.6	&89.7	&97.5	&93.5	&95.5	&90.6	&93.0 \\ 
    \bottomrule
    \end{tabular}
}
\end{table}

\subsubsection{RevDeep}\label{usv-seg:hkust-revdeep}
% \label{usv-seg:hkust-revdeep}
% \noindent
% \emph{Tuan-Anh Vu, Hai Nguyen-Truong, Sai-Kit Yeung}\\
% \texttt{\small{\{tavu,thnguyenab\}@connect.ust.hk, saikit@ust.hk}}\\
% \emph{Hong Kong University of Science and Technology (HKUST)}\\

% TODO: reference
Deeplab~\cite{chen2018deeplab} is one of the most influential neural architectures for semantic segmentation. In this report, we revisit Deeplabv3 and modify it to USV Obstacle Segmentation task. Our model is constructed by a Deeplabv3 model using a ResNet-101 backbone~\cite{he2016deep}. Firstly, we use combinations of data augmentations from color transform (color jitter, random gamma), noise transform (Gauss Noise, ISO Noise), and image transform (Horizontal and Vertical Flip, Shift, Rotate, Scale, Random Brightness and Contrast, and CLAHE). Then, we found that the efficiency of a neural network depends on loss functions, optimizers, learning rate schedulers, and hyperparameters. Our loss is the combination of CrossEntropy with label smoothing and Focal loss, then our loss is optimized using AdamW optimizer~\cite{loshchilov2018decoupled}, Cosine Decay, and learning rate = 1e-6. The network was trained on the mix of MaSTr1325~\cite{bovcon2019mastr}, and 153 additional hard examples from WaSR-T paper~\cite{Zust2022Temporal} to force the network to learn deeper features. Our network was trained with a batch size of 16 for 100 epochs and on a single NVIDIA RTX 3090 GPU. Finally, our model can achieve the average of F1 and F1D scores at 91.7\%.

\subsection{APTX003}
\label{usv-seg:xi-aptx}
\noindent
\emph{Zhuang Jia}\\
\texttt{\small{jiazhuang@xiaomi.com}}\\
\emph{Xiaomi Inc.}\\

Our model named APTX003 in leaderboard is based on DeepLabV3+ with ResNet101 backbone, we use the implementation of this model in Segmentation Models PyTorch~\cite{segmodelsrepo}. The segmentation network is trained on MaSTr1325 dataset for total 100 epochs using Adam optimizer with initial learning rate of 5e-4. After 80 epochs, the learning rate drops to 1e-5. The loss function is cross entropy loss.
For data augmentation, the input image is firstly resized to 512x512, then horizontal flip (vertical flip is not used concerning the relative positions of sky and water) and random brightness/contrast are applied to the resized image. In inference phase, conditional random field (CRF) is used to refine the output segmentation maps with the color information of input image. Moreover, as the large obstacles often tend to be tattered in the final predictions, with edges well detected yet planes may not, we conduct morphological post-processing to ”fill” the holes inside the obstacle regions.
The training and inference processes are all conducted in a Ubuntu server with Nvidia GeForce RTX 3090 GPU. Inference speed is circa 3fps including CRF and post-processing time.

\subsection{HRNet-OCR}
\label{usv-seg:hrnet}
\noindent
\emph{Sophia Yang$^1$, Chih-Chung Hsu$^2$, Xiu-Yu Hou$^2$, Yu-An Jhang$^2$}\\
\texttt{\small{sophias94171@gmail.com, cchsu@gs.ncku.edu.tw, \{xiuyu.hou.tw,styuanpj4\}@gmail.com }}\\
\emph{$^1$National Tsing Hua University (NTHU), $^2$National Cheng Kung University (NCKU)}\\

We propose a High-resolution network (HRNet) incorporating with Transformer as the downstream head, as suggested in \cite{HRNetv2}, for this challenge to effectively address the issues brought by small objects in Obstacle Segmentation track under USV sequences (Track 3). The team members include Chih-Chung Hsu, Xiu-Yu Hou, Yu-An Jhang, and Sophia Yang. We provide the implementation details and the network architecture in the following subsections.

\textbf{Improved HRNet-OCR:} The main issue in this challenge is that classifying the pixels in a given image into one of three classes: sky, water, or obstacle. We take advantage of HRNet for object segmentation, as well as adopt Object-Contextual
Representations (OCR) to achieve higher performance on the  MaSTr1325 datasets. The low-resolution features are upsampled with bilinear interpolation for multi-scale feature fusion.

\begin{figure}[t]
  \centering
  \includegraphics[width=\linewidth]{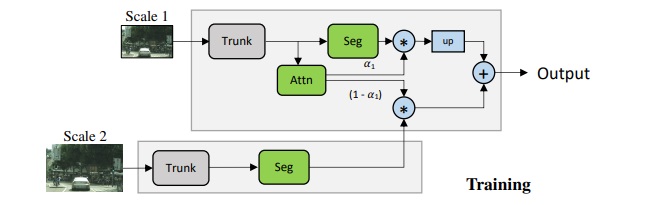}
  \caption{Hierarchical Attention Network Architecture}
  \label{fig:techreports/hrnet}
\end{figure}

\textbf{Network Architecture:} The network structure of the HRNet-OCR is presented in Figure \ref{fig:techreports/hrnet}. We adopt the pretrained weights based on HRNetV2 ImageNet to quickly and effectively train our HRNet-OCR. Note that we did not adopt the pretrained weights from segmentation datasets.

%-------------------------------------------------------------------------
\textbf{Training Tricks and Results:} We deploy the HRNet-OCR on a personal computer equipped with Tesla V100 with 4 FPS inference. We trained the model on four V100s with batch size 6 per GPU for 484 epochs. For training, the image size is 1024*512. We use SGD as the optimizer with weight decay $5\times 10e^{-5}$, and the learning rate is 0.01 with default learning rate decay scheduling.

\subsection{WaSR Baseline}
\label{usv-seg:wasr-baseline}
\emph{MaCVi Organizers}\\

WaSR~\cite{Bovcon2021} is a segmentation network based on the DeepLabv2 architecture. The decoder is redesigned and includes several attention refinement (ARM) and feature fusion (FFM) modules which adaptively reweigh the feature channels to improve the performance. Another critical component of WaSR is the auxiliary water separation loss, which encourages the separation of features of water and other classes early in the encoder. For training we follow the procedure in \cite{Bovcon2021}. We train the model on the MaSTr1325 dataset. ResNet-101 is used as the backbone. Image augmentation, including random rotation, horizontal flipping, random color transformations and adding noise, is used to introduce additional variation. We train for 50 epochs with a batch size of 8 per GPU on 2 V100 GPUs.

\subsection{DeepLabv3 Baseline}
\label{usv-seg:deeplab-baseline}
\emph{MaCVi Organizers}\\

We train the DeepLabv3~\cite{chen2017rethinking} model on the MaSTr1325 dataset. We use the built-in PyTorch implementation of the network. In training we follow the exact setup as WaSR (Section~\ref{usv-seg:wasr-baseline}) including image augmentation.

%USV-related technical reports
\section{USV-based Obstacle Detection}
\subsection{DetectoRS}
\label{usv-det:detectors}
\noindent
\emph{Lars Sommer$^1$$^2$, Raphael Spraul$^1$$^2$}\\
\texttt{\small{\{lars.sommer, raphael.spraul\}\\@iosb.fraunhofer.de}}\\
\emph{$^1$Fraunhofer IOSB, $^2$Fraunhofer Center for Machine Learning}\\
To generate our detections, we used DetectoRS~\cite{qiao2021detectors} with Cascade R-CNN and ResNet-50. 
For initialization, we used weights pre-trained on MS COCO. 
To account for small object dimensions, we set the “scales” parameter to 2, yielding smaller anchor boxes. 
All other parameters remained unchanged. 
SGD was used as optimizer with an initial learning rate of 0.02, a momentum of 0.9 and a weight decay of 0.0001. 
The model was trained for 12 epochs. 
For our submission, we used the model after eight epochs, as it achieved the highest AP values on a validation set.

We used five different datasets to train our model: ABOships~\cite{iancu2021aboships}, MODD~\cite{KristanCYB2015}, MODD2~\cite{bovcon2018stereo}, SeaDronesSee v2\footnote{\url{https://seadronessee.cs.uni-tuebingen.de/dataset}} and a small set of images with ships and buoys crawled from the Internet.
Except for SeaDronesSee, we re-annotated parts of the other dataset.
We used 305 images of ABOships, 269 images of MODD, 794 images of MODD2, 1142 images of SeaDronesSee v2 and 165 images from the Internet. 
For ABOships, we varied the input image scales between 1152x648 and 1600x1200 pixels.
For MODD, we varied the input image scales between 576x432 and 768x576 pixels.
For MODD2, we varied the input image scales between 1150x862 and 1600x1200 pixels.
For SeaDronesSee v2, we varied the input image scales between 1920x1080 and 2688x1512 pixels.
For images from the Internet, we varied the input image scales between 1728x972 and 2376x1296 pixels.
We further applied random gamma, random brightness contrast and RGB shift as data augmentation techniques.
For inference, we applied multiscale testing (1600x1200 and 2000x1500 pixels).
To reduce the number of false negatives, we filtered out detections with a confidence score below 0.3.

The implementation provided by MMDetection~\cite{mmdetection} - an open source object detection toolbox based on PyTorch – was used to train our detector. 
We used 2 Tesla V100 GPUS (CPU: Intel Xeon E5-2698 v4 @ 2.20GHz). 
The inference speed of the detector was about 5 FPS.

\subsection{PRBNet}
\label{usv-det:prbnet}
\noindent
\emph{Sophia Yang$^1$, Chih-Chung Hsu$^2$, Xiu-Yu Hou$^2$, Yu-An Chang$^2$}\\
\texttt{\small{ sophias94171@gmail.com, cchsu@gs.ncku.edu.tw, \{ xiuyu.hou.tw, styuanpj4\}@gmail.com}}\\
\emph{$^1$National Tsing Hua University, $^2$National Cheng Kung University}\\

We propose a Prior-Guided Parallel Residual Bi-Fusion Feature Pyramid Network (PPRB-FPN) for this challenge to effectively address the issues brought by small objects for Obstacle Detection for USV sequences (Track 4). The team members include Sophia Yang, Chih-Chung Hsu, Xiu-Yu Hou, and Yu-An Chang. We provide the implementation details as well as the network architecture in the following subsections.
% \subsection{Mission}

% The mission of the track is Obstacle Detection for USV sequences. There exist 3 challenges in this track: 1) small and unusual-seen objects; 2) Human and boat objects not labeled in suggested training dataset; and 3) Unrelated shore region to disturb the detection tasks.

\subsubsection{Proposed PPRB-FPN}
The main issue in this challenge is that the object or obstacles could be very small, leading to the information vanishing problem. We take advantage of PRB-FPN \cite{chen2021parallel} for small object detection, as well as adopt the state-of-the-art single-stage object detector, i.e., YOLOv7 \cite{wang2022yolov7} to effectively detect the obstacles. With the extensive data analysis of the given UAV sequences, the mask of the shores given by organizers is used in the training and inference phase to reduce the false alarm, termed as prior-based PRB-FPN.

\begin{figure}[tb]
  \centering
  \includegraphics[width=0.45 \textwidth]{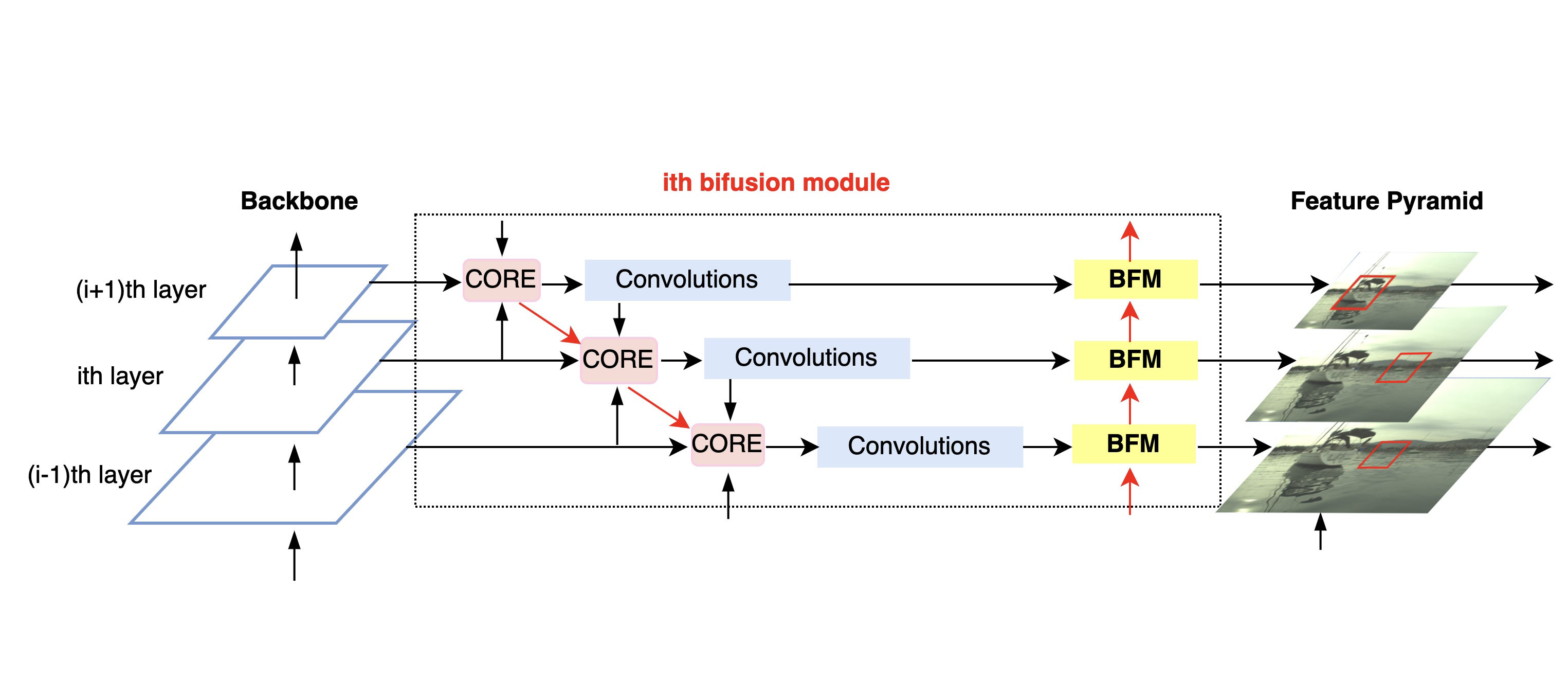}
  \caption{Bi-Fusion Feature Pyramid of PPRBNet
  }
  \label{fig:usv-det:prb}
\end{figure}

\noindent Network Architecture. {\bf PRBNet}\cite{chen2021parallel}: The network structure of the PRBNet is presented in Figure \ref{fig:usv-det:prb}. We enlarge the input size to 1280 to achieve better accuracy on small objects and adopt the pretrained weights based on MS-COCO to quickly and effectively train our PRB-FPN.
\\

\noindent {\bf Prior-Guided Labeling and Masking}: Most of the obstacles are not presented in MS-COCO dataset, so we label all of the objects not classified as human and boat into other. While this approach can easily use the pretrained model with other datasets, it also triggers a lot of false positives on unrelated regions. We choose to filter these false alarms out by applying the prior that objects on shores are actually not affecting USV sailing.

%-------------------------------------------------------------------------
\subsubsection{Training Tricks and Results}
We deploy our method on a machine with Intel Xeon Gold 6248R and Tesla V100-DGXS and achieve a 6 fps frame rate, which is mostly taken for inferencing with PRBNet. The total number of the training process is 500, 500 with the AdamW optimizer. The step decay learning rate 0.001. Then the learning rate is multiplied by a factor 0.01 at the 400, 000 steps and 450, 000 steps, respectively. PRB model were trained on a single V100 with batch size 64 and 400 epochs. The data augmentation strategies used in this paper are Mixout, random rotation, CutMix, and color jitters. 

\subsection{OceanU}
\label{usv-det:oceanu}
\noindent
\emph{Simon Yang$^1$, Mau-Tsuen Yang$^2$}\\
\texttt{\small{lordolddog@gmail.com}}\\
\emph{$^1$National Taiwan Ocean University, Taiwan, $^2$National Dong-Hwa University, Taiwan}\\

\noindent Neural network architecture: YOLOv7\cite{yolov7github}.

\noindent Adaptations of E-ELAN: The computational block is modified by using expand, shuffle, merge cardinality to continuously enhance the learning ability without destroying the original gradient path.

\noindent Detection: \texttt{python detect.py --weights yolov7-e6e.pt --conf 0.1 –iou 0.25 --img-size 1280}

\noindent Training: \texttt{python train\_aux.py --workers 4 --device 0 --batch-size 16 --data data/coco.yaml --img 1280 1280 --cfg cfg/training/yolov7-e6e.yaml --weights 'yolov7-e6e.pt' --name yolov7-e6e --hyp data/hyp.scratch.p6.yaml}

\noindent Train dataset: MS COCO dataset

\noindent Local machine: Nvidia GeForce RTX 2060 GPU / Intel i7-9700 CPU 

\noindent Inference speed: 50ms on GPU / 3000ms on CPU

\noindent Adaptations for MODS: There are 80 categories in COCO dataset. We replace type 0 by ‘person’, type 8 by ‘ship’, and all other types by ‘other’ so only three categories are recognized.

\noindent Observations: Instead of detecting everything, should focus on detecting objects on the water area.

{\small
\bibliographystyle{ieee_fullname}
\bibliography{egbib}
}

\end{document}